\documentclass[10pt,twocolumn,letterpaper]{article}

\usepackage{cvpr}
\usepackage{times}
\usepackage{epsfig}
\usepackage{graphicx}
\usepackage{amsmath}
\usepackage{amssymb}
\usepackage{mathptmx}
\usepackage{caption}
\usepackage{subcaption}
\usepackage{comment}

\def\E{{\rm E}}
\def\N{{\rm N}}
\def\KL{{\rm KL}}

\def\P{P_{\rm data}}

\def\tY{\tilde{Y}}
\def\tn{\tilde{n}}
\def\hY{\hat{Y}}

\usepackage{url}            
\usepackage{booktabs}       
\usepackage{amsfonts}       
\usepackage{nicefrac}       
\usepackage{microtype}      
\usepackage{algorithm}
\usepackage{algorithmic}

\usepackage[pagebackref=true,breaklinks=true,letterpaper=true,colorlinks,bookmarks=false]{hyperref}

 \cvprfinalcopy 

\def\cvprPaperID{4307} 

\ifcvprfinal\fi
\begin{document}

\title{Learning Descriptor Networks for 3D Shape Synthesis and Analysis}

\newcommand*\samethanks[1][\value{footnote}]{\footnotemark[#1]}

\author{Jianwen Xie$^{1}$\thanks{Equal contributions.} , Zilong Zheng$^{2}$\samethanks[1] , Ruiqi Gao$^{2}$, Wenguan Wang$^{2,3}$, Song-Chun Zhu$^{2}$, Ying Nian Wu$^{2}$\\
$^{1}$Hikvision Research Institute 
$^{2}$University of California, Los Angeles 
$^{3}$Beijing Institute of Technology\\
}


\maketitle

\begin{abstract}


This paper proposes a 3D shape descriptor network, which is a deep convolutional energy-based model, for  modeling volumetric shape patterns. The maximum likelihood training of the model follows an ``analysis by synthesis'' scheme and can be interpreted as a mode seeking and mode shifting process. The model can synthesize 3D shape patterns by sampling  from the probability distribution via MCMC such as Langevin dynamics. The model can be used to train a 3D generator network via MCMC teaching. The conditional version of the 3D shape descriptor net can be used for 3D object recovery and 3D object super-resolution. Experiments demonstrate that the proposed model can generate realistic 3D shape patterns and can be useful for 3D shape analysis.

\end{abstract}

\section{Introduction}

\subsection{Statistical models of 3D shapes}

Recently, with the introduction of large 3D CAD datasets, e.g., ShapeNet \cite{wu20153d, chang2015shapenet}, some interesting attempts \cite{girdhar2016learning, su2015multi,qi2016volumetric} have been made on object recognition and synthesis based on voxelized 3D shape data. From the perspective of statistical modeling, the existing 3D models can be grouped into two main categories: (1) 3D discriminators, such as Voxnet \cite{maturana2015voxnet}, which aim to learn a mapping from 3D voxel input to semantic labels for the purpose of 3D object classification and recognition, and (2) 3D generators, such as 3D-GAN\cite{3dgan}, which are in the form of latent variable models that assume that the 3D voxel signals are generated by some latent variables. The training of discriminators usually relies on big data with annotations and is accomplished by a direct minimization of the prediction errors, while the training of the generators  learns a mapping from the latent space to 3D voxel data space. 

The generator model, while useful for synthesizing 3D shape patterns, involves a challenging inference step (i.e., sampling from the posterior distribution) in maximum likelihood learning, therefore variational inference \cite{kingma2013auto} and adversarial learning \cite{goodfellow2014generative, radford2015unsupervised, 3dgan} methods are commonly used, where an extra network is incorporated into the learning algorithm to get around the difficulty of the posterior inference. 

The past few years have witnessed impressive progress on developing  discriminator models and generator models for 3D shape data, however, there has not been much work in the literature on modeling 3D shape data based on energy-based models. We call this type of models the descriptive models or descriptor networks following \cite{zhu2003statistical}, because the models describe the data based on bottom-up descriptive features learned from the data. The focus of the present paper is to develop a volumetric 3D descriptor network for voxelized shape data. It can be considered an alternative to 3D-GAN \cite{3dgan} for 3D shape generation. 

\subsection{3D shape descriptor network}
Specifically, we present a novel framework for probabilistic modeling of volumetric shape patterns  by combining the merits of energy-based model \cite{Lecun2006} and volumetric convolutional neural network \cite{maturana2015voxnet}. The model is a probability density function directly defined on voxelized shape signal, and the model is in the form of a deep convolutional energy-based model, where the feature statistics or the energy function is defined by a bottom-up volumetric ConvNet that maps the 3D shape signal to the features. 
We call the proposed model the 3D DescriptorNet, because it uses a volumetric ConvNet to extract 3D shape features from the voxelized data.  

The training of the proposed model follows an ``analysis by synthesis'' scheme \cite{grenander2007pattern}. Different from the variational inference or adversarial learning, the proposed model does not need to incorporate an extra inference network or an adversarial discriminator in the learning process.  The learning and sampling process is guided by the same set of parameters of a single model, which makes it a particularly natural and statistically rigorous framework for probabilistic 3D shape modeling. 

Modeling 3D shape data by a probability density function provides distinctive advantages: First, it is able to synthesize realistic 3D shape patterns by sampling examples from the distribution via MCMC, such as Langevin dynamics. Second, the model can be modified into a conditional version, which is useful for 3D object recovery and 3D object super-resolution. Specifically, a conditional probability density function that maps the corrupted (or low resolution) 3D object to the recovered (or high resolution) 3D object is trained, and then the 3D recovery (or 3D super-resolution) can be achieved by sampling from the learned conditional distribution given the corrupted or low resolution 3D object as the conditional input. Third, the model can be used in a cooperative training scheme \cite{xie2016cooperative}, as opposed to adversarial training, to train a 3D generator model via MCMC teaching. The training of 3D generator in such a scheme is stable and does not encounter mode collapsing issue. Fourth, the model is useful for semi-supervised learning. After learning the model from unlabeled data, the learned features can be used to train a classifier on the labeled data. 

We show that the proposed 3D DescriptorNet can be used to synthesize realistic 3D shape patterns, and its conditional version is useful for 3D object recovery and 3D object super-resolution. The 3D generator trained by 3D DescriptorNet in a cooperative scheme carries semantic information about 3D objects. The feature maps trained by 3D DescriptorNet in an unsupervised manner are useful for 3D object classification.   

\subsection{Related work}

\textit{3D object synthesis.} Researchers in the fields of graphics and vision have studied the 3D object synthesis problems \cite{blanz1999morphable, carlson1982algorithm, kalogerakis2012probabilistic}. However, most of these object synthesis methods are nonparametric and they generate new patterns by retrieving and merging parts from an existing database. Our model is a parametric probabilistic model that requires learning from the observed data.  3D object synthesis can be achieved by running MCMC such as Langevin dynamics to draw samples from the learned distribution.

\textit{3D deep learning.} Recently, the vision community has witnessed the success of deep learning, and  researchers have used  the models in the field of deep learning, such as convolutional deep belief network \cite{wu20153d}, deep convolutional neural network \cite{maturana2015voxnet}, and deep convolutional generative adversarial nets (GAN) \cite{3dgan}, to model 3D objects for the sake of synthesis and analysis. Our proposed 3D model is also powered by the ConvNets. It incorporates a bottom-up 3D ConvNet structure for defining the probability density, and learns the parameters of the ConvNet by an ``analysis by synthesis'' scheme.   

\textit{Descriptive models for synthesis.} 
Our model is related to the following descriptive models. The FRAME (Filters, Random field, And Maximum Entropy) \cite{zhu1998filters} model, which was developed for modeling  stochastic textures. The sparse FRAME model \cite{xie2015learning, xie2016inducing}, which was used for modeling object patterns. 
Inspired by the successes of deep convolutional neural networks (CNNs or ConvNets), \cite{lu2015learning} proposes a deep FRAME model, where the linear filters used in the original FRAME model are replaced by the non-linear filters at a certain convolutional layer of a pre-trained deep ConvNet. 
Instead of using filters from a pre-trained ConvNet, \cite{XieLuICML} learns the ConvNet filters from the observed data by maximum likelihood estimation. The resulting model is called generative ConvNet, which can be considered a recursive multi-layer generalization of the original FRAME model. 


Building on the early work of \cite{tu2007learning}, recently \cite{jin2017introspective, lazarow2017introspective} have developed an introspective learning method to learn the energy-based model, where the energy function is discriminatively learned. 

\subsection{Contributions}
(1) We propose a 3D deep convolutional energy-based model that we call  3D DescriptorNet for  modeling 3D object patterns by combining the volumetric ConvNets \cite{maturana2015voxnet} and the generative ConvNets \cite{XieLuICML}. (2) We present a mode seeking and mode shifting interpretation of the learning process of the model. (3) We present an adversarial interpretation of the zero temperature limit of the learning process. (4) We propose a conditional learning method for recovery tasks. (5) we  propose  metrics that can be useful for evaluating 3D generative models. (6) A 3D cooperative training scheme is provided as an alternative to the adversarial learning method to train 3D generator.    


\section{3D DescriptorNet}
 
\subsection{Probability density}
The 3D DescriptorNet is a 3D deep convolutional energy-based model defined on the volumetric data $Y$, which is in the form of exponential tilting of a reference distribution \cite{XieLuICML}: 
\begin{eqnarray} 
   p(Y; \theta) = \frac{1}{Z(\theta)} \exp\left[ f(Y; \theta)\right] p_0(Y), 
\end{eqnarray}
 where $p_0(Y)$ is the reference distribution such as Gaussian white noise model, i.e.,
$ p_0(Y) \propto \exp \left( -{\|Y\|^2}/{2 s^2}\right), 
$  
 $f(Y; \theta)$ is defined by a bottom-up 3D volumetric ConvNet whose parameters are denoted by $\theta$. $Z(\theta) = \int  \exp\left[ f(Y; \theta)\right] p_0(Y) dY$ is the normalizing constant or partition function that is analytically intractable. The energy function is
 \begin{eqnarray} 
   {\cal E}(Y; \theta) = \frac{\|Y\|^2}{2s^2} - f(Y; \theta). 
\end{eqnarray}
We may also take $p_0(Y)$ as uniform distribution within a bounded range. Then ${\cal E}(Y; \theta) = - f(Y; \theta)$. 

 \subsection{Analysis by synthesis} 
 The maximum likelihood estimation (MLE) of the 3D DescriptorNet follows an ``analysis by synthesis'' scheme. Suppose we observe 3D training examples $\{Y_i, i = 1, ..., n\}$ from an unknown data distribution $\P(Y)$. The MLE seeks to maximize the log-likelihood function 
$   L(\theta) = \frac{1}{n} \sum_{i=1}^{n} \log p(Y_i; \theta). 
$ 
If the sample size $n$ is large, the maximum likelihood estimator minimizes $\KL(\P \parallel p_\theta)$, the Kullback-Leibler divergence from the data distribution $\P$ to the model distribution $p_\theta$. 
The gradient of the $L(\theta)$ is 
  \begin{eqnarray} 
  L'(\theta) = \frac{1}{n} \sum_{i=1}^{n} \frac{\partial}{\partial \theta} f(Y_i; \theta) - \E_{\theta} \left[\frac{\partial}{\partial \theta} f(Y; \theta)\right],  \label{eq:lD}
\end{eqnarray} 
where  $\E_{\theta}$ denotes the expectation with respect to $p(Y; \theta)$. The expectation term in equation (\ref{eq:lD}) is due to
$\frac{\partial}{\partial \theta}  \log Z(\theta) = \E_{\theta}[\frac{\partial}{\partial \theta}  f(Y; \theta)]$, which is analytically intractable and has to be approximated by MCMC, such as Langevin  dynamics,  which iterates the following step: 
\begin{eqnarray}
   Y_{\tau+\Delta \tau} &=& Y_\tau - \frac{\Delta \tau}{2} \frac{\partial}{\partial Y}  {\cal E}(Y_\tau; \theta) + \sqrt{\Delta \tau}\epsilon_{\tau} \nonumber\\ 
   &=& Y_\tau - \frac{\Delta \tau}{2} \left[ \frac{Y_\tau}{s^2} - \frac{\partial}{\partial Y} f(Y_\tau; \theta) \right] + \sqrt{ \Delta \tau}\epsilon_{\tau},
    \label{eq:LangevinD}
\end{eqnarray}
where $\tau$ indexes the time steps of the Langevin dynamics, $\Delta \tau$ is the discretized step size, and $\epsilon_{\tau} \sim \N(0, I)$ is the Gaussian white noise term.  The Langevin dynamics consists of a  deterministic part, which is a gradient descent on a landscape defined by ${\cal E}(Y; \theta)$, and a stochastic part, which is a Brownian motion that helps  the chain to escape spurious local minima of the energy ${\cal E}(Y; \theta)$. 

Suppose we draw $\tilde{n}$ samples $\{ \tY_i, i=1,...,\tilde{n} \}$ from the distribution $p(Y;\theta)$ by running $\tilde{n}$ parallel chains of Langevin dynamics according to (\ref{eq:LangevinD}). The gradient of the log-likelihood $L{(\theta)}$ can be approximated by 
  \begin{eqnarray} 
  L'(\theta) \approx \frac{1}{n} \sum_{i=1}^{n} \frac{\partial}{\partial \theta} f(Y_i; \theta) - \frac{1}{\tilde{n}} \sum_{i=1}^{ \tilde{n} } \frac{\partial}{\partial \theta} f(\tilde Y_i; \theta).  \label{eq:lD2}
\end{eqnarray}

 \subsection{Mode seeking and mode shifting} 
 
The above ``analysis by synthesis'' learning scheme can be interpreted as a mode seeking and mode shifting process. We can rewrite equation (\ref{eq:lD2}) in the form of  
 \begin{eqnarray} 
  L'(\theta) \approx \frac{\partial}{\partial \theta} \left[ \frac{1}{\tilde{n}} \sum_{i=1}^{ \tilde{n} }   {\cal E}(\tilde Y_i; \theta) - \frac{1}{n} \sum_{i=1}^{n}   {\cal E}(Y_i; \theta) \right]. \label{eq:lD3}
\end{eqnarray}
We define a value function
 \begin{eqnarray} 
 V(\{\tY_i\}; \theta) = \frac{1}{\tilde{n}} \sum_{i=1}^{ \tilde{n} }   {\cal E}(\tilde Y_i; \theta) - \frac{1}{n} \sum_{i=1}^{n}   {\cal E}(Y_i; \theta). \label{eq:V}
\end{eqnarray}
The equation (\ref{eq:lD3}) reveals that the gradient of the log-likelihood $L{(\theta)}$ coincides with the gradient of $V$. 

The sampling step in (\ref{eq:LangevinD}) can be interpreted as mode seeking, by finding low energy modes or high probability modes in the landscape defined by $ {\cal E}(Y;\theta)$ via stochastic gradient descent (Langevin dynamics) and placing the synthesized examples around the modes. It seeks to decrease $V$.   The learning step can be interpreted as mode shifting (as well as mode creating and mode sharpening) by shifting the low energy modes from the synthesized examples $\{\tY_i\}$ toward the observed examples $\{Y_i\}$. It seeks to increase $V$. 


 
  The training algorithm of the 3D DescriptorNet is presented in Algorithm \ref{code:3D}.

\begin{algorithm}[h]
\caption{3D DescriptorNet}
\label{code:3D}
\begin{algorithmic}[1]
\REQUIRE ~~\\
(1) training data $\{Y_i, i=1,...,n\}$; (2) number of Langevin steps $l$; (3) number of learning iterations $T$.

\ENSURE~~\\
(1) estimated parameters $\theta$; (2) synthesized examples $\{\tY_i, i = 1, ..., \tilde{n}\}$ 

\item[]
\STATE Let $t\leftarrow 0$, initialize $\theta^{(0)}$, initialize $\tY_i$, for $i = 1, ..., \tilde{n}$. 
\REPEAT 
\STATE \textbf{Mode seeking}: For each $i$, run $l$ steps of Langevin dynamics to revise $\tY_i$, i.e., starting from the current $\tY_i$, each step follows equation (\ref{eq:LangevinD}). 
\STATE \textbf{Mode shifting}: Update $\theta^{(t+1)} = \theta^{(t)} + \gamma_t L'(\theta^{(t)}) $,  with learning rate $\gamma_t$, where $L'(\theta^{(t)})$ is computed according to (\ref{eq:lD2}). 
\STATE Let $t \leftarrow t+1$
\UNTIL $t = T$
\end{algorithmic}
\end{algorithm}

  \subsection{Alternating back-propagation} 

Both mode seeking (sampling) and mode shifting (learning) steps involve the derivatives of $f(Y;\theta)$ with respect to $Y$ and $\theta$ respectively. Both derivatives can be computed efficiently by back-propagation. The algorithm is thus in the form of alternating back-propagation that iterates the following two steps: (1) Sampling back-propagation: Revise the synthesized examples by Langevin dynamics or gradient descent.
(2) Learning back-propagation: Update the model parameters
given the synthesized and the observed examples by gradient ascent. 

\subsection{Zero temperature limit} 


We can add a temperature term to the model $p_T(Y; \theta) = \exp(- {\cal E}(Y; \theta)/T)/Z_T(\theta)$, where the original model corresponds to $T = 1$. At zero temperature limit as $T \rightarrow 0$, the Langevin sampling will become gradient descent where the noise term diminishes in comparison to the gradient descent term. The resulting algorithm approximately solves the minimax problem below
  \begin{eqnarray} 
\max_{\theta} \min_{\{\tY_i\}}  V(\{\tY_i\}; \theta)  \label{eq:minimax}
\end{eqnarray}
with $\tY_i$ initialized from an initial distribution and approaching local modes of $V$.  We can regularize either the diversity of $\{\tY_i\}$ or the smoothness of ${\cal E}(Y; \theta)$.  This is an adversarial interpretation of the learning algorithm. It is also a generalized version of herding \cite{welling2009herding} and is related to \cite{arjovsky2017wasserstein}. In our experiments, we find that disabling the noise term of the Langevin dynamics in the later stage of the learning process often leads to better synthesis results.  Ideally the learning algorithm should create a large number of local modes with similar low energies to capture the diverse observed examples as well as unseen examples.

\subsection{Conditional learning for recovery}

 The conditional distribution $p(Y | C(Y) = c; \theta)$ can be derived from $p(Y; \theta)$. This conditional form of the 3D DescriptorNet can be used for recovery tasks such as inpainting and super-resolution. In inpinating, $C(Y)$ consists of the visible part of $Y$. In super-resolution, $C(Y)$ is the low resolution version of $Y$. For such tasks, we can learn the model from the fully observed training data $\{Y_i, i = 1, ..., n\}$ by maximizing the conditional log-likelihood 
 \begin{eqnarray}
     L(\theta) = \frac{1}{n} \sum_{i=1}^{n} \log p(Y_i \mid C(Y_i) = c_i; \theta), 
 \end{eqnarray} 
 where $c_i$ is the observed value of $C(Y_i)$. 
The learning and sampling algorithm is essentially the same as maximizing the original log-likelihood, except that in the Langevin sampling step, we need to sample from the conditional distribution, which amounts to fixing $C(Y_\tau)$ in the sampling process. 
The  zero temperature limit (with the noise term in the Langevin dynamics disabled) approximately solves the following minimax problem 
  \begin{eqnarray} 
\max_{\theta} \min_{\{\tY_i: C(\tY_i) = {c}_i\}}  V(\{\tY_i\}; \theta).  \label{eq:minimax1}
\end{eqnarray}

\section{Teaching 3D generator net}


We can let a 3D generator network learn from the MCMC sampling of the 3D DescriptorNet, so that the 3D generator network can be used as an approximate direct sampler of the 3D DescriptorNet. 

\subsection{3D generator model} 

The 3D generator model \cite{goodfellow2014generative} is a 3D non-linear multi-layer generalization of the traditional factor analysis model. The generator model has the following form
\begin{eqnarray} 
& Z \sim {\rm N}(0, I_d);\nonumber \\ 
 & Y = g(Z; \alpha)  + \epsilon; \epsilon \sim {\rm N}(0, \sigma^2 I_D). \label{eq:NFA}
 \end{eqnarray}
 where $Z$ is a $d$-dimensional vector of latent factors that follow ${\rm N}(0, 1)$ independently, and the 3D object $Y$ is generated by first sampling $Z$ from its known prior distribution ${\rm N}(0, I_d)$ and then transforming $Z$ to the $D$-dimensional $Y$ by a top-down deconvolutional network $g(Z; \alpha)$ plus the white noise $\epsilon$.  $\alpha$ denotes the parameters of the generator.

\subsection{MCMC teaching of 3D generator net} 

The 3D generator model can be trained simultaneously with the 3D DescriptorNet  in a cooperative training scheme \cite{xie2016cooperative}. The basic idea is to use the 3D generator to generate examples to initialize a finite step Langevin dynamics for training the 3D DescriptorNet. In return, the 3D generator learns from how the Langevin dynamics changes the initial examples it generates. 


 
Specifically, in each iteration, (1) We generate $Z_i$ from its known prior distribution, and then generate the initial synthesized examples  by $\hY_i = g(Z_i; \alpha)  + \epsilon_i$ for $i=1,...,\tilde n$. (2) Starting from the initial examples $\{\hY_i\}$, we sample from the 3D DescriptorNet by running a finite number of steps of MCMC such as Langevin dynamics  to obtain the revised synthesized examples $\{\tY_i\}$. (3) We then update the parameters $\theta$ of the 3D DescriptorNet based on $\{\tY_i\}$ according to (\ref{eq:lD2}), and update the parameters $\alpha$ of the 3D generator by gradient descent 
\begin{eqnarray}
\Delta \alpha \propto - \frac{\partial}{\partial \alpha} \left[\frac{1}{\tn} \sum_{i=1}^{\tilde n} \|\tY_i - g(Z_i; \alpha)\|^2\right]. \label{eq:alpha}
\end{eqnarray}
 We call it MCMC teaching because the revised examples $\{\tY_i\}$ generated by the finite step MCMC are used to teach $g(Z; \alpha)$. 
For each $\tY_i$,  the latent factors $Z_i$ are known to the 3D generator, so that there is no need to infer $Z_i$, and the learning becomes a much simpler supervised learning problem. 
Algorithm \ref{code:3} presents a full description of the learning of a 3D DescriptorNet with a 3D generator as a sampler.   

\begin{algorithm}[h]
\caption{MCMC teaching of 3D generator net}
\label{code:3}
\begin{algorithmic}[1]

\REQUIRE~~\\
(1) training examples $\{Y_i, i=1,...,n\}$, (2) numbers of Langevin steps $l$, (3) number of learning iterations $T$.

\ENSURE~~\\
(1) estimated parameters $\theta$ and $\alpha$, (2) synthetic examples $\{\hY_i, \tY_i, i= 1, ..., \tilde{n}\}$ 

\item[]
\STATE Let $t\leftarrow 0$, initialize $\theta$ and $\alpha$.
\REPEAT 
\STATE {\bf Initializing mode seeking}: For $i = 1, ..., \tn$, generate $Z_i \sim \N(0, I_d)$, and generate $\hY_i = g(Z_i; \alpha^{(t)}) + \epsilon_i$. 
\STATE {\bf Mode seeking}: For $i = 1, ..., \tn$,  starting from $\hY_i$, run $l$ steps of Langevin  dynamics to obtain $\tY_i$,  each step 
following equation (\ref{eq:LangevinD}). 
\STATE {\bf Mode shifting}: Update $\theta^{(t+1)} = \theta^{(t)} + \gamma_t L'(\theta^{(t)})$,  where $L'(\theta^{(t)})$ is computed according to (\ref{eq:lD2}). 
\STATE {\bf Learning from mode seeking}: Update $\alpha^{(t+1)}$ according to (\ref{eq:alpha}).
\STATE Let $t \leftarrow t+1$
\UNTIL $t = T$
\end{algorithmic}
\end{algorithm}

\section{Experiments}
\textbf{Project page}: The code and more results and details can be found at 
\url{http://www.stat.ucla.edu/~jxie/3DDescriptorNet/3DDescriptorNet.html}


\subsection{3D object synthesis} 
\label{Exp:objectSynthesis}

\begin{figure*}[h]
	\centering	
\textbf{{\footnotesize obs1}}\hspace{6mm} \textbf{{\footnotesize obs2}} \hspace{6mm} \textbf{{\footnotesize obs3}} \hspace{6mm} \textbf{{\footnotesize syn1}} \hspace{6mm} \textbf{{\footnotesize syn2}} \hspace{6mm} \textbf{{\footnotesize syn3}} \hspace{6mm} \textbf{{\footnotesize syn4}} \hspace{6mm}
\textbf{{\footnotesize syn5}} \hspace{6mm} \textbf{{\footnotesize syn6}} \hspace{6mm} \textbf{{\footnotesize nn1}} \hspace{6mm}
\textbf{{\footnotesize nn2}} \hspace{6mm} \textbf{{\footnotesize nn3}} \hspace{6mm}
\textbf{{\footnotesize nn4}} \\
	\rotatebox{90}{\hspace{4mm}\textbf{{\footnotesize chair}}}	
	\includegraphics[height=.08\linewidth]{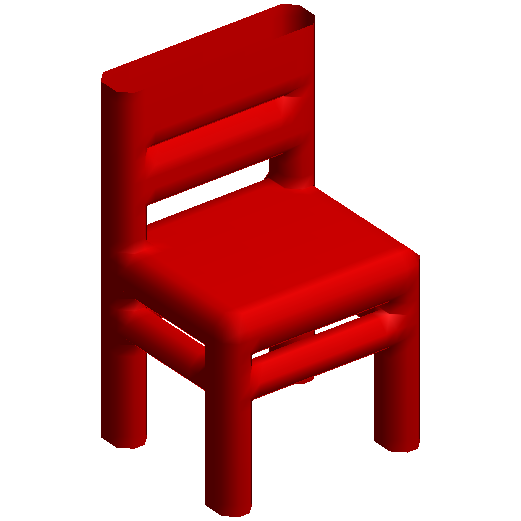} \hspace{-3mm} 
	\includegraphics[height=.08\linewidth]{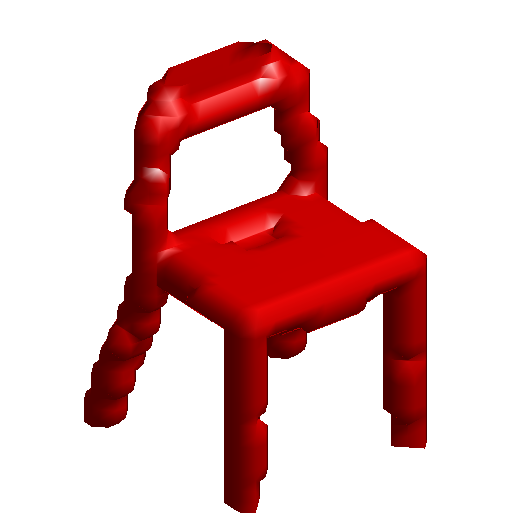} \hspace{-3mm} 
	\includegraphics[height=.08\linewidth]{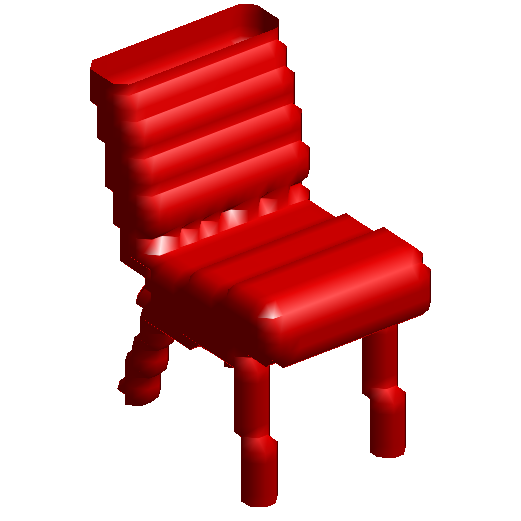} \hspace{-3mm}
    \includegraphics[height=.08\linewidth]{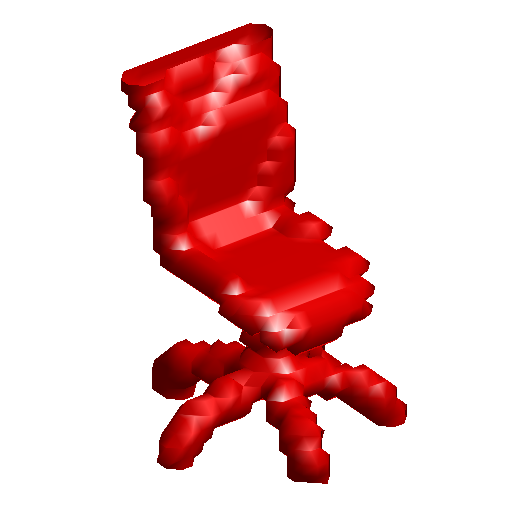} \hspace{-3mm}	
    \includegraphics[height=.08\linewidth]{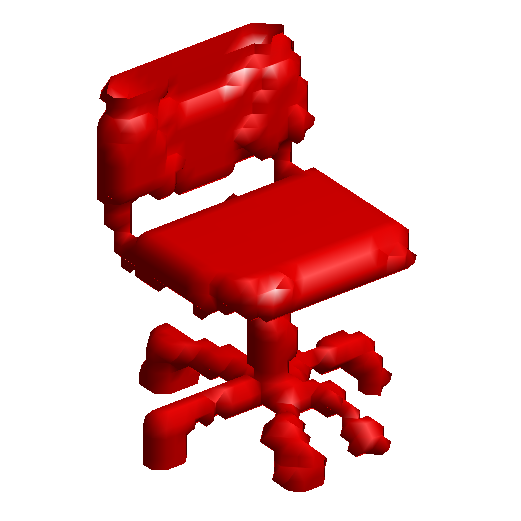} \hspace{-3mm}	
    \includegraphics[height=.08\linewidth]{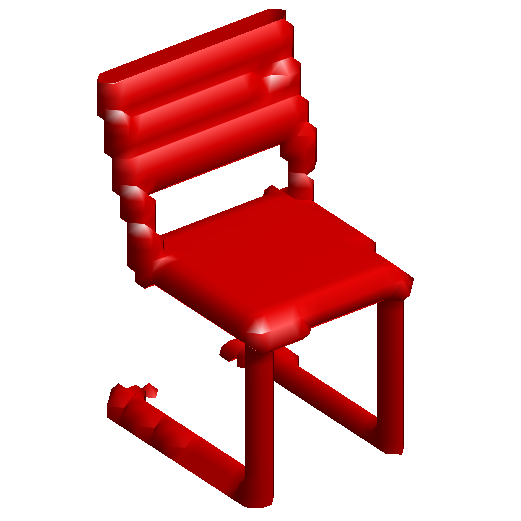} \hspace{-3mm}	
    \includegraphics[height=.08\linewidth]{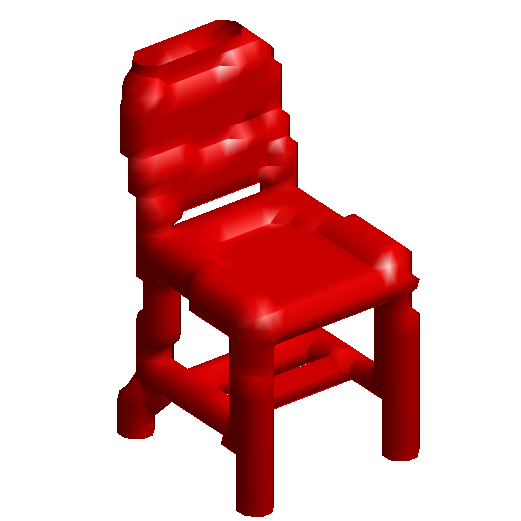} \hspace{-3mm}	
    \includegraphics[height=.08\linewidth]{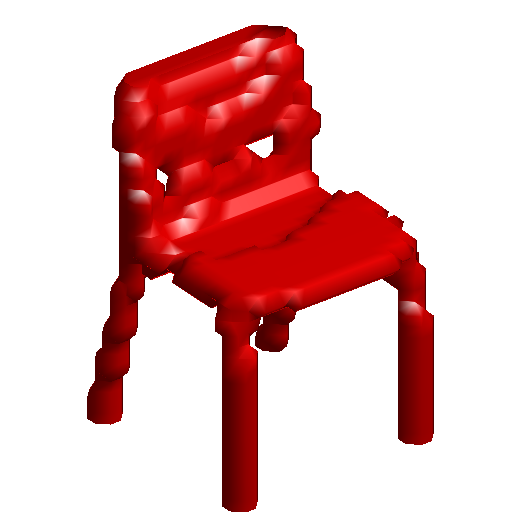} \hspace{-2mm}	
     \includegraphics[height=.08\linewidth]{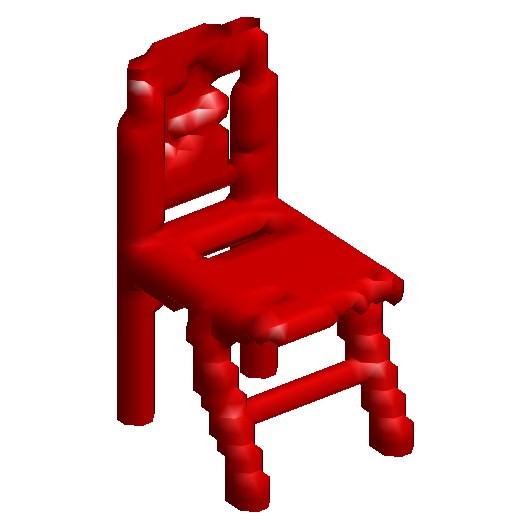}  \hspace{-2mm}   
     \includegraphics[height=.08\linewidth]{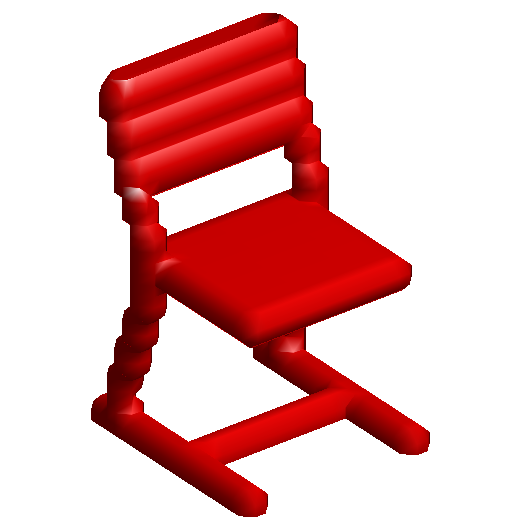} \hspace{-3mm}
     \includegraphics[height=.08\linewidth]{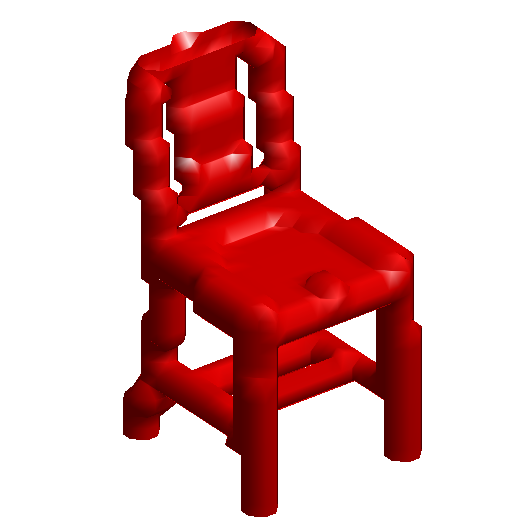}
      \hspace{-3mm}
     \includegraphics[height=.08\linewidth]{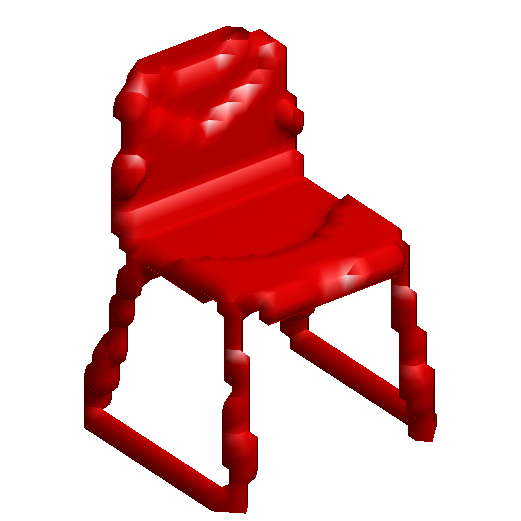} \hspace{-3mm}
     \includegraphics[height=.08\linewidth]{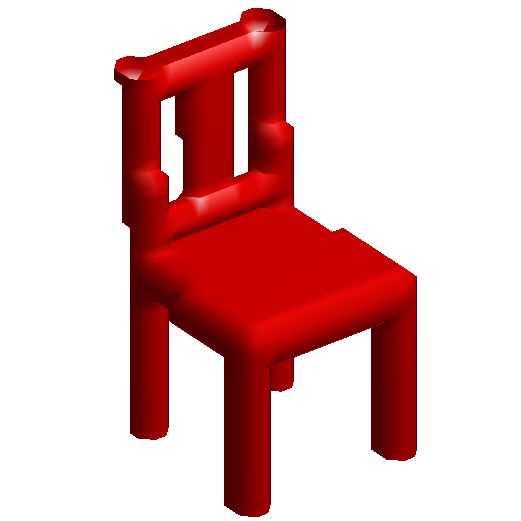}     
      \\ 
    \rotatebox[origin=l]{90}{\hspace{2mm} \textbf{{\footnotesize bed}}}
	\includegraphics[height=.07\linewidth]{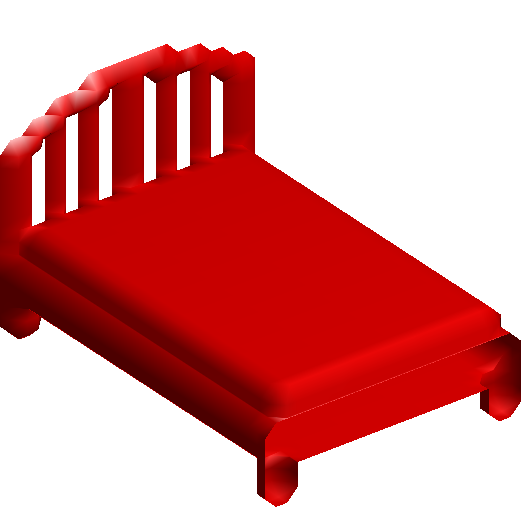} \hspace{-1mm}
	\includegraphics[height=.07\linewidth]{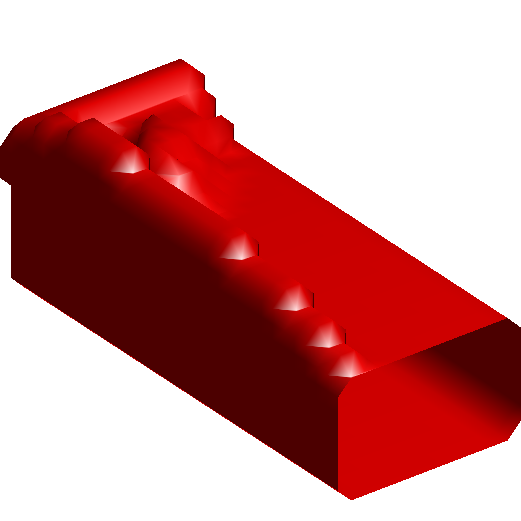} \hspace{-1mm}
	\includegraphics[height=.07\linewidth]{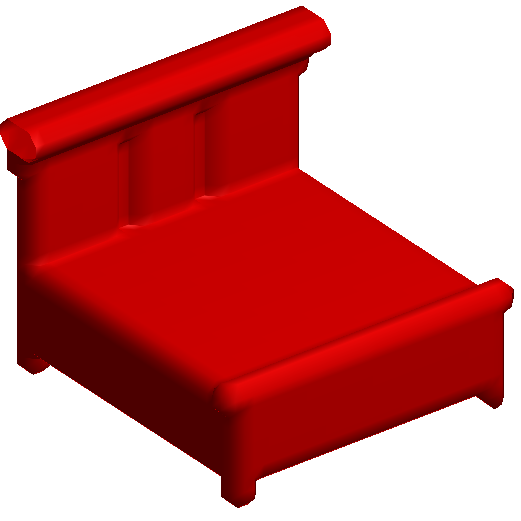}  
	\includegraphics[height=.07\linewidth]{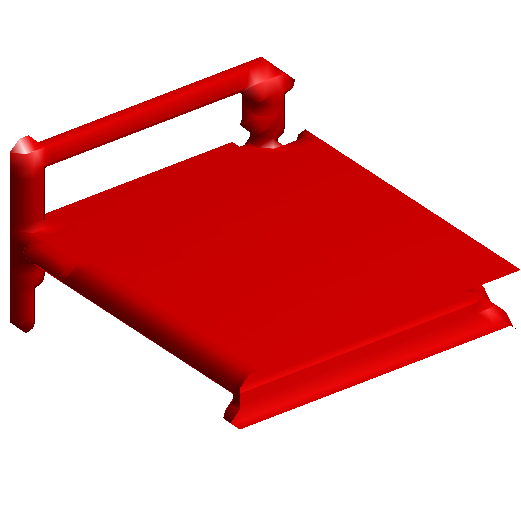}  \hspace{-1mm}
	\includegraphics[height=.07\linewidth]{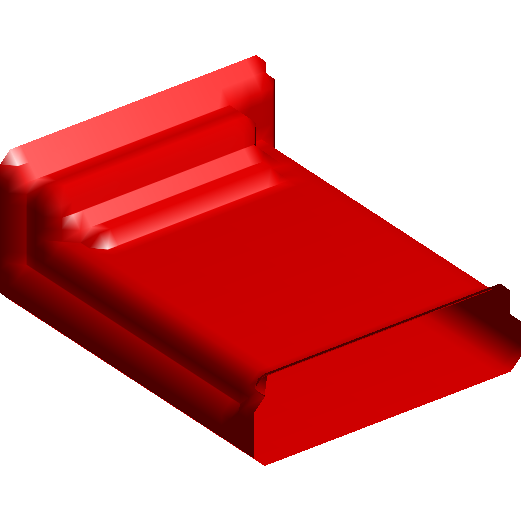}  \hspace{-1mm}
    \includegraphics[height=.07\linewidth]{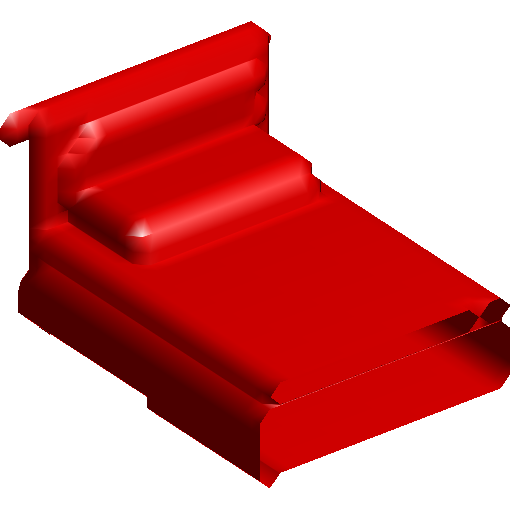}  \hspace{-1mm}
    \includegraphics[height=.07\linewidth]{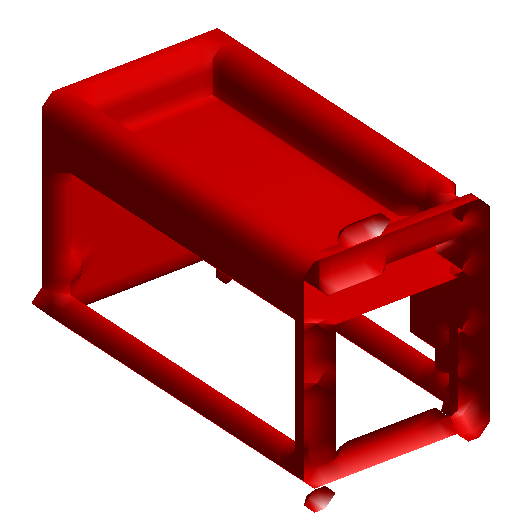}  \hspace{-1mm}
    \includegraphics[height=.07\linewidth]{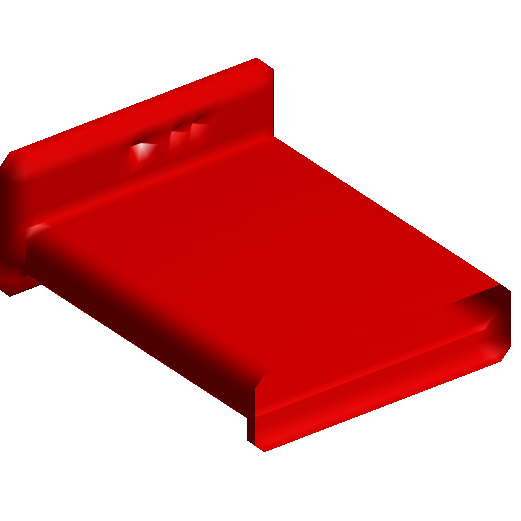} \hspace{-1mm}
    \includegraphics[height=.07\linewidth]{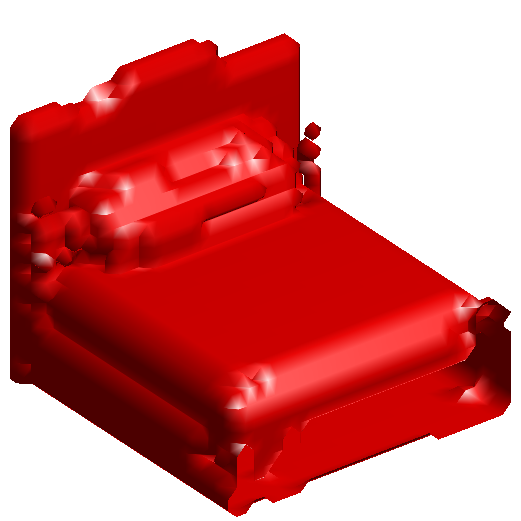}     
    \includegraphics[height=.07\linewidth]{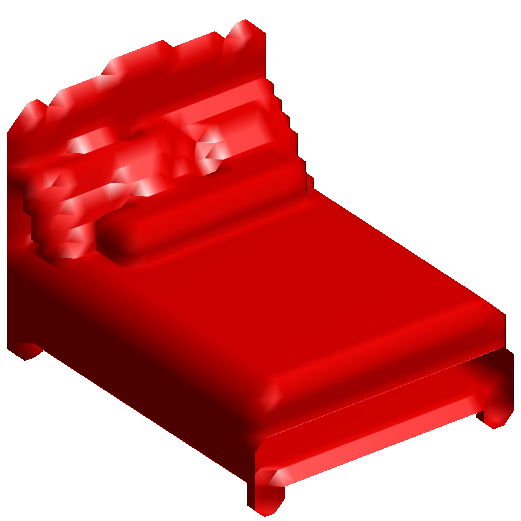}  \hspace{-1mm}
    \includegraphics[height=.07\linewidth]{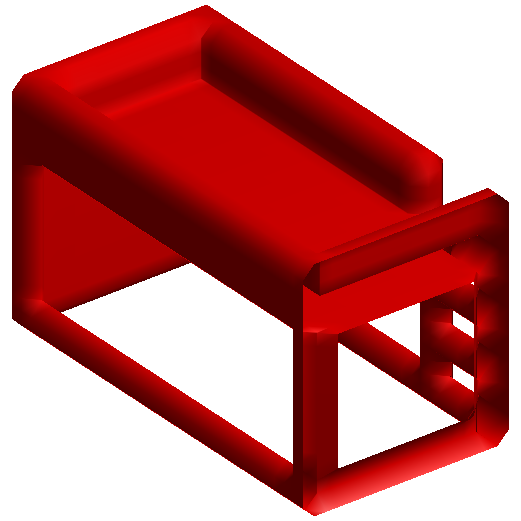}  \hspace{-1mm} 
    \includegraphics[height=.07\linewidth]{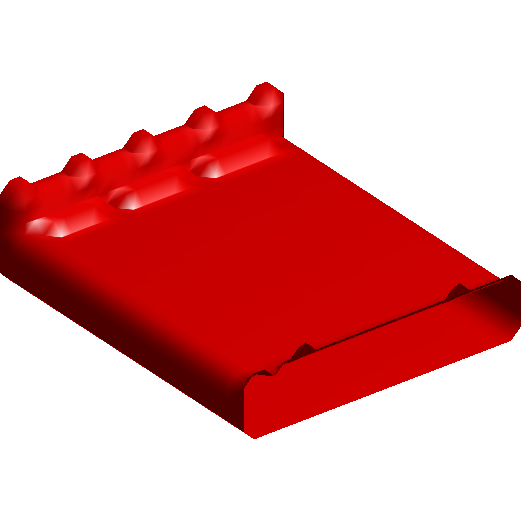} \hspace{-1mm}
    \includegraphics[height=.07\linewidth]{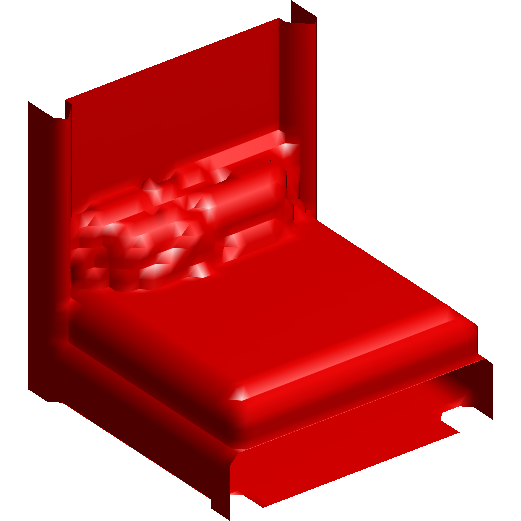}     
      \\
    \rotatebox[origin=l]{90}{\hspace{2mm} \textbf{{\footnotesize sofa}}}
    \includegraphics[height=.07\linewidth]{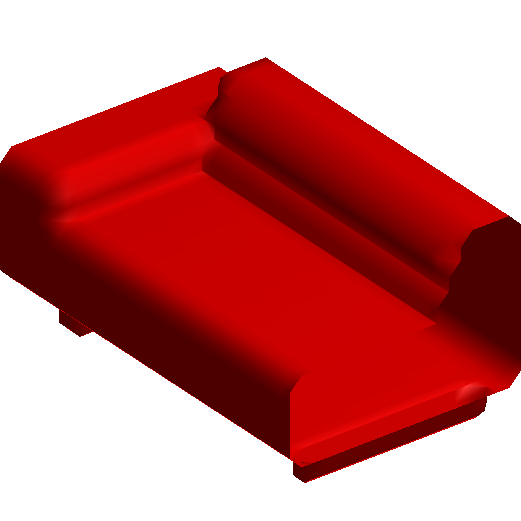}           \hspace{-1mm}
     \includegraphics[height=.07\linewidth]{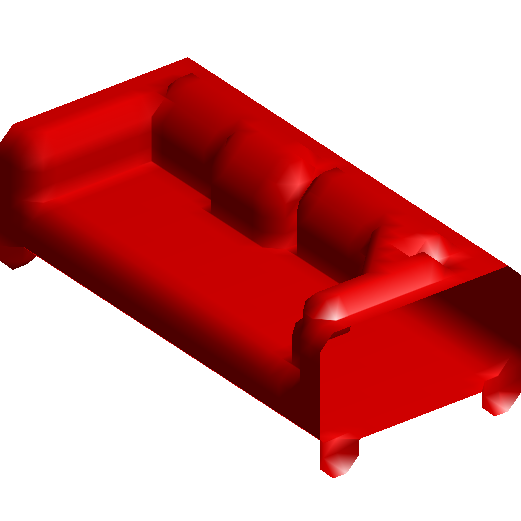}           \hspace{-1mm}
     \includegraphics[height=.07\linewidth]{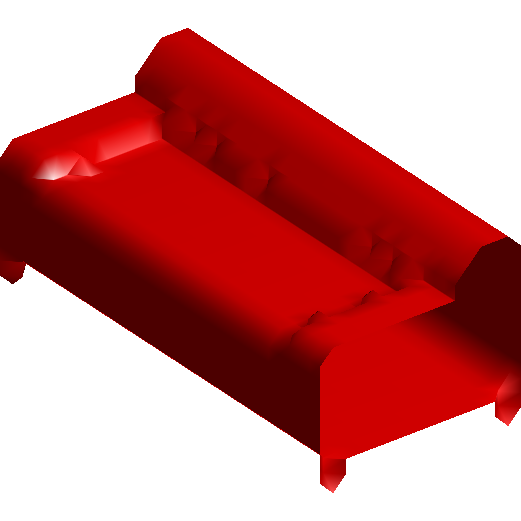}           
     \includegraphics[height=.07\linewidth]{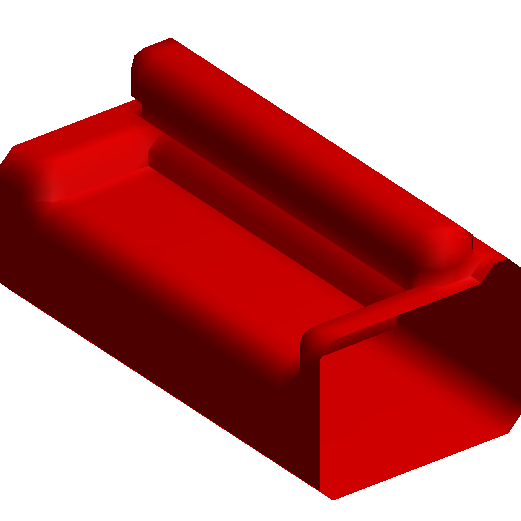}           \hspace{-1mm}  
      \includegraphics[height=.07\linewidth]{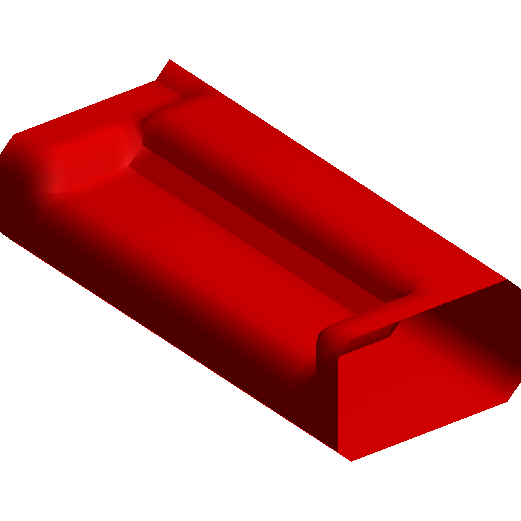}           \hspace{-1mm}    
      \includegraphics[height=.07\linewidth]{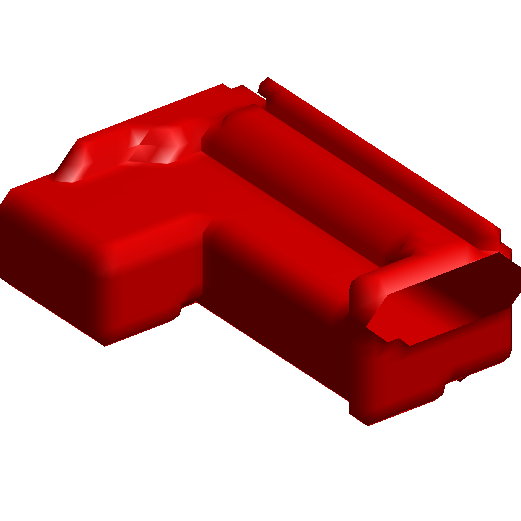}           \hspace{-1mm}
      \includegraphics[height=.07\linewidth]{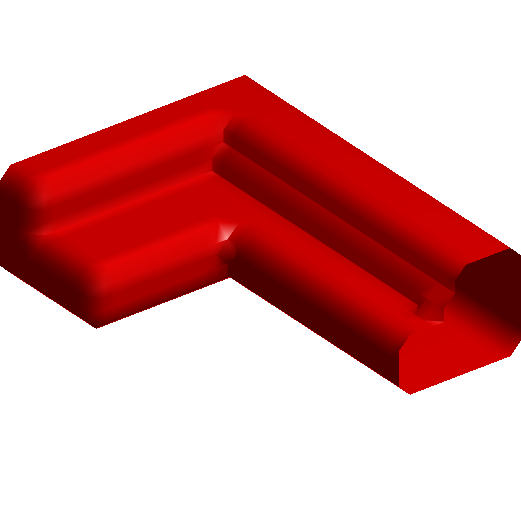}           \hspace{-1mm}     
      \includegraphics[height=.07\linewidth]{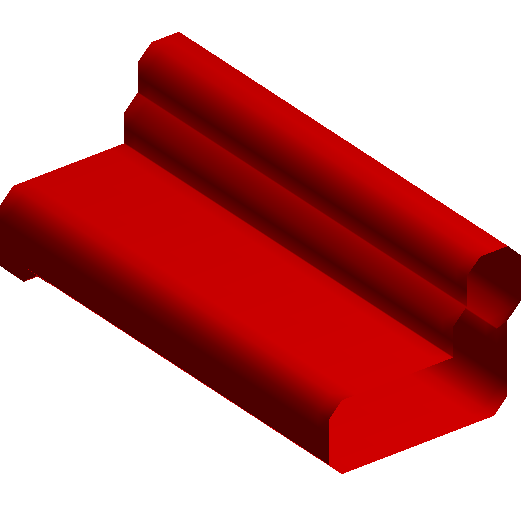}           \hspace{-1mm} 
      \includegraphics[height=.07\linewidth]{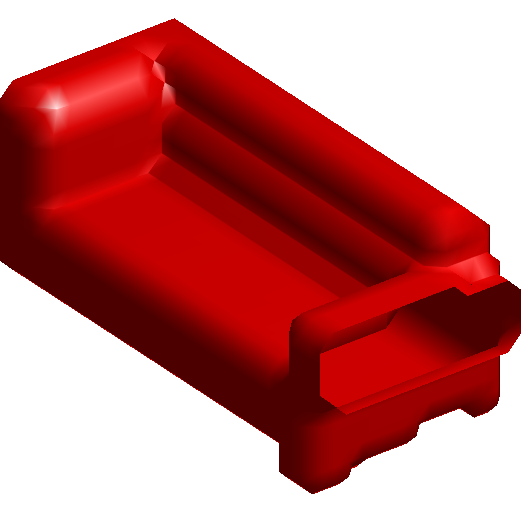}           
      \includegraphics[height=.07\linewidth]{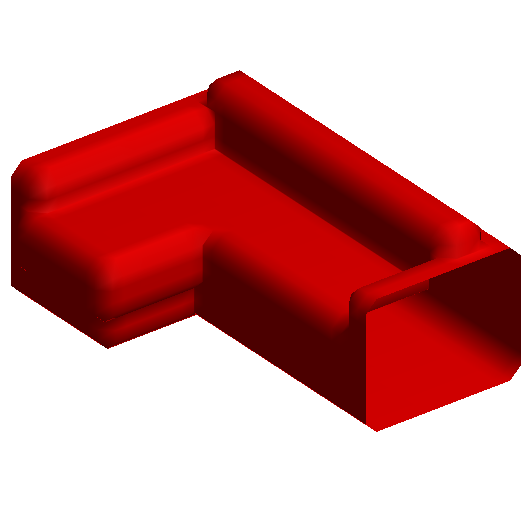}           \hspace{-1mm}
      \includegraphics[height=.07\linewidth]{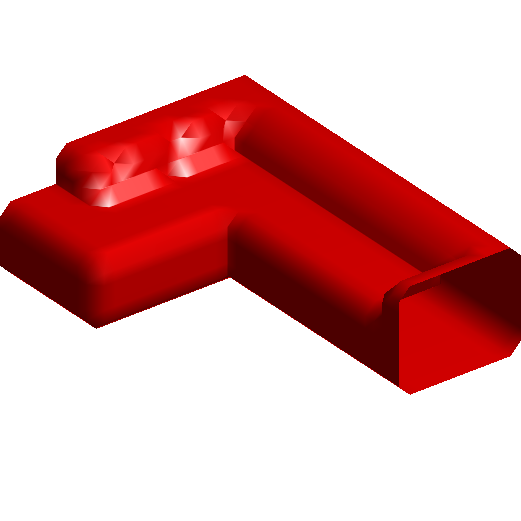}           \hspace{-1mm}     
      \includegraphics[height=.07\linewidth]{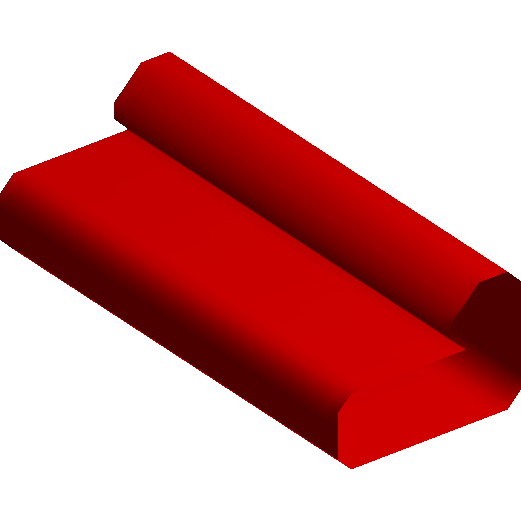}           \hspace{-1mm} 
      \includegraphics[height=.07\linewidth]{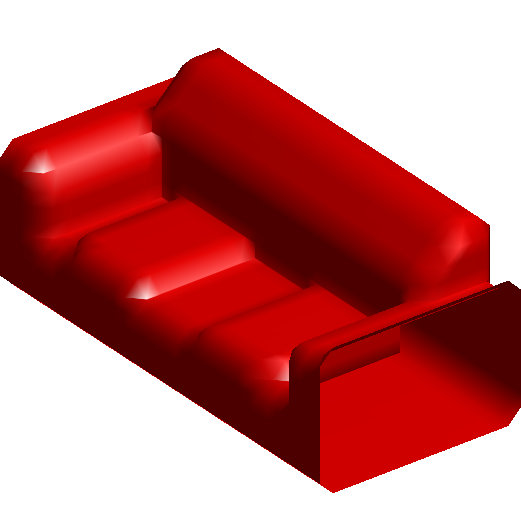}           \hspace{-1mm}     
      \\
     \rotatebox[origin=l]{90}{\hspace{2mm} \textbf{{\footnotesize table}}}
      \includegraphics[height=.07\linewidth]{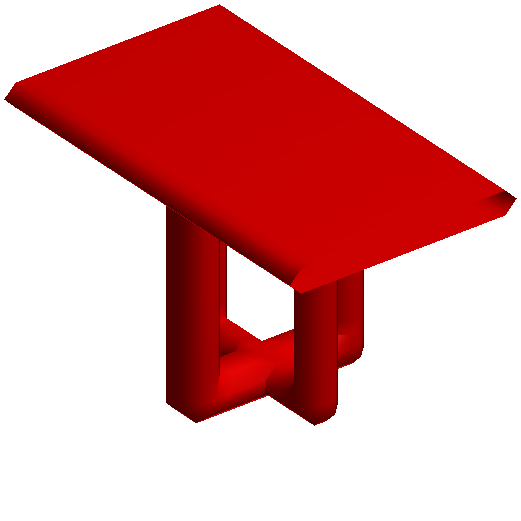}           \hspace{-1mm} 
      \includegraphics[height=.07\linewidth]{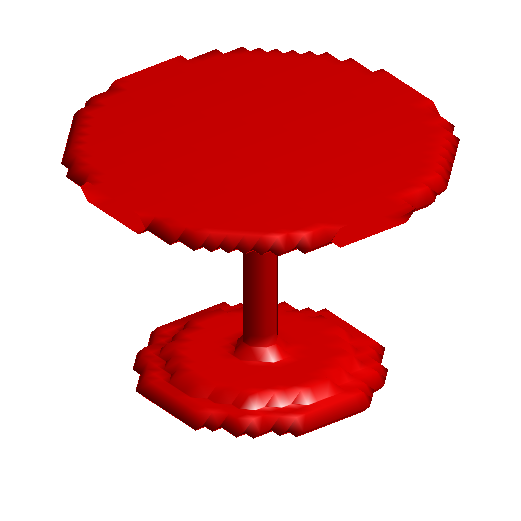}           \hspace{-1mm} 
       \includegraphics[height=.07\linewidth]{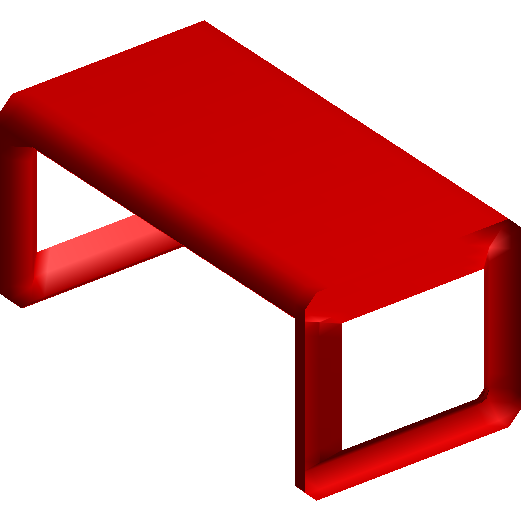}           
      \includegraphics[height=.07\linewidth]{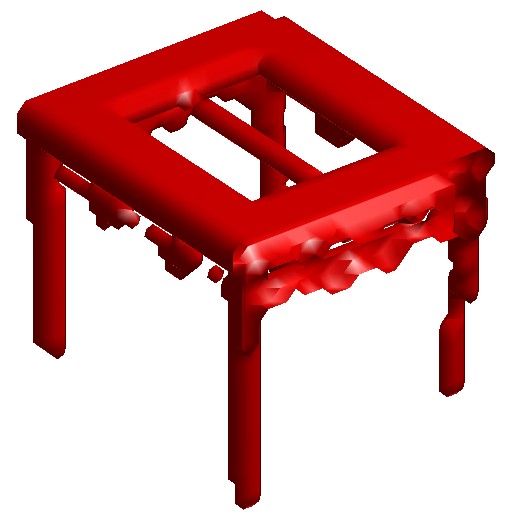}           \hspace{-1mm} 
     \includegraphics[height=.07\linewidth]{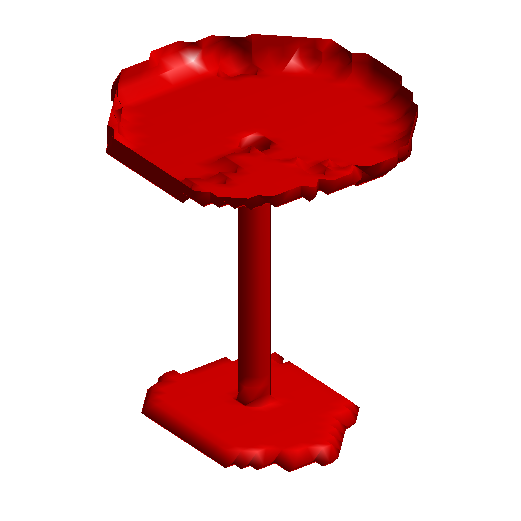}           \hspace{-1mm} 
     \includegraphics[height=.07\linewidth]{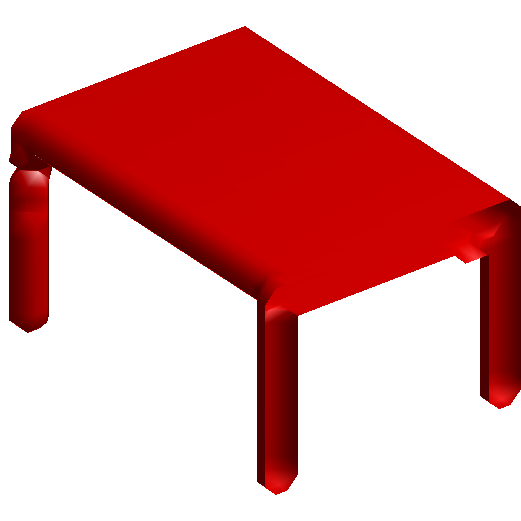}           \hspace{-1mm} 
      \includegraphics[height=.07\linewidth]{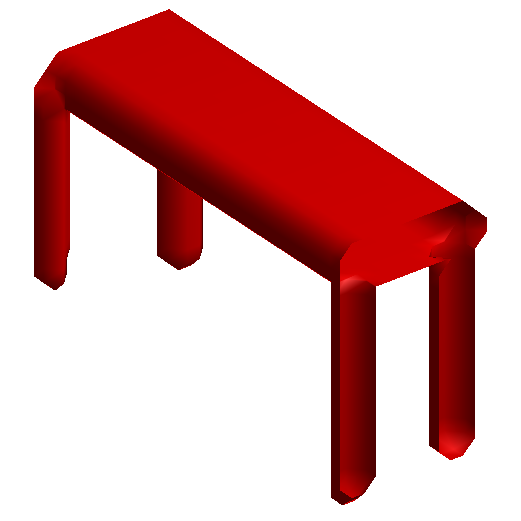}           \hspace{-1mm} 
      \includegraphics[height=.07\linewidth]{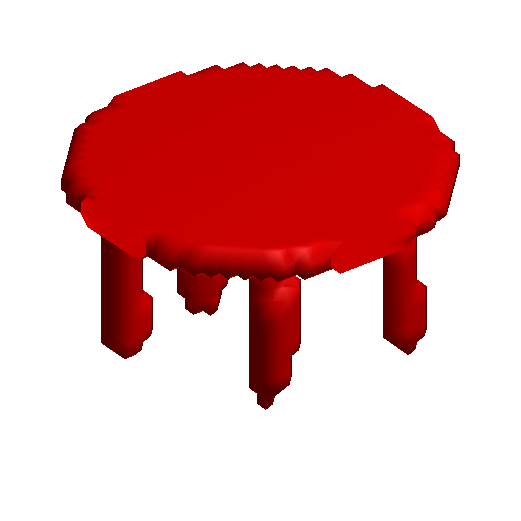}           \hspace{-1mm} 
     \includegraphics[height=.07\linewidth]{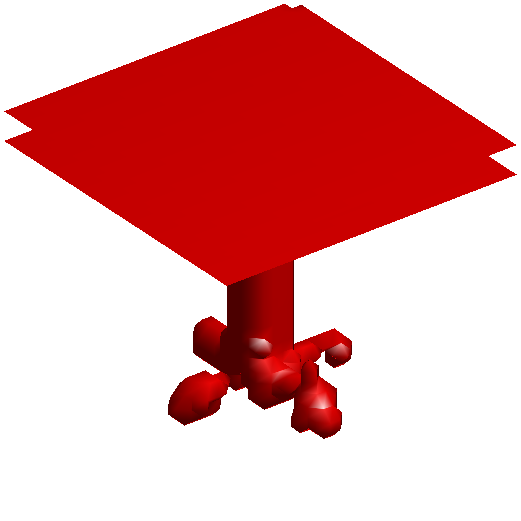}                
      \includegraphics[height=.07\linewidth]{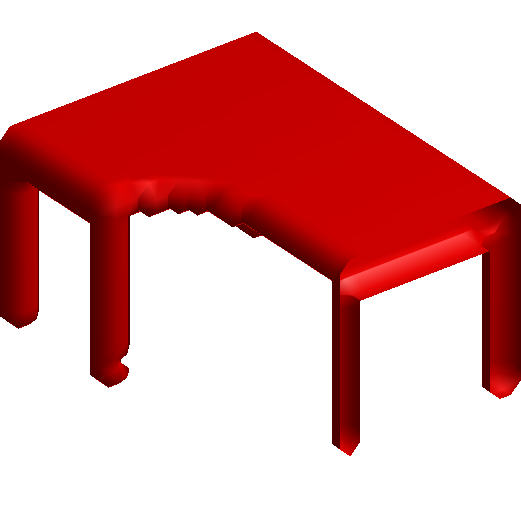}           \hspace{-1mm} 
      \includegraphics[height=.07\linewidth]{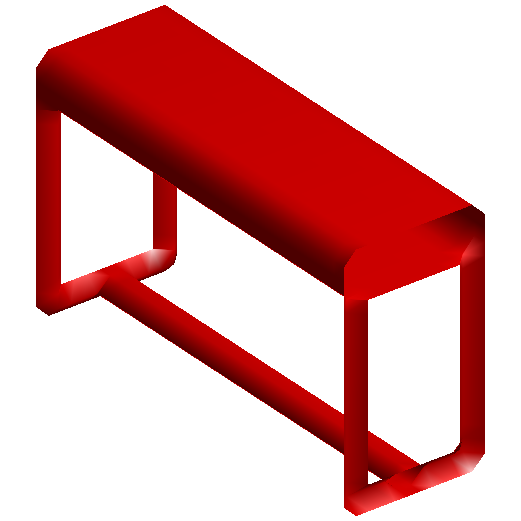}           \hspace{-1mm} 
     \includegraphics[height=.07\linewidth]{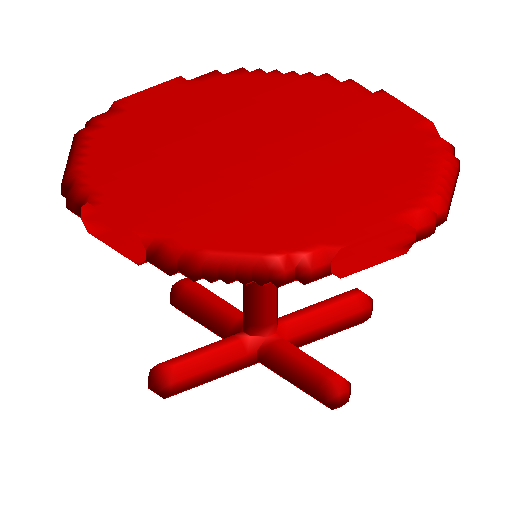}           \hspace{-1mm} 
    \includegraphics[height=.07\linewidth]{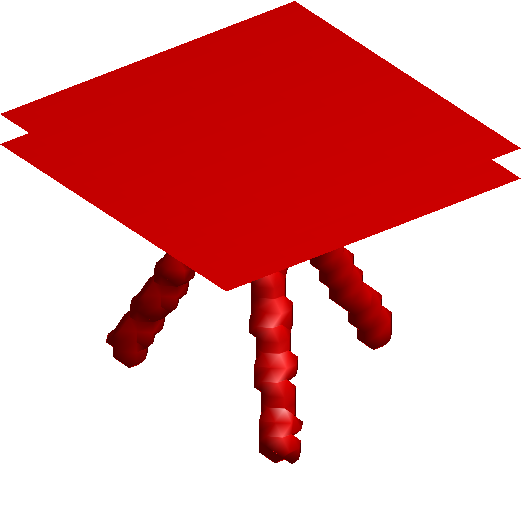}           \\
    \rotatebox[origin=l]{90}{\hspace{1mm}\textbf{{\footnotesize dresser}}}
    \includegraphics[height=.07\linewidth]{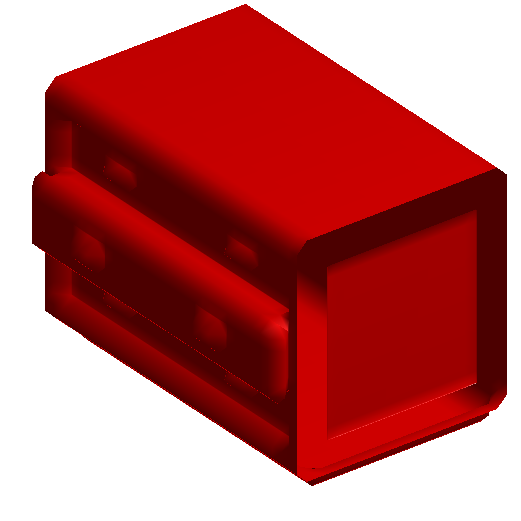}          \hspace{-1mm}  
     \includegraphics[height=.07\linewidth]{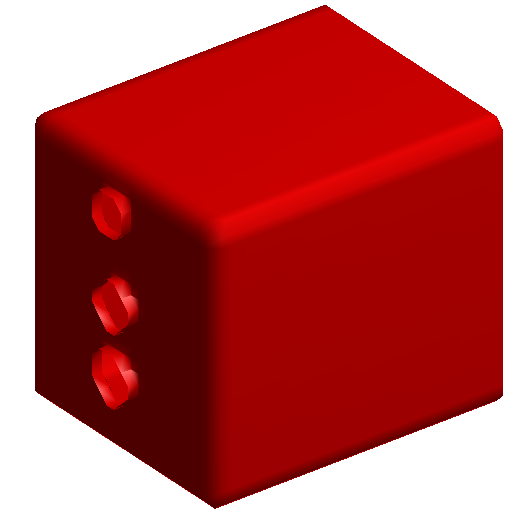}          \hspace{-1mm}  
      \includegraphics[height=.07\linewidth]{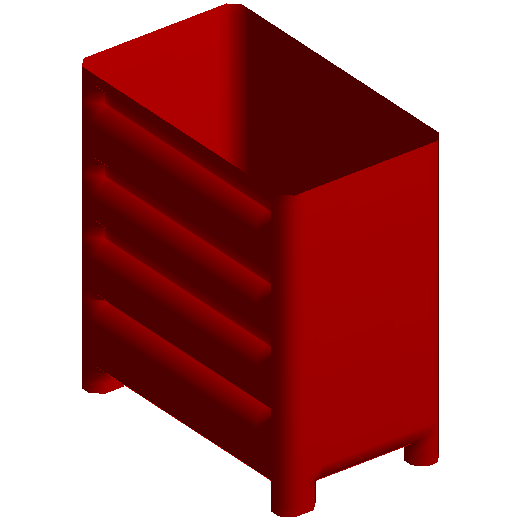}         
    \includegraphics[height=.07\linewidth]{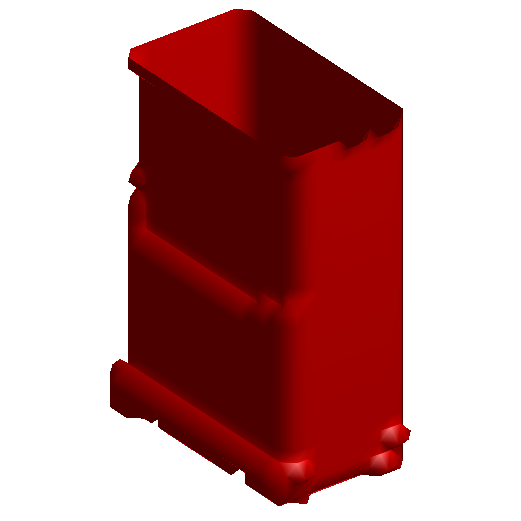}          \hspace{-1mm} 
     \includegraphics[height=.07\linewidth]{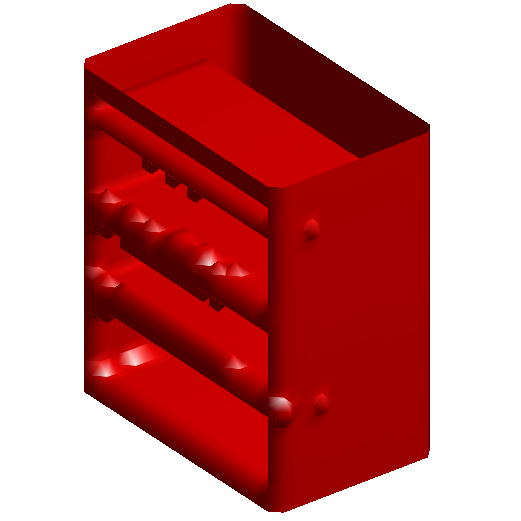}          \hspace{-1mm} 
     \includegraphics[height=.07\linewidth]{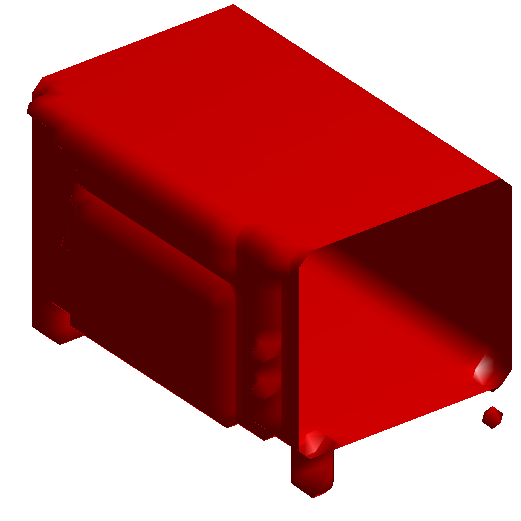}          \hspace{-1mm} 
     \includegraphics[height=.07\linewidth]{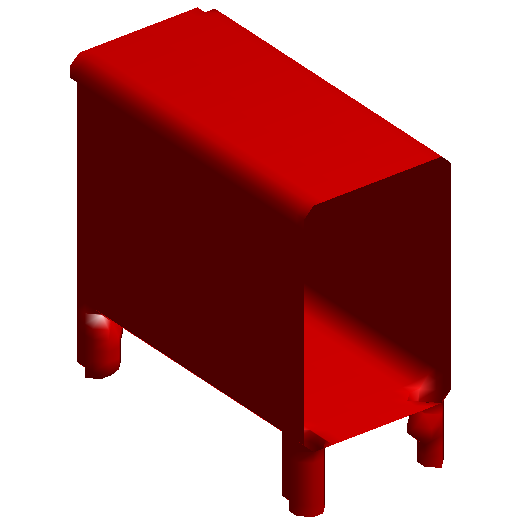}          \hspace{-1mm} 
     \includegraphics[height=.07\linewidth]{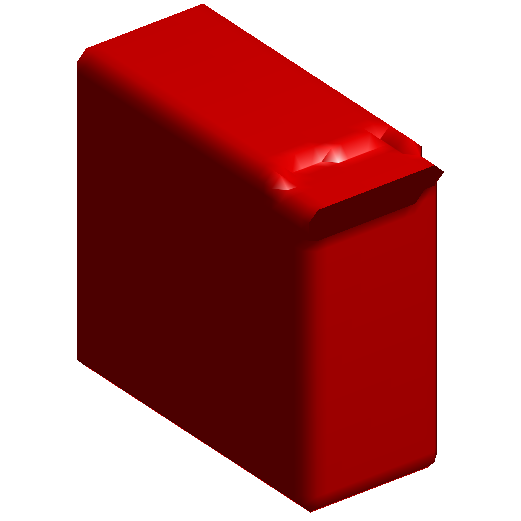}          \hspace{-1mm}  
     \includegraphics[height=.07\linewidth]{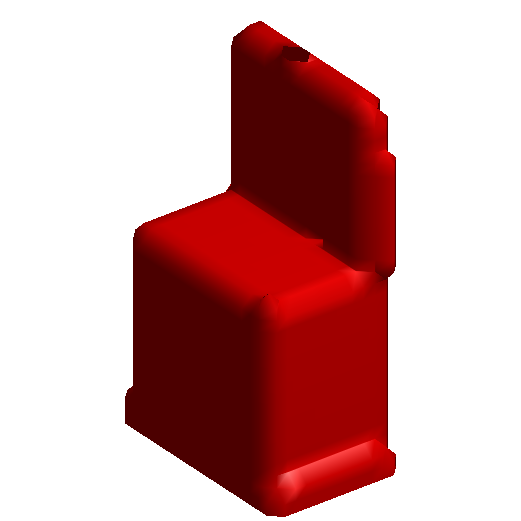}          
      \includegraphics[height=.07\linewidth]{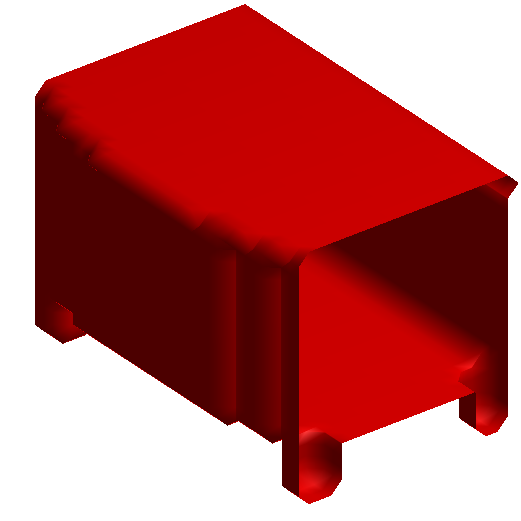}          \hspace{-1mm} 
     \includegraphics[height=.07\linewidth]{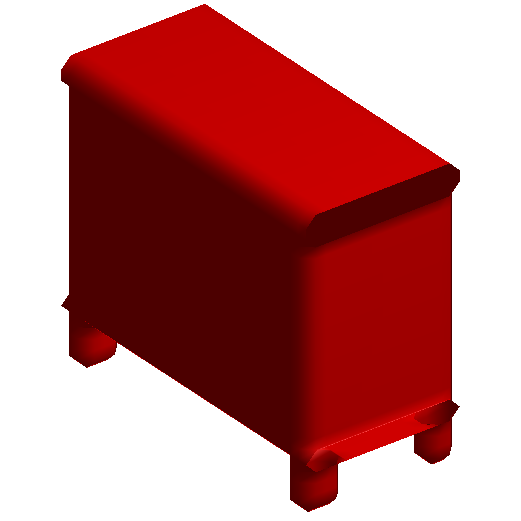}          \hspace{-1mm} 
     \includegraphics[height=.07\linewidth]{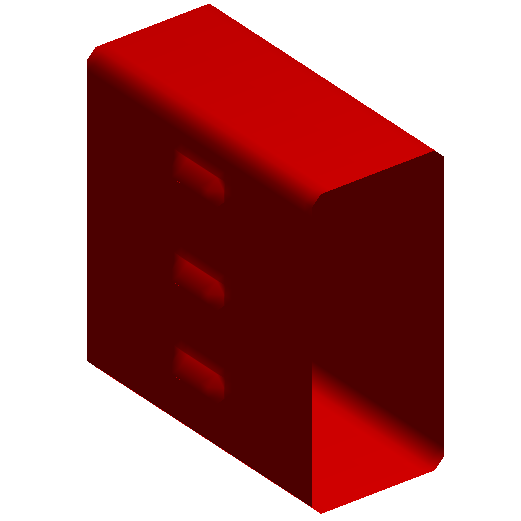}          \hspace{-1mm} 
     \includegraphics[height=.07\linewidth]{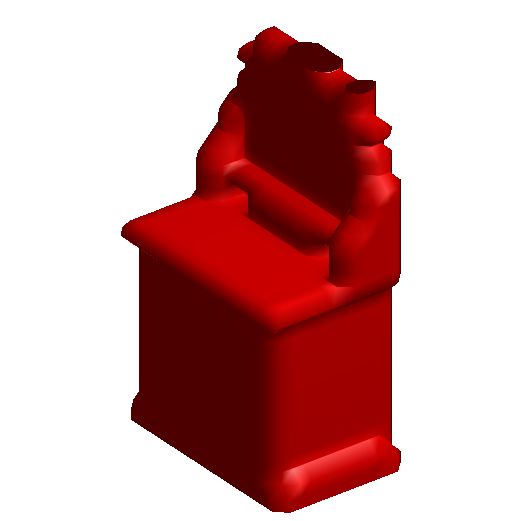}          \hspace{-1mm} \\
     \rotatebox[origin=l]{90}{\hspace{3mm}\textbf{{\footnotesize toilet}}}
        \includegraphics[height=.08\linewidth]{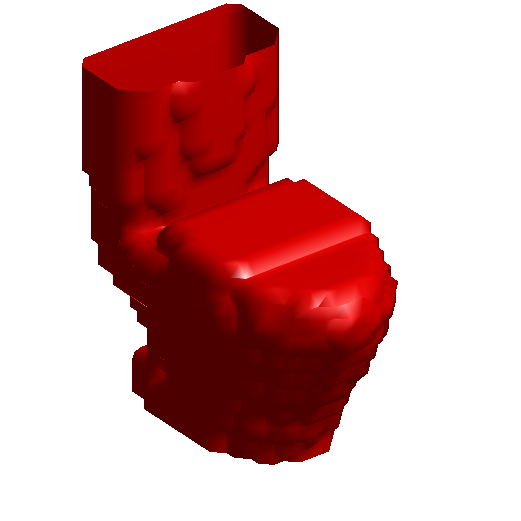} \hspace{-3.5mm}   
       \includegraphics[height=.08\linewidth]{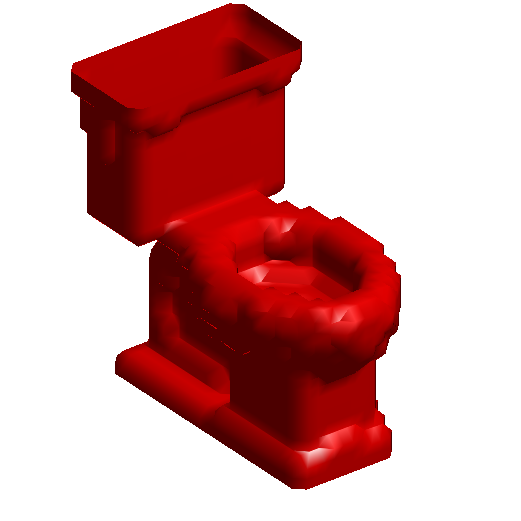}\hspace{-2.5mm}               
        \includegraphics[height=.08\linewidth]{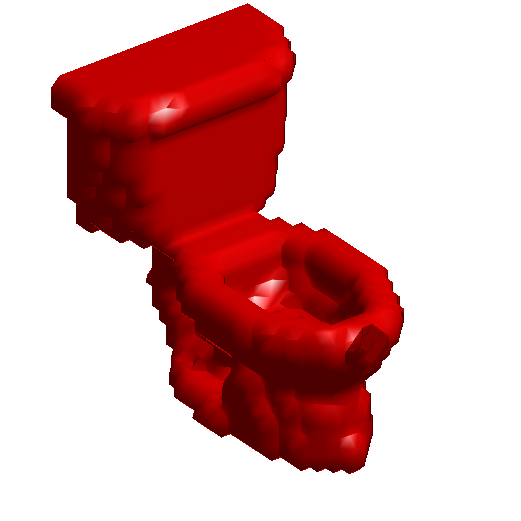}         \hspace{-2.5mm}  
       \includegraphics[height=.08\linewidth]{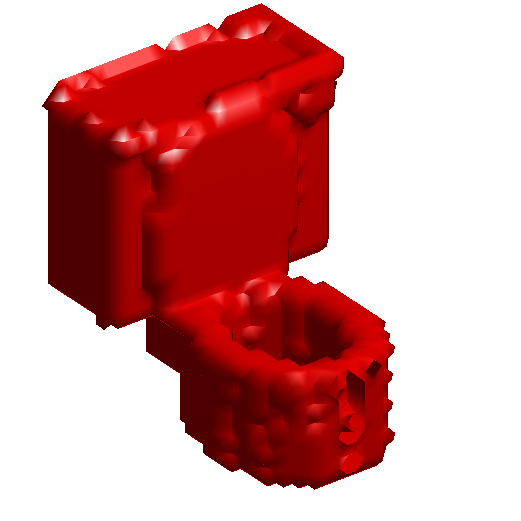}          \hspace{-2.5mm}   
     \includegraphics[height=.08\linewidth]{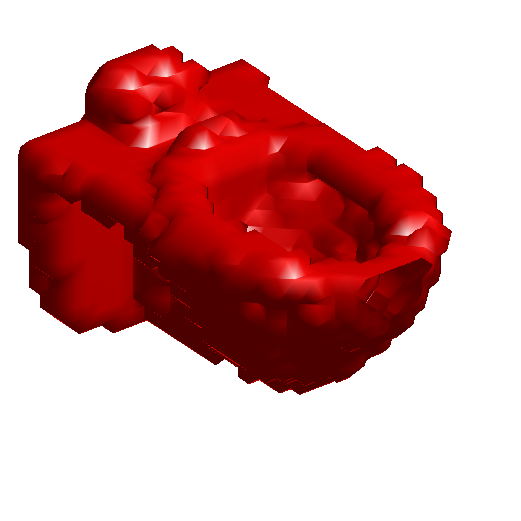}          \hspace{-2.5mm}   
      \includegraphics[height=.08\linewidth]{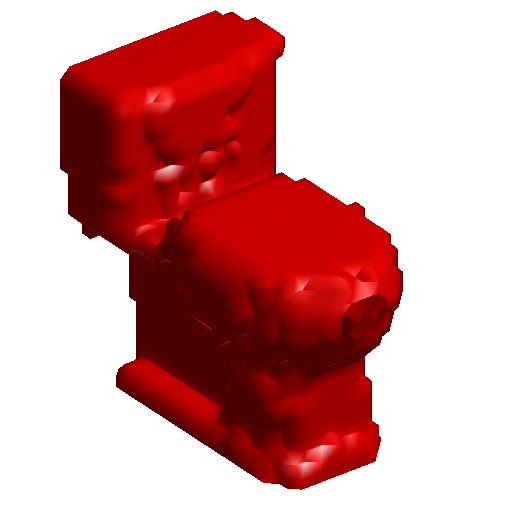}          \hspace{-2.5mm}   
     \includegraphics[height=.08\linewidth]{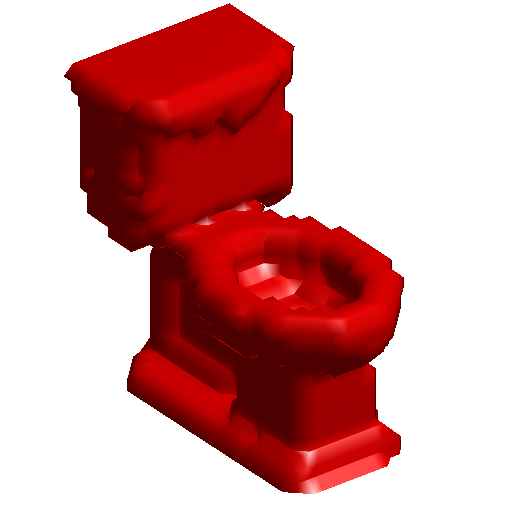}          \hspace{-3.5mm}   
     \includegraphics[height=.08\linewidth]{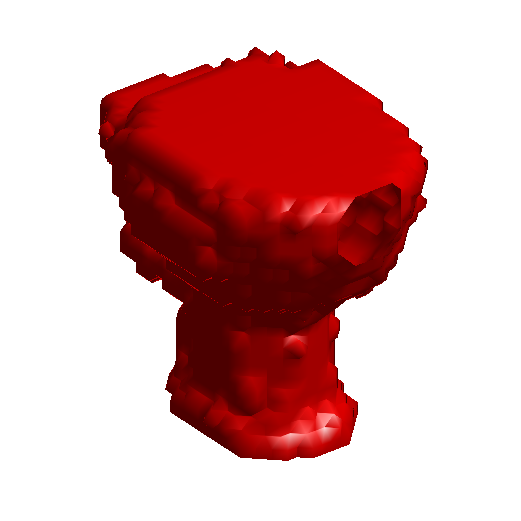}          \hspace{-2.5mm}   
     \includegraphics[height=.08\linewidth]{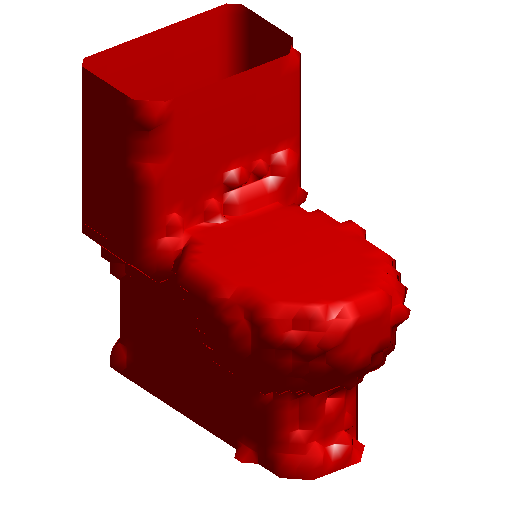}        \hspace{-2.5mm}        
     \includegraphics[height=.08\linewidth]{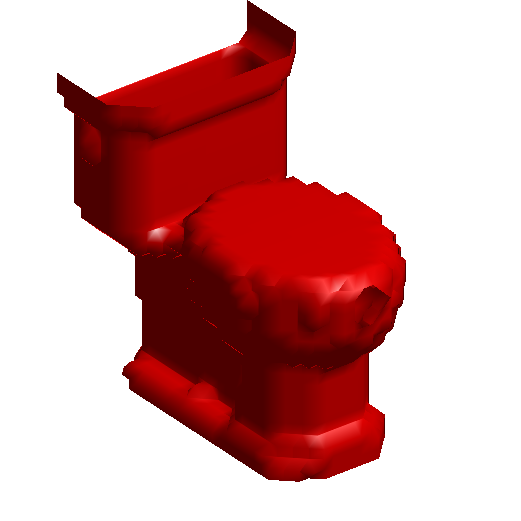}          \hspace{-3mm}   
     \includegraphics[height=.08\linewidth]{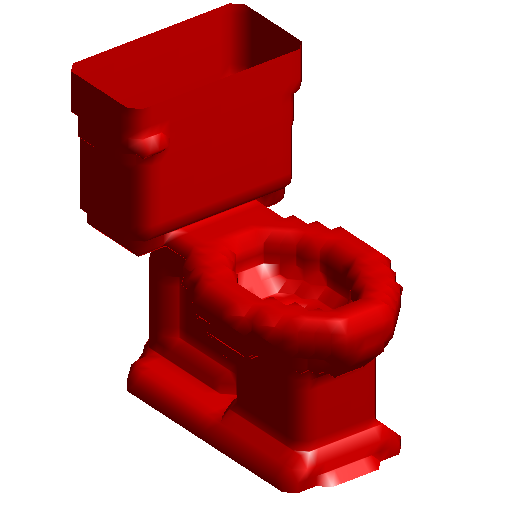}          \hspace{-3mm}   
     \includegraphics[height=.08\linewidth]{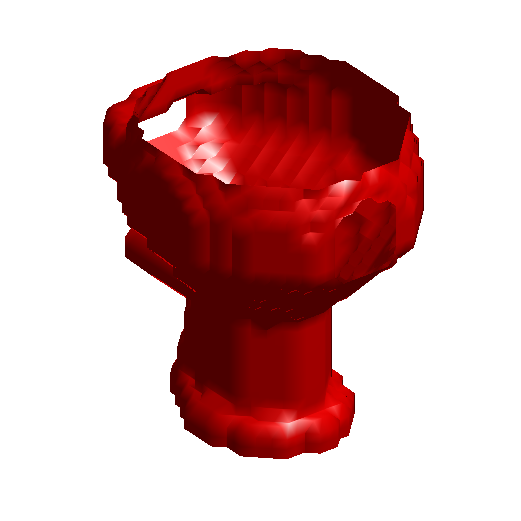}          \hspace{-2.5mm}  
     \includegraphics[height=.08\linewidth]{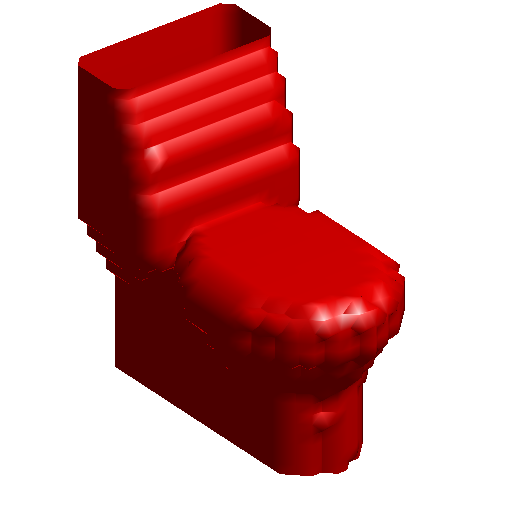}          \\
	\caption{Generating 3D objects. Each row displays one experiment, where the first three 3D objects are some observed examples, columns 4, 5, 6, 7, 8, and 9 are 6 of the synthesized 3D objects sampled from the learned model by Langevin dynamics. For the last four synthesized objects (shown in columns 6, 7, 8, and 9), their nearest neighbors retrieved from the training set are shown in columns 10, 11, 12, and 13. }	
	\label{exp:synthesis}
\end{figure*}

We conduct experiments on synthesizing 3D objects of categories from ModelNet dataset \cite{wu20153d}. Specifically, we use ModelNet10, a 10-category subset of ModelNet which is commonly used as benchmark for 3D object analysis. The categories are chair, sofa, bathtub, toilet, bed, desk, table, nightstand, dresser, and monitor. The size of the training set for each category ranges from 100 to 700.

For qualitative experiment, we learn one 3-layer 3D DescriptorNet for each object category in ModelNet10. The first layer has 200 $16\times16\times16$ filters with sub-sampling of 3, the second layer has 100 $6\times6\times6$ filters with sub-sampling of $2$, and the final layer is a fully connected layer with a single filter that covers the whole voxel grid. We add ReLU layers between convolutional layers. We fix the standard deviation of the reference distribution of the model to be $s=0.5$. The number of Langevin dynamics steps in each learning iteration is $l$=20 and the step size $\Delta \tau =0.1$. We use Adam \cite{kingma2015adam} for optimization with $\beta_1=0.5$ and $\beta_2=0.999$. The learning rate is 0.001. The number of learning iterations is $3,000$. We disable the noise term in the Langevin step after $100$ iterations. The training data are of size $32 \times 32 \times 32$ voxels, whose values are 0 or 1. We prepare the training data by subtracting the mean value from the data. Each voxel value of the synthesized data is discretized into 0 or 1 by comparing with a threshold 0.5. The mini-batch size is 20. The number of parallel sampling chains for each batch is 25.

Figure \ref{exp:synthesis} displays the observed 3D objects randomly sampled from the training set, and the synthesized 3D objects generated by our models for categories chair, bed, sofa, table, dresser, and toilet. 
We visualize volumetric data via isosurfaces in our paper. To show that our model can synthesize new 3D objects beyond the training set, we compare the synthesized patterns with their nearest neighbors in the training set. The retrieved nearest neighbors are based on $\ell_2$ distance in the voxel space. As shown in Figure \ref{exp:synthesis}, our model can synthesize realistic 3D shapes, and the generated 3D objects are similar, but not identical, to the training set.  

To quantitatively evaluate our model, we adopt the Inception score proposed by \cite{warde2016improving}, which uses a reference convolutional neural network to compute  
\begin{eqnarray} 
I(\{\tY_i,i=1,...,\tilde{n}\})=\exp\left( \E_{\tY}\left[ \KL( p(c|\tY)\parallel p(c))\right]\right), \nonumber 
\end{eqnarray} 
where $c$ denotes category, $\{\tY_i,i=1,...,\tilde{n}\}$ are synthesized examples sampled from the model, $p(c|\tY)$ is obtained from the output of the reference network, and $p(c)
\approx \frac{1}{\tilde{n}} \sum_{i=1}^{\tilde{n}} p(c|\tY_i)$. Both a low entropy conditional category distribution $p(c|\tY)$ (i.e., the  network classifies a given sample with high certainty) and a high entropy category distribution $p(c)$ (i.e., the network identifies a wide variety of categories among the generated samples) can lead to a high inception score. In our experiment, we use a state-of-the-art 3D multi-view convolutional neural network \cite{qi2016volumetric} trained on ModelNet dataset for 3D object classification as the reference network.

We learn a single model from mixed 3D objects from the training sets of 10 3D object categories of ModelNet10 dataset.
Table \ref{tb:inception} reports the Inception scores of our model as well as a comparison with some baseline models including 3D-GAN \cite{3dgan}, 3D ShapeNets \cite{wu20153d}, and 3D-VAE \cite{kingma2013auto}.

We also evaluate the quality of the synthesized 3D shapes by the model learned from single category by using average softmax class probability that reference network assigns to the synthesized examples for the underlying category. 
Table \ref{synthesisSingleExp} displays the results for all 10 categories. It can be seen that our model generates 3D shapes with higher softmax class probabilities than other baseline models.


\begin{table}
\centering
\begin{small}
\caption{Inception scores of different methods of learning from 10 3D object categories.}\label{tb:inception}
\vspace{-2mm}
\begin{tabular}{|l|r|}
\hline 
Method &  Inception score \\ \hline \hline
3D ShapeNets \cite{wu20153d} & 4.126$\pm$0.193    \\ \hline
3D-GAN \cite{3dgan}& 8.658$\pm$0.450     \\ \hline
3D VAE \cite{kingma2013auto} & 11.015$\pm$0.420\\ \hline
3D DescriptorNet (ours) &  \textbf{11.772$\pm$0.418} \\ \hline
\end{tabular}
\end{small}
\end{table}
\vspace{-2mm}

\begin{table}[h]
\caption{Softmax class probability}\label{synthesisSingleExp}
\vspace{-2mm}
\centering
\begin{small}
\begin{tabular}{|c|c|c|c|c|}
\hline
category & ours   & \cite{3dgan}   & \cite{kingma2013auto}  &  \cite{wu20153d}  \\ \hline \hline

bathtub    & \textbf{0.8348} & 0.7017 & 0.7190 & 0.1644 \\ \hline

bed     & \textbf{0.9202} & 0.7775 & 0.3963 & 0.3239 \\ \hline

chair   & \textbf{0.9920} & 0.9700  & 0.9892 & 0.8482 \\ \hline

desk      & \textbf{0.8203} & 0.7936 & 0.8145 & 0.1068  \\ \hline

dresser      & \textbf{0.7678} & 0.6314 & 0.7010 & 0.2166  \\ \hline

monitor & \textbf{0.9473} & 0.2493   & 0.8559 & 0.2767\\ \hline

night stand    & \textbf{0.7195} & 0.6853 & 0.6592 & 0.4969\\ \hline

sofa   & \textbf{0.9480} &  0.9276  & 0.3017 & 0.4888\\ \hline  

table    & \textbf{0.8910} & 0.8377 & 0.8751 &0.7902 \\ \hline
toilet   & \textbf{0.9701} &  0.8569  & 0.6943 & 0.8832\\ \hline \hline

Avg.   & \textbf{0.8811} &  0.7431   & 0.7006 & 0.4596\\ \hline 

\end{tabular}
%
%
\end{small}
\end{table}
\vspace{-2mm}

\subsection{3D object recovery}
\label{Exp:objectRecovery}
We then test the conditional 3D DescriptorNet on the 3D object recovery task.  On each testing 3D object, we randomly corrupt some voxels of the 3D object. We then seek to recover the corrupted voxels by sampling from the conditional distribution $p(Y_M|Y_{\tilde{M}}; \theta)$ according to the learned model $p(Y; \theta)$, where $M$ and $\tilde{M}$ denote the corrupted and uncorrupted voxels, and $Y_M$ and $Y_{\tilde{M}}$ are the corrupted part and the uncorrupted part of the 3D object $Y$ respectively. The sampling of $p(Y_M|Y_{\tilde{M}}; \theta)$ is again accomplished by the Langevin dynamics, which is the same as the Langevin dynamics that samples from the full distribution $p(Y; \theta)$, except that we fix the uncorrupted part $Y_{\tilde{M}}$ and only update the corrupted part $Y_M$ throughout the Langevin dynamics. In the learning stage, we learn the model from the fully observed training 3D objects. To specialize the learned model to this recovery task, we learn the conditional distribution $p(Y_M|Y_{\tilde{M}}; \theta)$ directly. That is, in the learning stage, we also randomly corrupt each fully observed training 3D object $Y$, and run Langevin dynamics by fixing $Y_M$ to obtain the synthesized 3D object. The parameters $\theta$ are then updated by gradient ascent according to (\ref{eq:lD2}). The network architecture for recovery is the same as the one used in Section \ref{Exp:objectSynthesis} for synthesis. The number of Langevin dynamics steps for recovery in each iteration is set to be $l=90$ and the step size is $\Delta \tau =0.07$. The number of learning iterations is $1,000$. The size of the mini-batch is 50. The 3D training data are of size $32 \times 32 \times 32$ voxels. 

\begin{figure}
	\centering
	\rotatebox[origin=l]{90}{\hspace{1mm}\textbf{{\scriptsize original}}}
	\includegraphics[height=.139\linewidth]{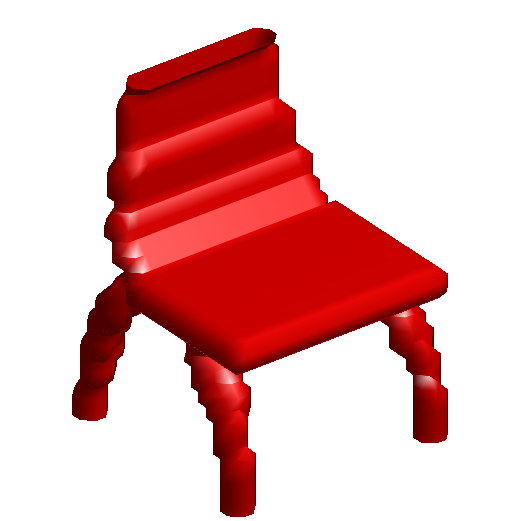} \hspace{-1.8mm}
	\includegraphics[height=.139\linewidth]{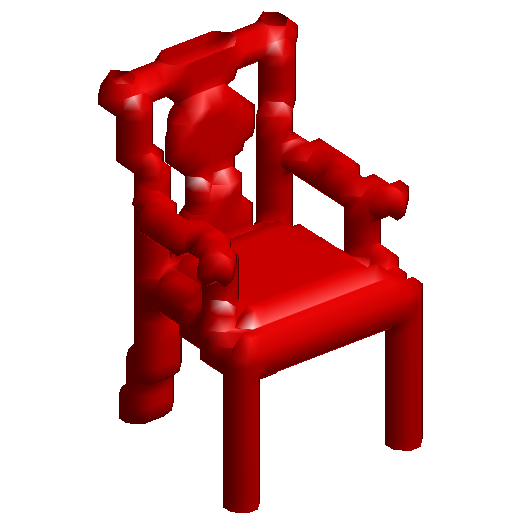} \hspace{-1.8mm}
    \includegraphics[height=.139\linewidth]{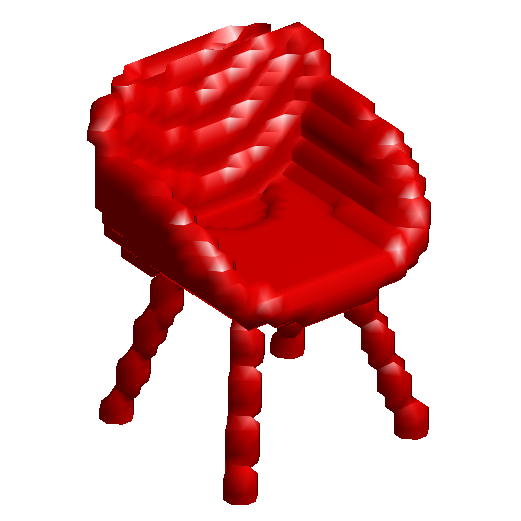} \hspace{-1.8mm}
    \includegraphics[height=.139\linewidth]{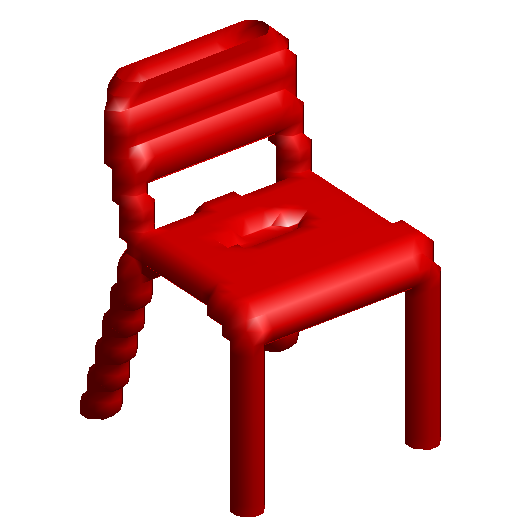} \hspace{-1.8mm}
    \includegraphics[height=.139\linewidth]{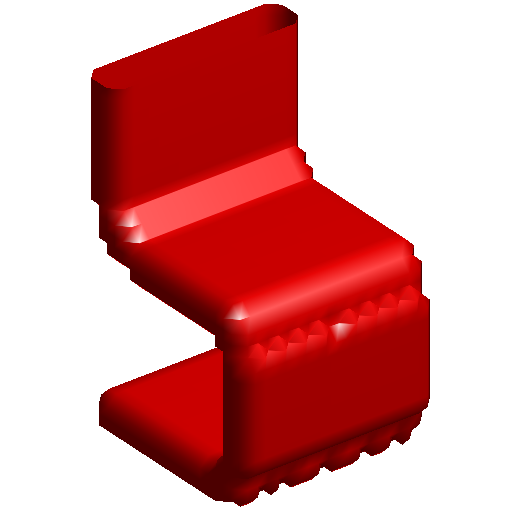} \hspace{-1.8mm}
    \includegraphics[height=.139\linewidth]{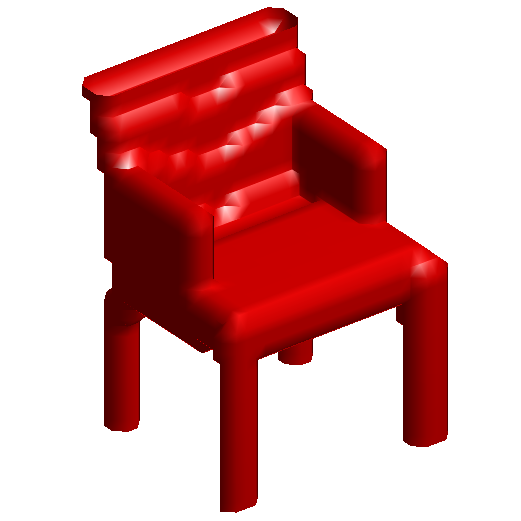} \hspace{-1.8mm}    
    \includegraphics[height=.139\linewidth]{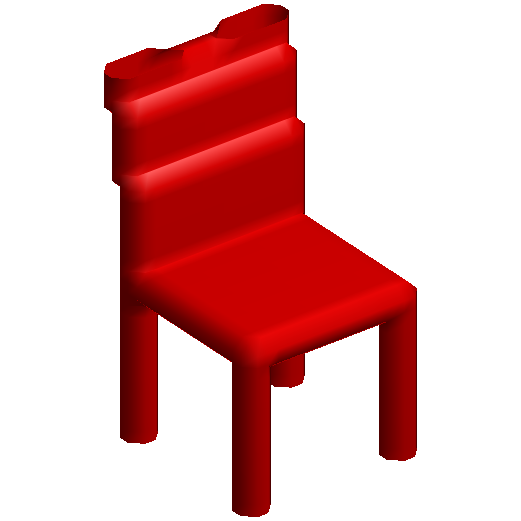}   
     \\
    \rotatebox[origin=l]{90}{\hspace{1mm}\textbf{{\scriptsize corrupted}}}
	\includegraphics[height=.139\linewidth]{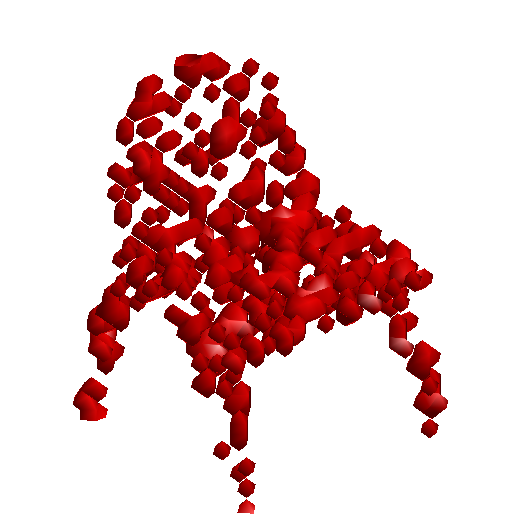} \hspace{-1.8mm}
	\includegraphics[height=.139\linewidth]{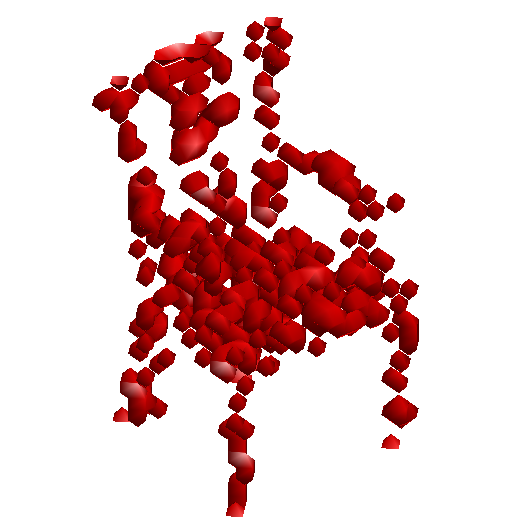} \hspace{-1.8mm}
    \includegraphics[height=.139\linewidth]{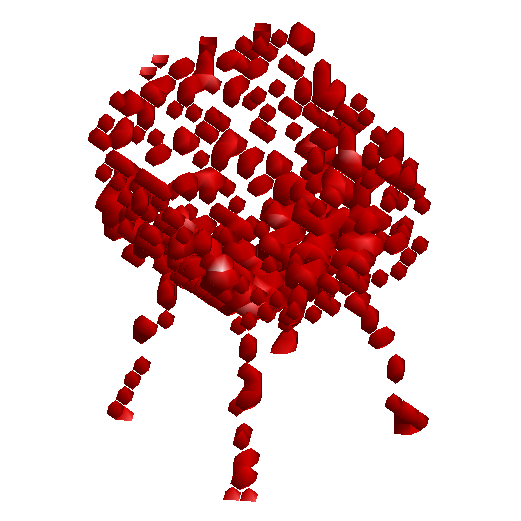} \hspace{-1.8mm}
    \includegraphics[height=.139\linewidth]{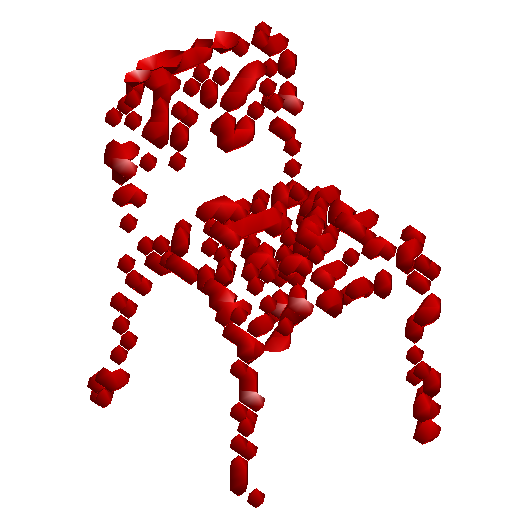} \hspace{-1.8mm}
    \includegraphics[height=.139\linewidth]{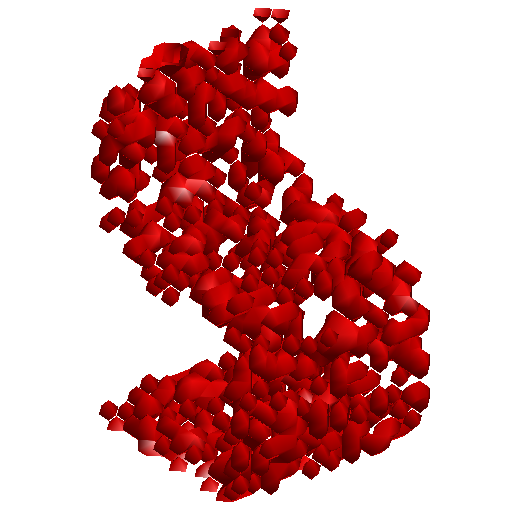} \hspace{-1.8mm}
    \includegraphics[height=.139\linewidth]{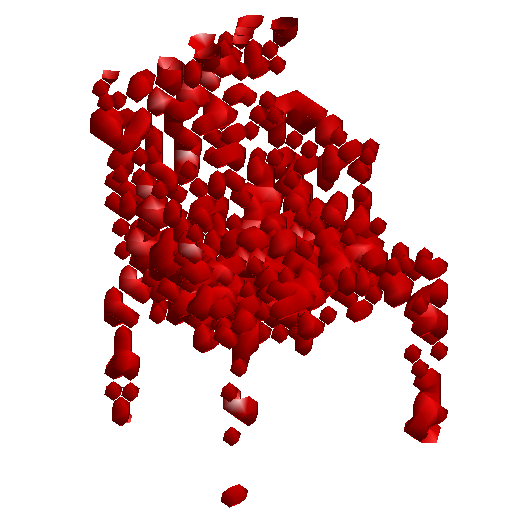} \hspace{-1.8mm}   
    \includegraphics[height=.139\linewidth]{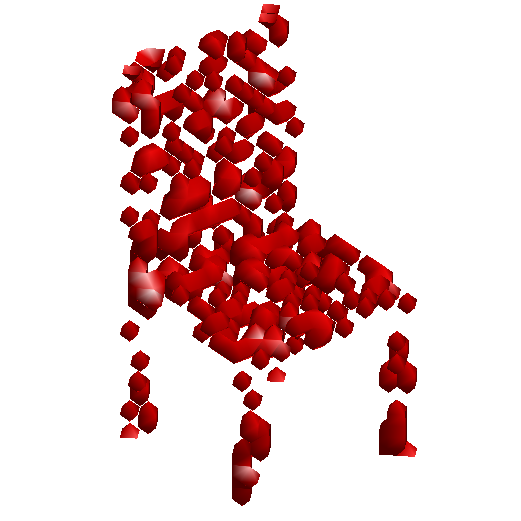} 
     \\
     \rotatebox[origin=l]{90}{\hspace{1mm}\textbf{{\scriptsize recovered}}}
     \includegraphics[height=.139\linewidth]{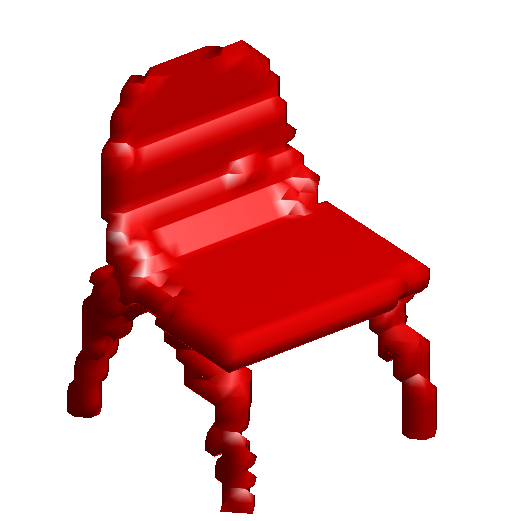} \hspace{-1.8mm}
     \includegraphics[height=.139\linewidth]{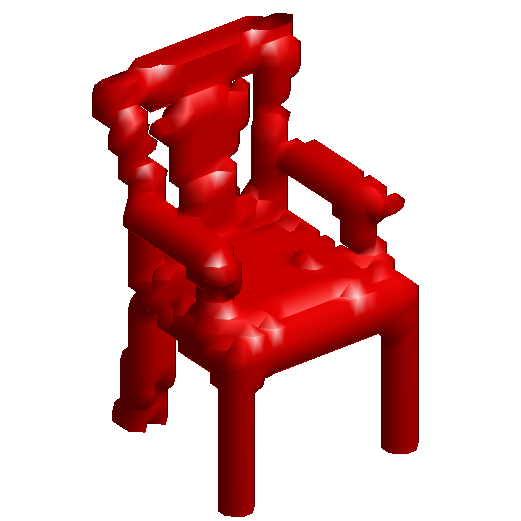} \hspace{-1.8mm}
    \includegraphics[height=.139\linewidth]{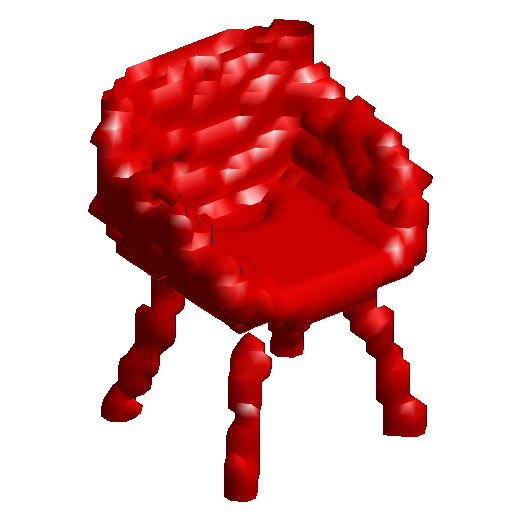} \hspace{-1.8mm}
    \includegraphics[height=.139\linewidth]{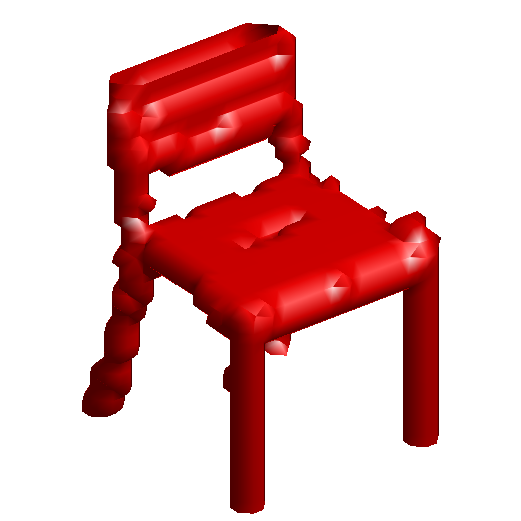} \hspace{-1.8mm}
    \includegraphics[height=.139\linewidth]{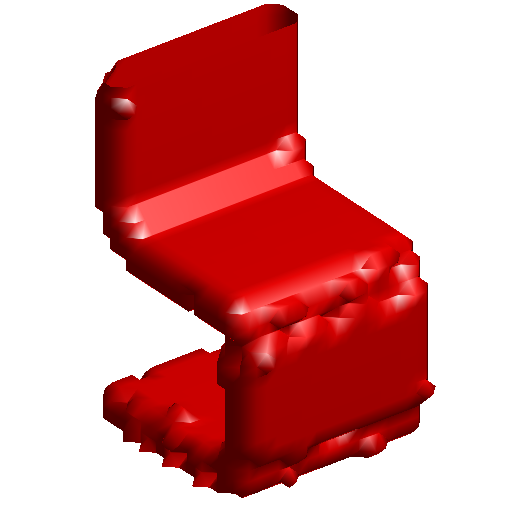} \hspace{-1.8mm}
    \includegraphics[height=.139\linewidth]{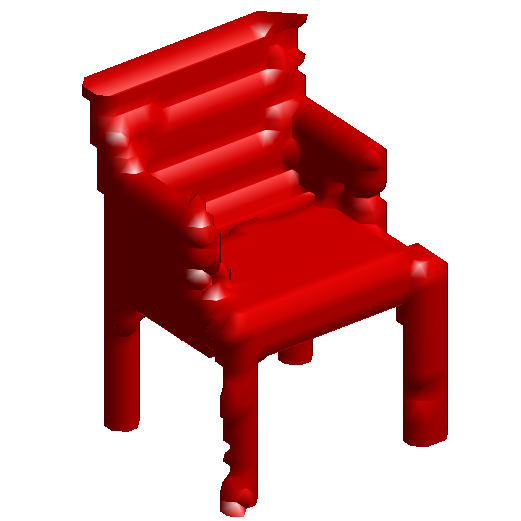} \hspace{-1.8mm}
    \includegraphics[height=.139\linewidth]{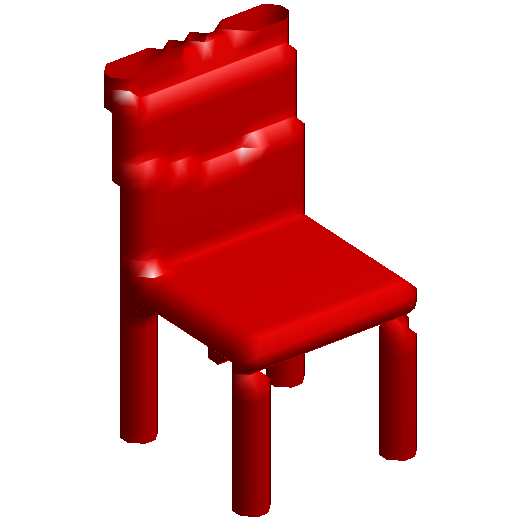}\\
    (a) chair\\
    \vspace{2mm}
    \rotatebox[origin=l]{90}{\hspace{1mm}\textbf{{\scriptsize original}}}
    \includegraphics[height=.139\linewidth]{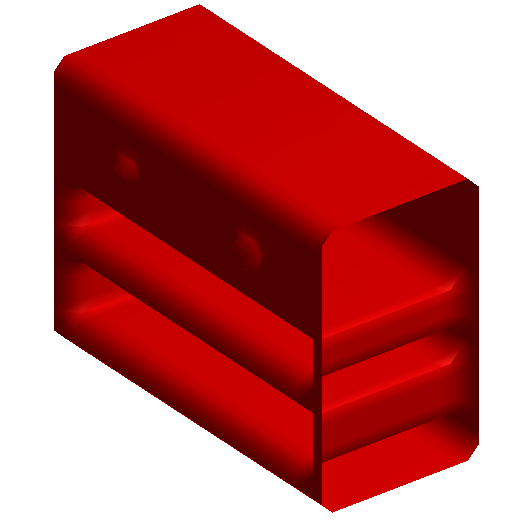} \hspace{-1.8mm}
    \includegraphics[height=.139\linewidth]{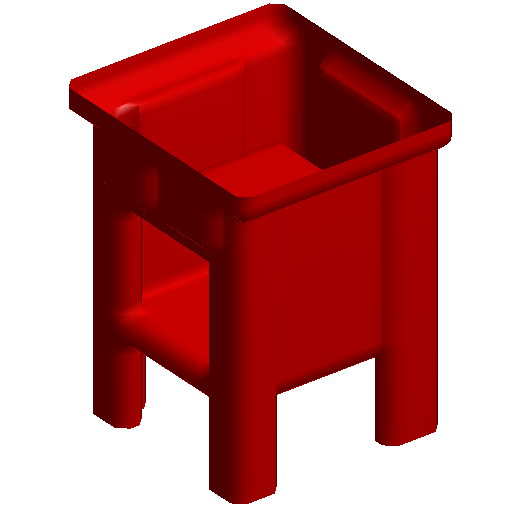} \hspace{-1.8mm}
    \includegraphics[height=.139\linewidth]{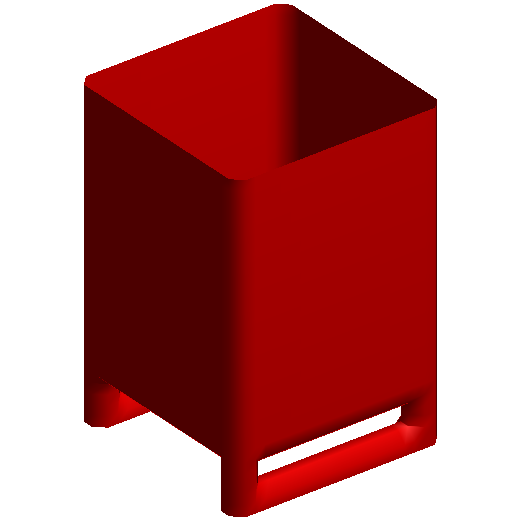} \hspace{-1.8mm}
    \includegraphics[height=.139\linewidth]{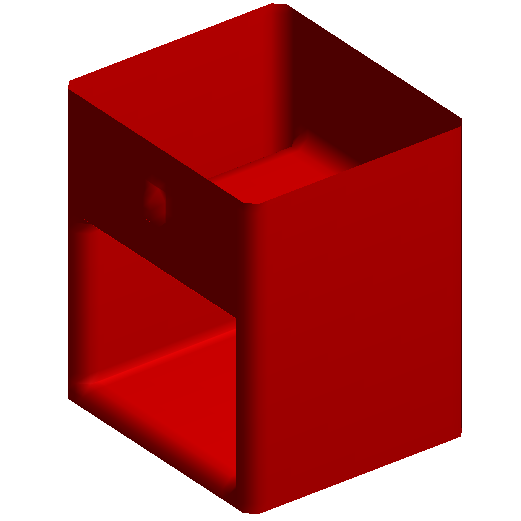} \hspace{-1.8mm}
    \includegraphics[height=.139\linewidth]{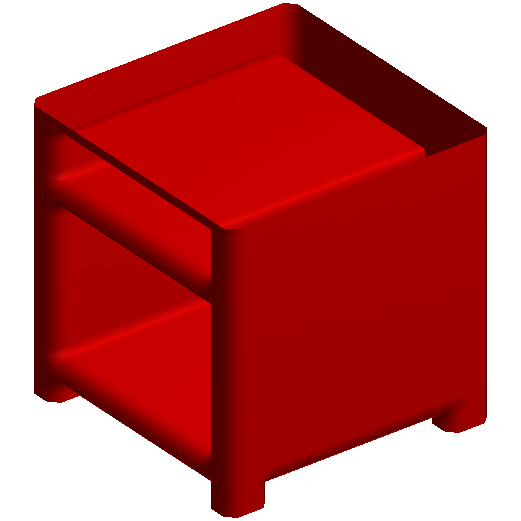} \hspace{-1.8mm}
    \includegraphics[height=.139\linewidth]{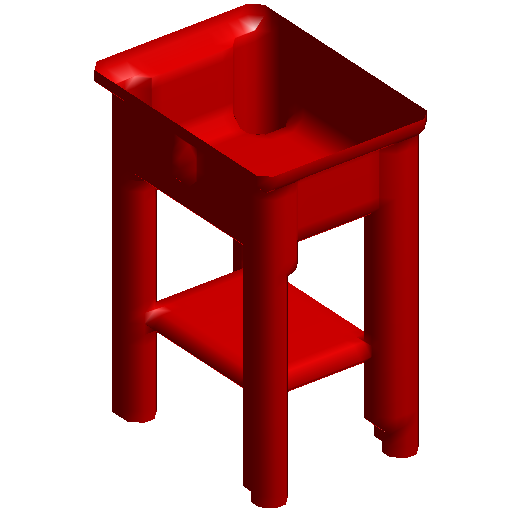} \hspace{-1.8mm}
    \includegraphics[height=.139\linewidth]{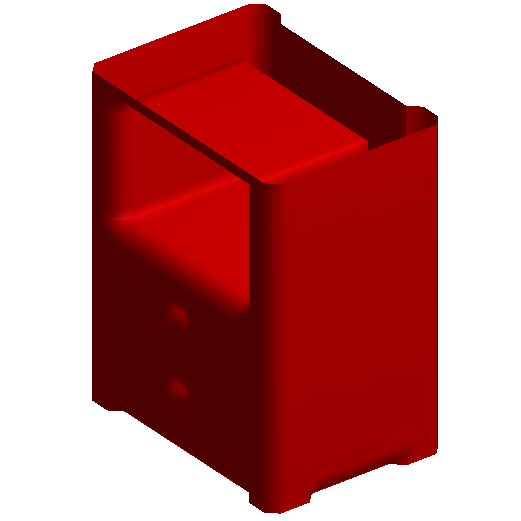} \hspace{-1.8mm}\\
    \rotatebox[origin=l]{90}{\hspace{1mm}\textbf{{\scriptsize corrupted}}}
    \includegraphics[height=.139\linewidth]{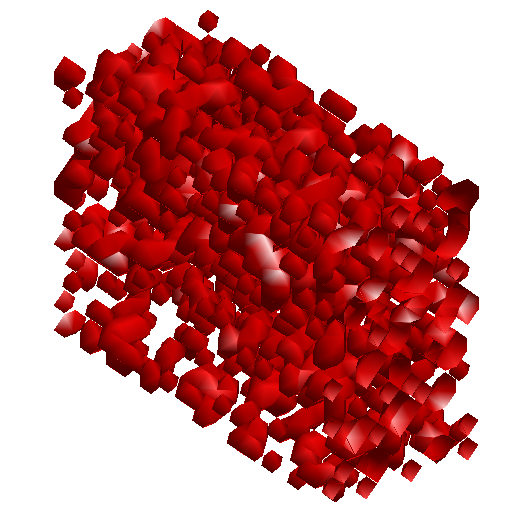} \hspace{-1.8mm}
    \includegraphics[height=.139\linewidth]{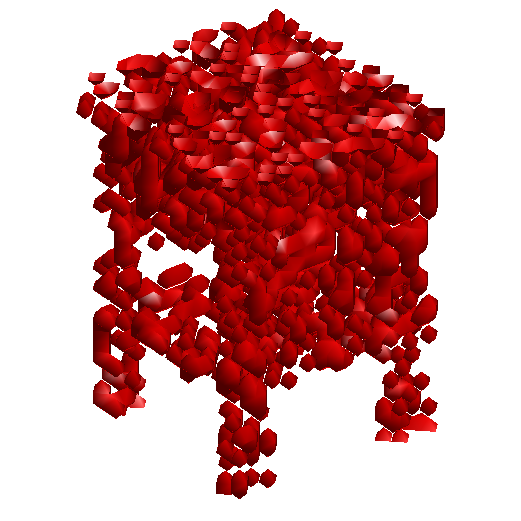} \hspace{-1.8mm}
    \includegraphics[height=.139\linewidth]{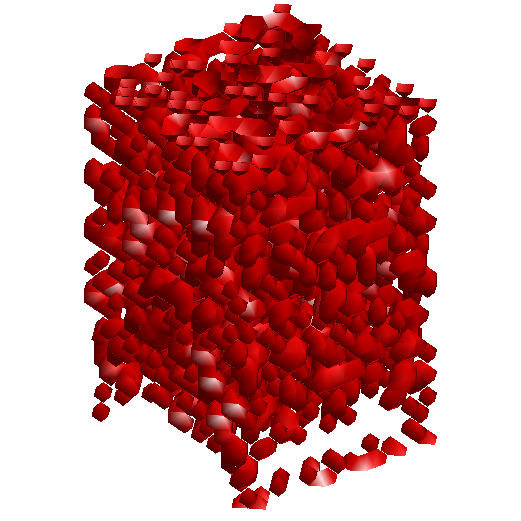} \hspace{-1.8mm}
    \includegraphics[height=.139\linewidth]{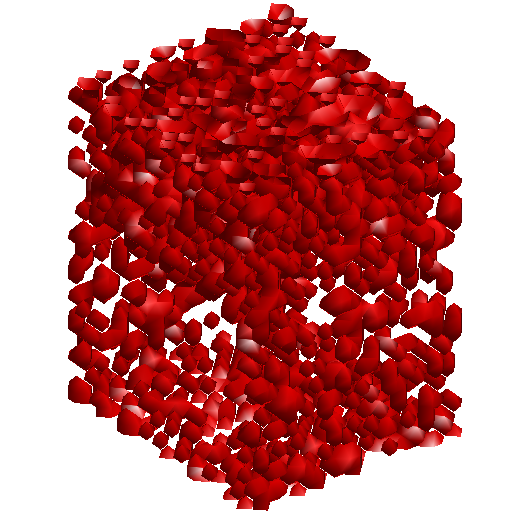} \hspace{-1.8mm}
    \includegraphics[height=.139\linewidth]{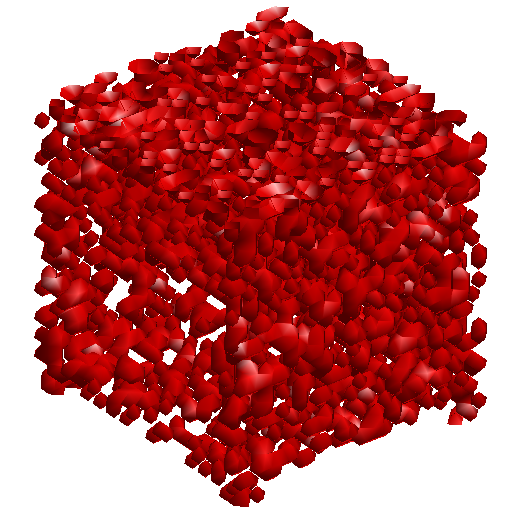} \hspace{-1.8mm}
    \includegraphics[height=.139\linewidth]{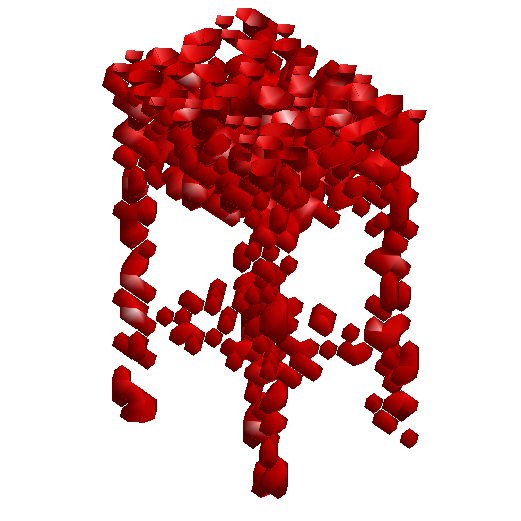} \hspace{-1.8mm}
    \includegraphics[height=.139\linewidth]{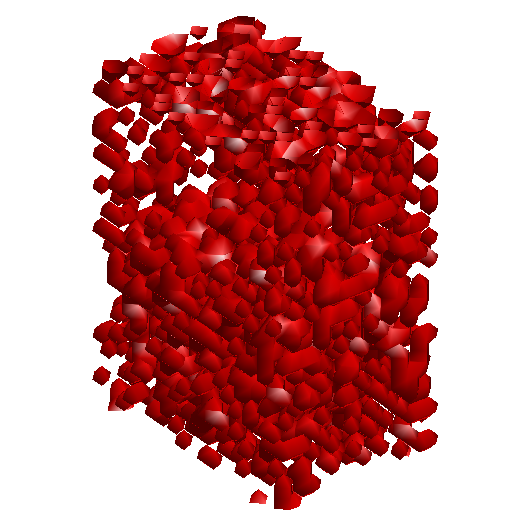} \hspace{-1.8mm}\\
    \rotatebox[origin=l]{90}{\hspace{1mm}\textbf{{\scriptsize recovered}}}
    \includegraphics[height=.139\linewidth]{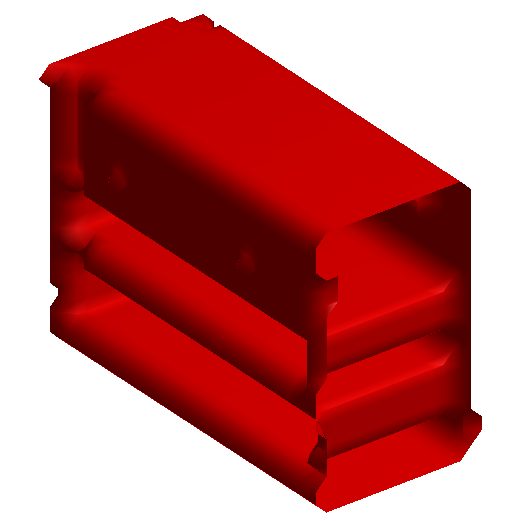} \hspace{-1.8mm}
    \includegraphics[height=.139\linewidth]{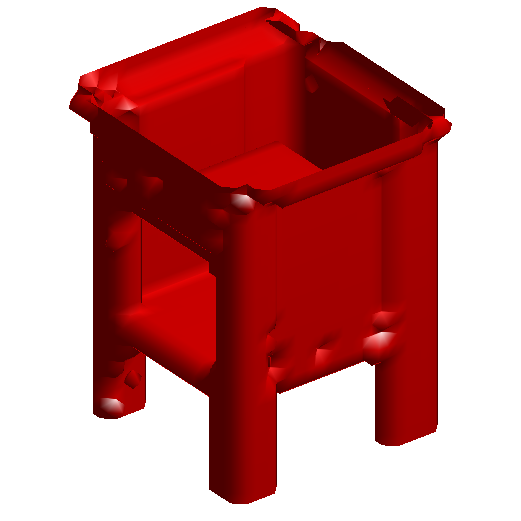} \hspace{-1.8mm}
    \includegraphics[height=.139\linewidth]{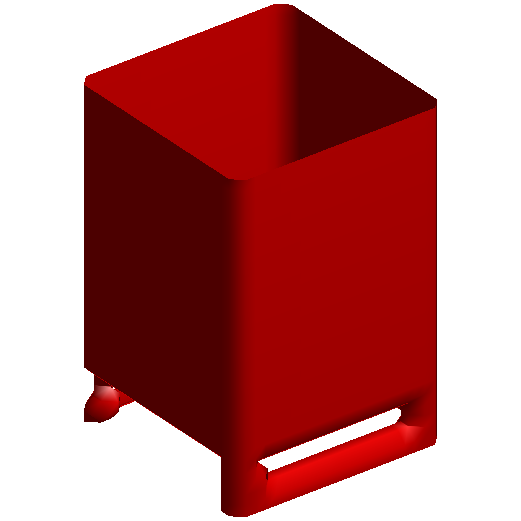} \hspace{-1.8mm}
    \includegraphics[height=.139\linewidth]{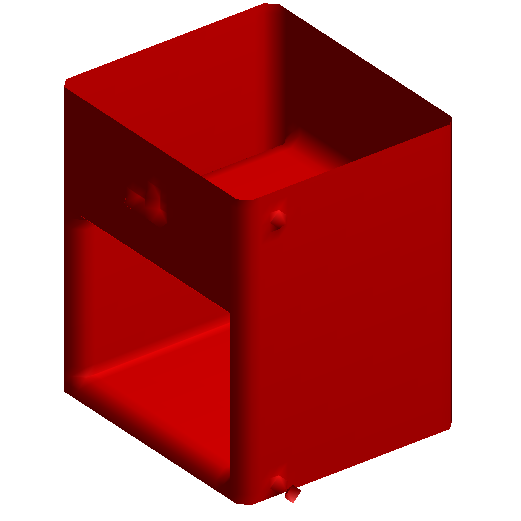} \hspace{-1.8mm}
    \includegraphics[height=.139\linewidth]{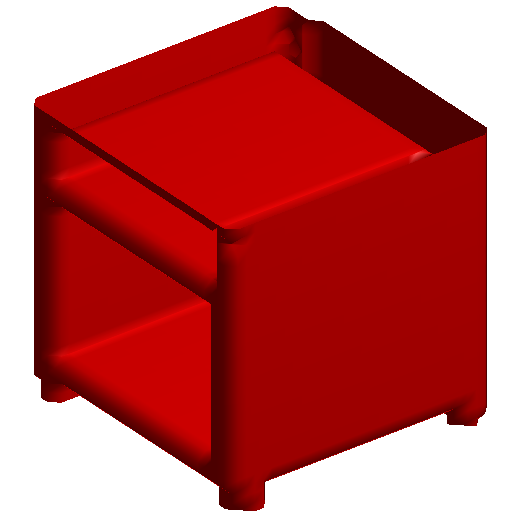} \hspace{-1.8mm}
    \includegraphics[height=.139\linewidth]{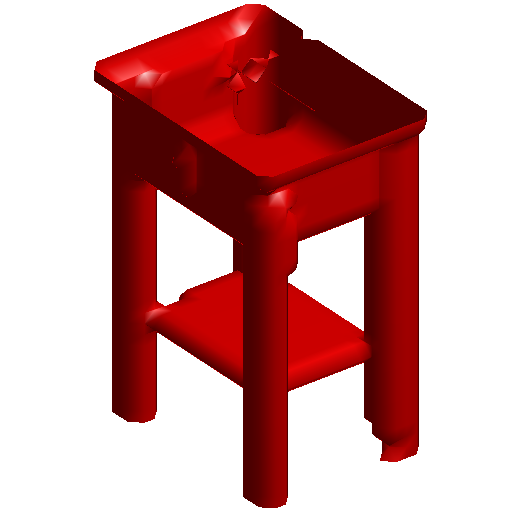} \hspace{-1.8mm}
    \includegraphics[height=.139\linewidth]{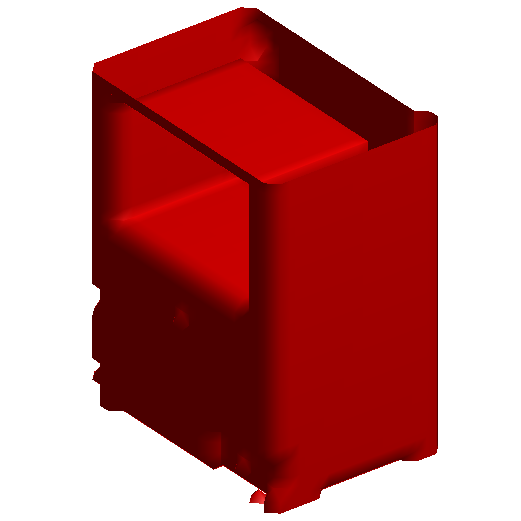} \hspace{-1.8mm}\\
	(b) night stand	
	\caption{3D object recovery by sampling from the conditional 3D DescriptorNet models. In each category, the first row displays the original 3D objects, the second row shows the corrupted 3D objects, and the third row displays the recovered 3D objects by running Langevin dynamics starting from the corrupted objects. (a) chair, (b) night stand.}	
	\label{fig:recovery}
\end{figure}

 After learning the model, we recover the corrupted voxels in each testing data $Y$ by sampling from $p(Y_M|Y_{\tilde{M}}, \theta)$ by running 90 Langevin dynamics steps. In the training stage, we randomly corrupt $70\%$ of each training 3D shape. In the testing stage, we experiment with the same percentage of corruption. We compare our method with 3D-GAN and 3D ShapeNets.
 We measure the recovery error by the average of per-voxel differences between the original testing data and the corresponding recovered data on the corrupted voxels. Table \ref{exp:recovery} displays the numerical comparison results for the 10 categories. Figure \ref{fig:recovery} displays some examples of 3D object recovery. For each experiment, the first row displays the original 3D objects, the second row displays the corrupted 3D objects, and the third row displays the recovered 3D objects that are sampled from the learned conditional distributions given the corrupted 3D objects as inputs.
 
 \begin{table}[h]
\caption{Recovery errors in occlusion experiments}\label{exp:recovery}
\vspace{-2mm}
\centering
\begin{small}
\begin{tabular}{|c|c|c|c|}
\hline
 category          & ours   & \cite{3dgan}      & \cite{wu20153d}  \\ \hline \hline
bathtub       & \textbf{0.0152} & 0.0266  & 0.0621  \\ \hline
bed   & \textbf{0.0068} & 0.0240  & 0.0617  \\ \hline
chair      & \textbf{0.0118} & 0.0238 & 0.0444 \\ \hline
desk    & \textbf{0.0122} & 0.0298& 0.0731    \\ \hline
dresser & \textbf{0.0038} &  0.0384  & 0.1558 \\ \hline
monitor & \textbf{0.0103} &  0.0220  & 0.0783  \\ \hline
night stand    & \textbf{0.0080} & 0.0248 & 0.2925  \\ \hline
sofa   & \textbf{0.0068} &  0.0186   & 0.0563 \\ \hline 
table    & \textbf{0.0051} & 0.0326 & 0.0340  \\ \hline
toilet   & \textbf{0.0119} & 0.0180 & 0.0977  \\ \hline
\hline
Avg.   & \textbf{0.0092} &  0.0259    & 0.0956  \\ \hline
\end{tabular}
\end{small}
\end{table}
\vspace{-2mm}

\subsection{3D object super-resolution}

\begin{figure}
	\centering
	\rotatebox[origin=l]{90}{\hspace{1mm}\textbf{{\scriptsize original}}}
	\includegraphics[height=.147\linewidth]{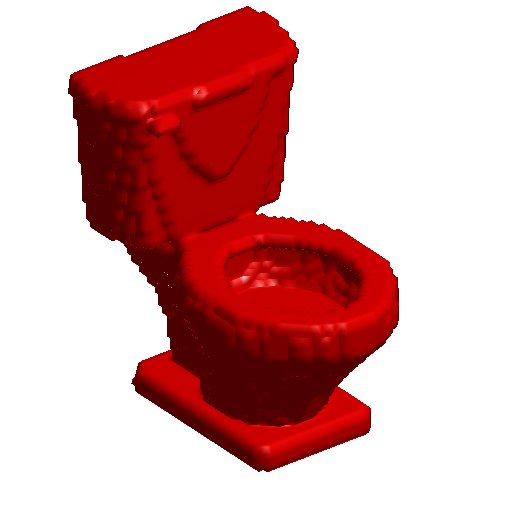}\hspace{-1.8mm}
	\includegraphics[height=.147\linewidth]{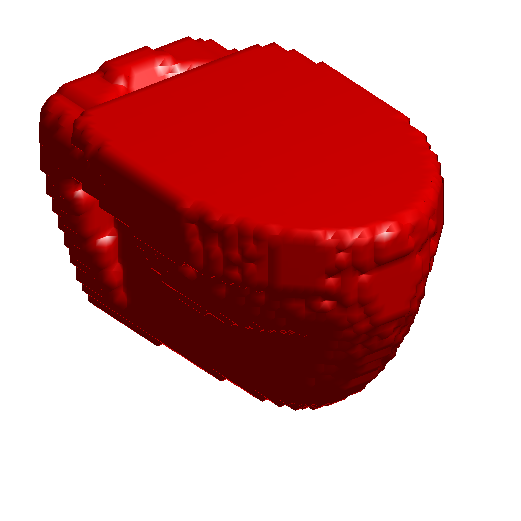}\hspace{-1.1mm}
	\includegraphics[height=.147\linewidth]{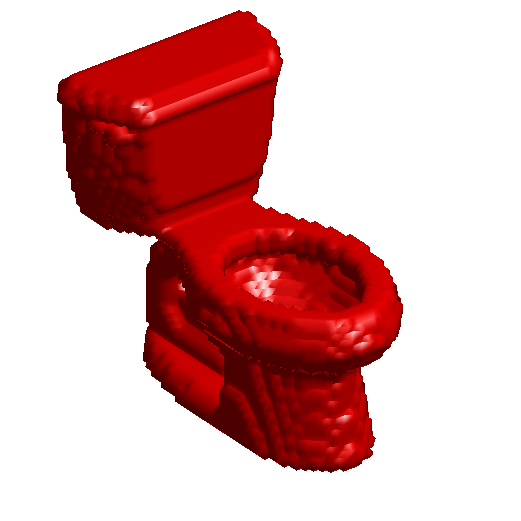}\hspace{-1.8mm}
	\includegraphics[height=.147\linewidth]{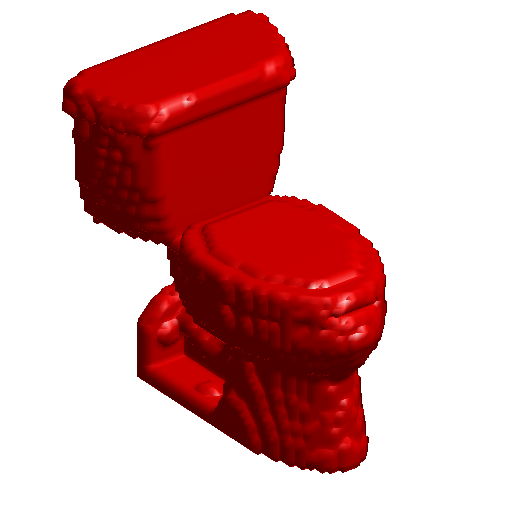}\hspace{-1.8mm}
	\includegraphics[height=.147\linewidth]{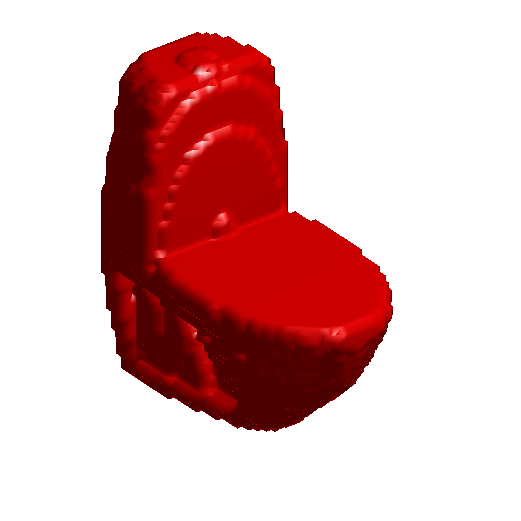}\hspace{-1.8mm}
	\includegraphics[height=.147\linewidth]{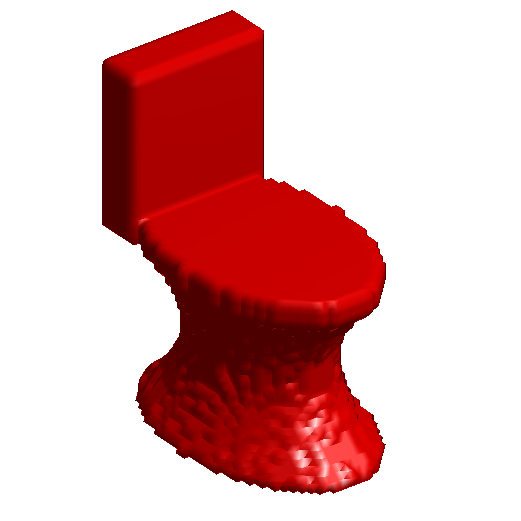}\hspace{-2.2mm}
	\includegraphics[height=.147\linewidth]{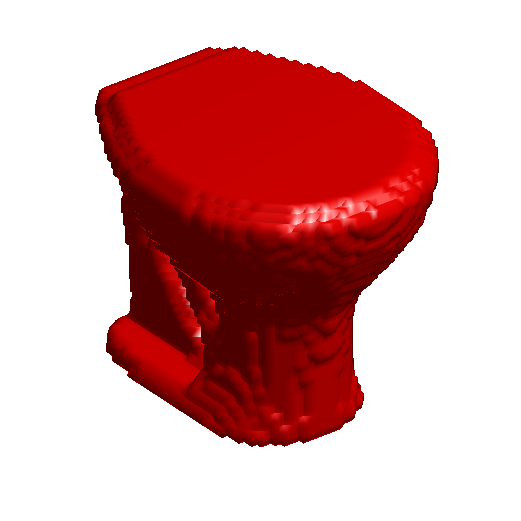}
	\\
	 \rotatebox[origin=l]{90}{\hspace{1mm}\textbf{{\scriptsize low res.}}}
	\includegraphics[height=.147\linewidth]{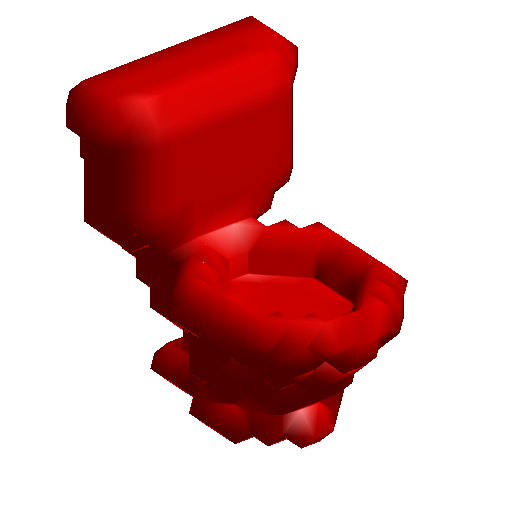}\hspace{-1.4mm}
    \includegraphics[height=.147\linewidth]{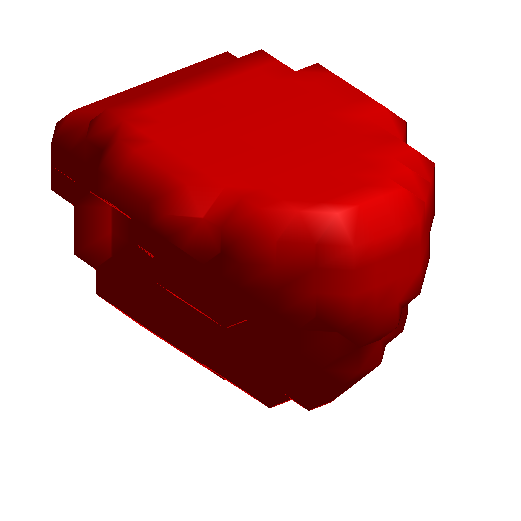}\hspace{-.4mm}
    \includegraphics[height=.147\linewidth]{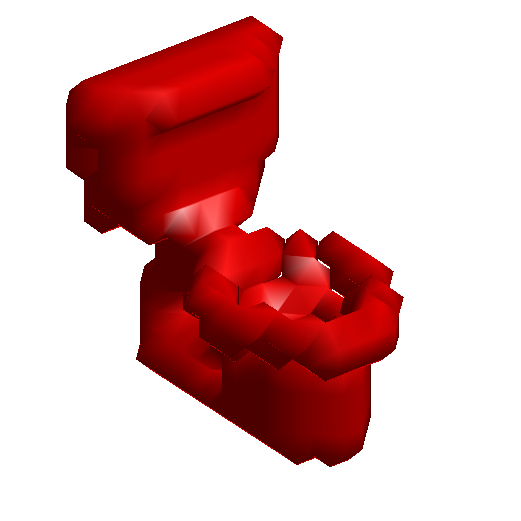}\hspace{-2.5mm}
    \includegraphics[height=.147\linewidth]{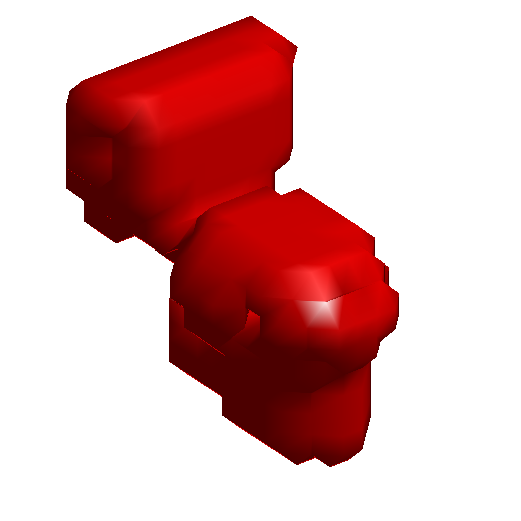}	\hspace{-2.9mm}
    \includegraphics[height=.147\linewidth]{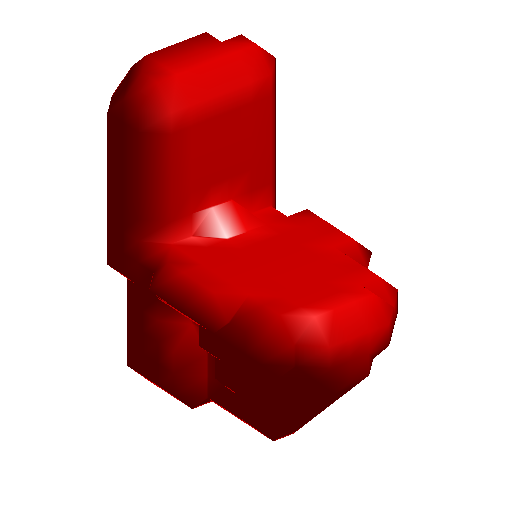}	\hspace{-2.5mm}
    \includegraphics[height=.147\linewidth]{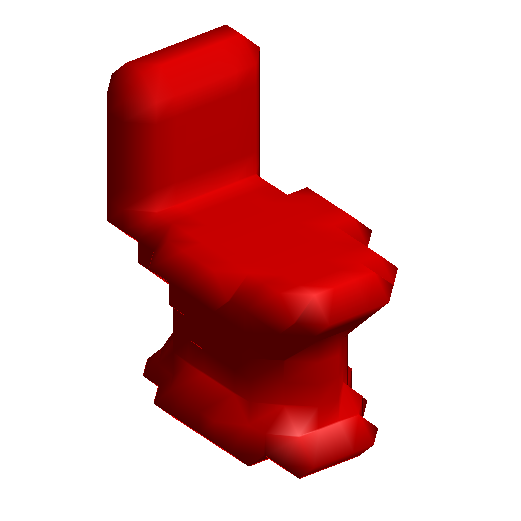}	\hspace{-2.5mm}
    \includegraphics[height=.147\linewidth]{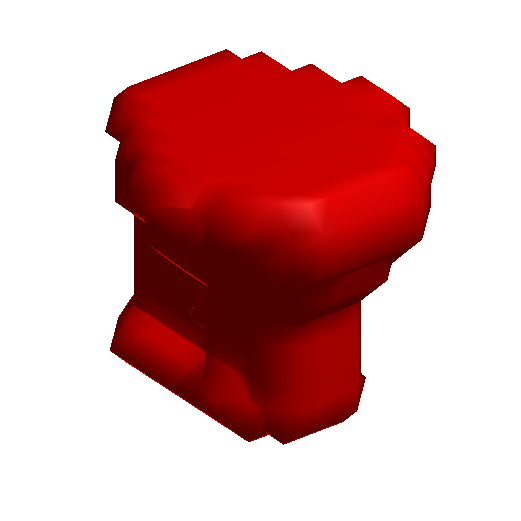}
   	\\
   	\rotatebox[origin=l]{90}{\hspace{1mm}\textbf{{\scriptsize high res.}}}
	\includegraphics[height=.147\linewidth]{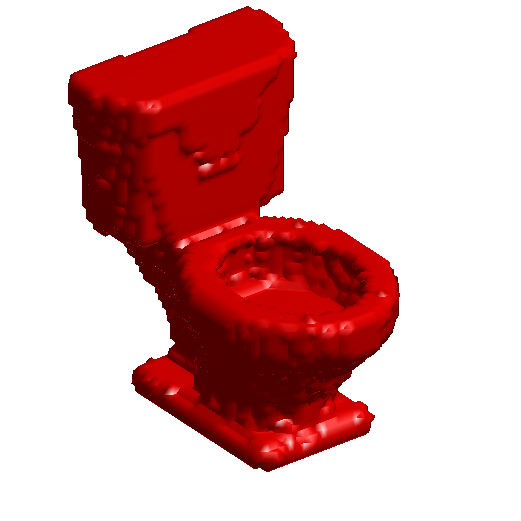}\hspace{-1.8mm}
	\includegraphics[height=.147\linewidth]{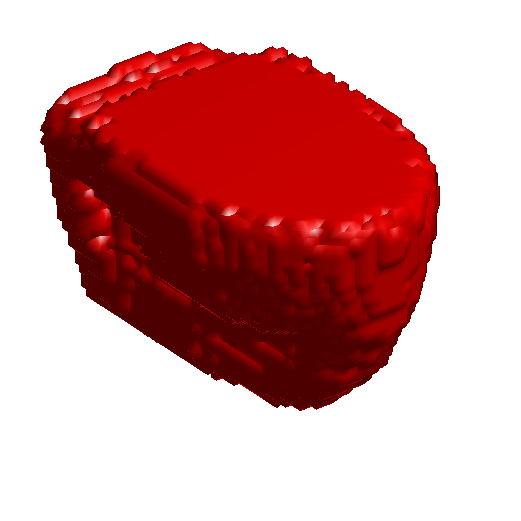}\hspace{-1.1mm}
	\includegraphics[height=.147\linewidth]{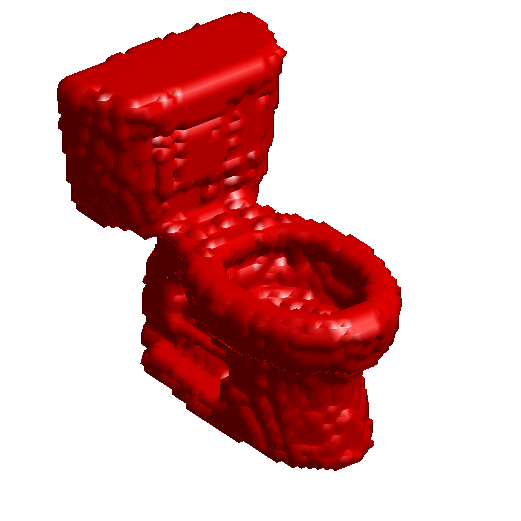}\hspace{-1.8mm}
	\includegraphics[height=.147\linewidth]{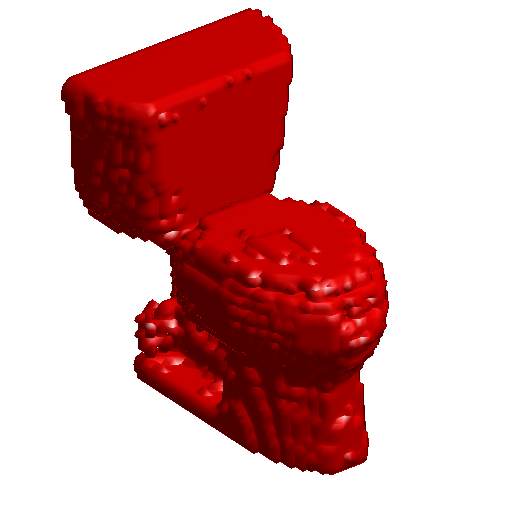}\hspace{-1.8mm}
	\includegraphics[height=.147\linewidth]{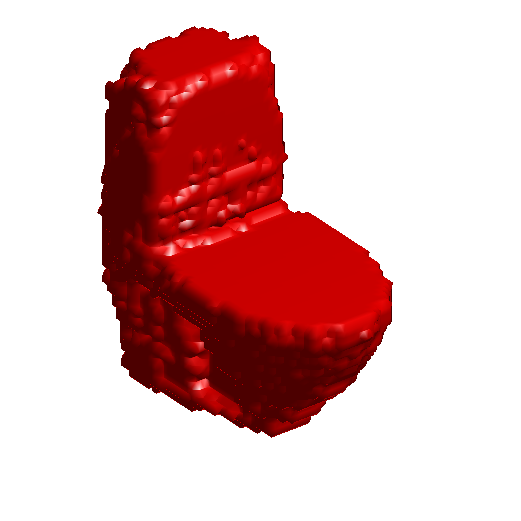}\hspace{-1.8mm}
	\includegraphics[height=.147\linewidth]{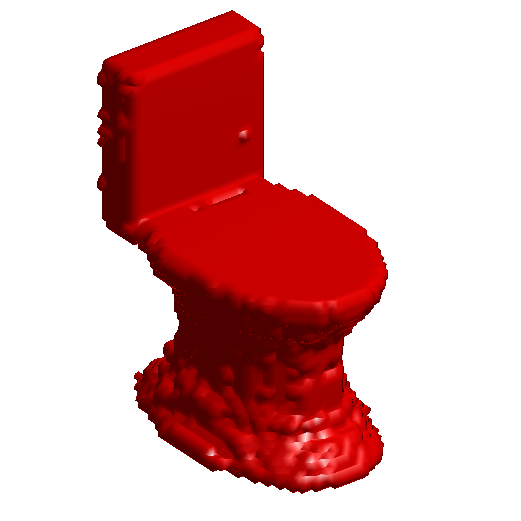}\hspace{-2.2mm}
	\includegraphics[height=.147\linewidth]{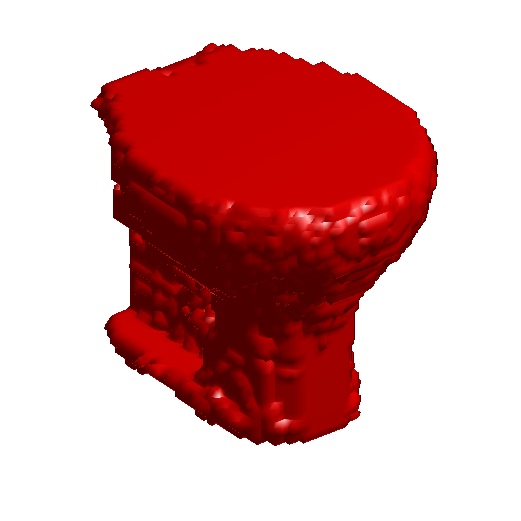}\\

		\caption{3D object super-resolution by conditional 3D DescriptorNet. The first row displays some original 3D objects ($64 \times 64 \times 64$ voxels). The second row shows the corresponding low resolution 3D objects ($16 \times 16 \times 16$ voxels). The last row displays the corresponding super-resolution results which are obtained by sampling from the conditional 3D DescriptorNet by running 10 steps of Langevin dynamics initialized with the objects shown in the second row.}	
	\label{exp:superResolution}
\end{figure}

\begin{figure*}
\centering
\begin{minipage}[b]{.34\textwidth}
  \centering
    \rotatebox[origin=l]{90}{\hspace{3mm}\textbf{{\footnotesize toilet}}} \hspace{-1mm}
     \includegraphics[width=.25\linewidth]{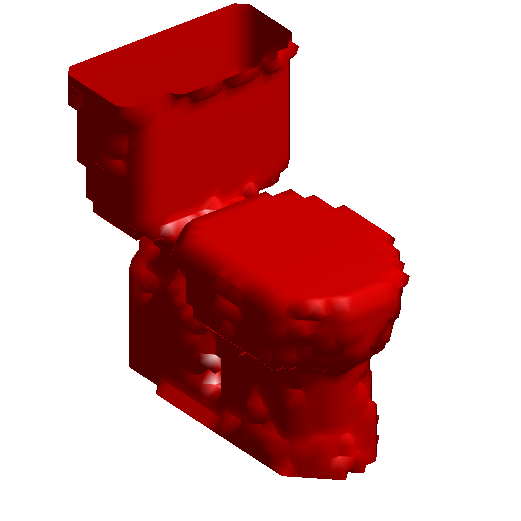}\hspace{-2mm} 
     \includegraphics[width=.25\linewidth]{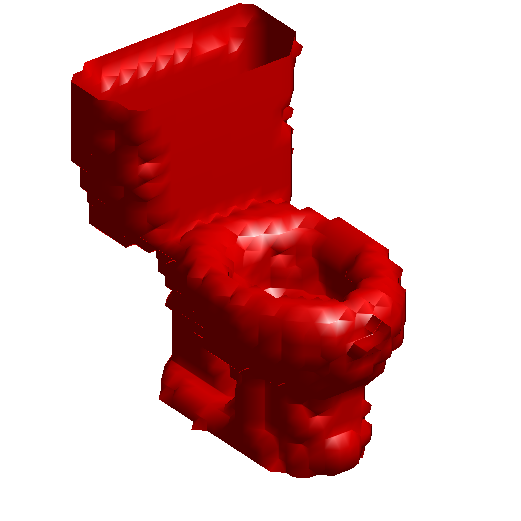}\hspace{-2mm} 
     \includegraphics[width=.25\linewidth]{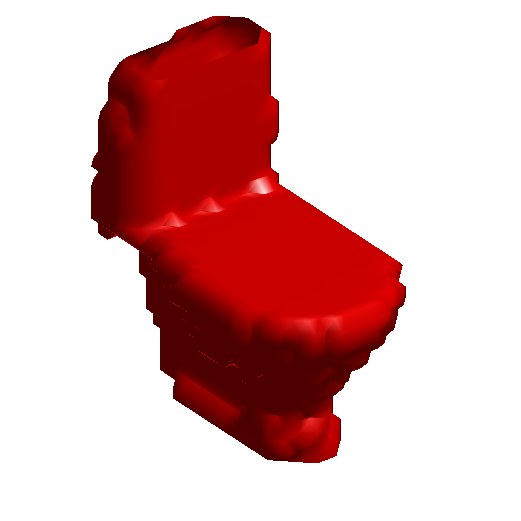}\hspace{-2mm} 
     \includegraphics[width=.25\linewidth]{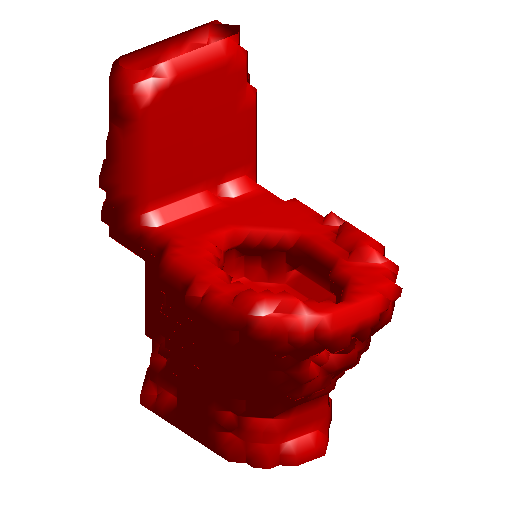}\hspace{-2mm}  \\	
	\rotatebox[origin=l]{90}{\hspace{3mm}\textbf{{\footnotesize sofa}}}
	\includegraphics[width=.23\linewidth]{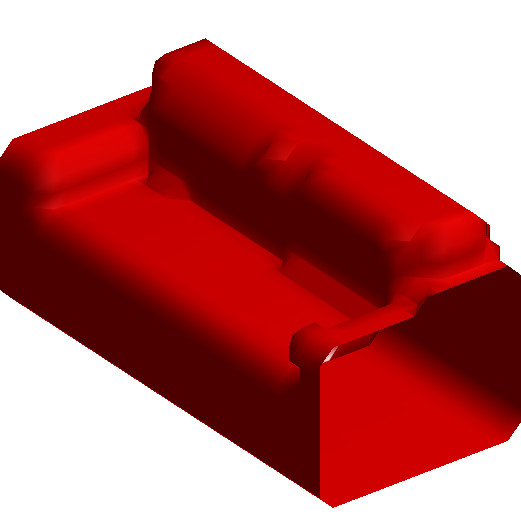}\hspace{-0.8mm}
	\includegraphics[width=.23\linewidth]{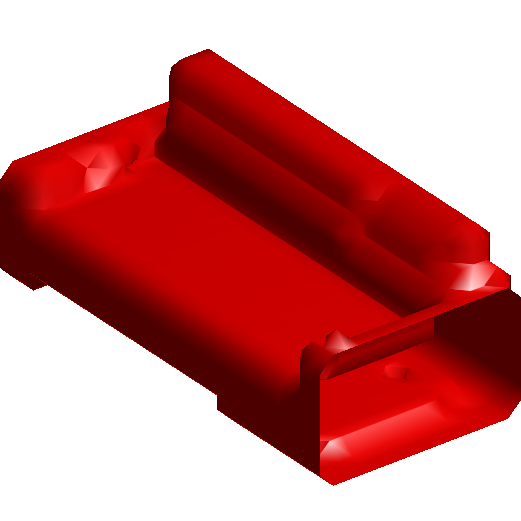}\hspace{-0.8mm}
	\includegraphics[width=.23\linewidth]{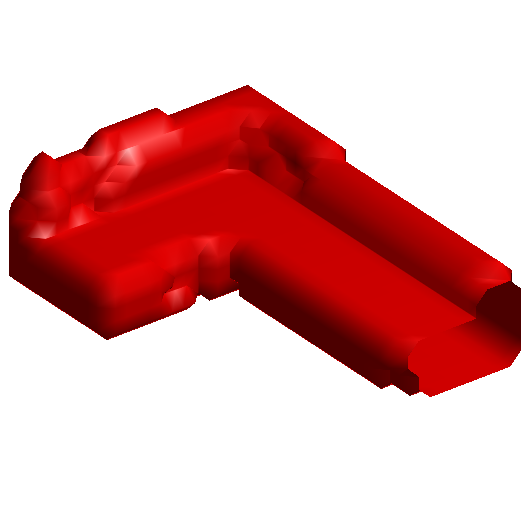}\hspace{-0.8mm}
	\includegraphics[width=.23\linewidth]{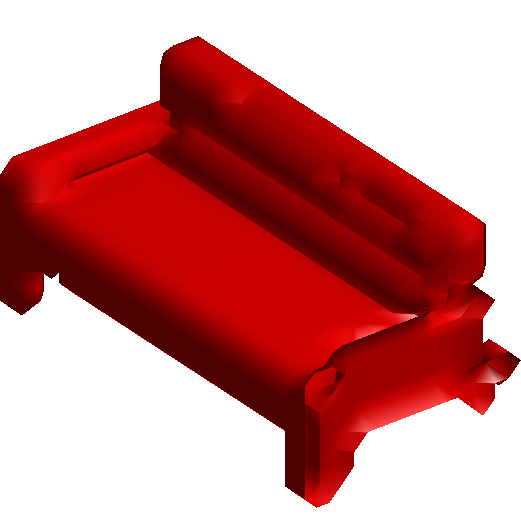}\hspace{-0.8mm}

  \caption{Synthesis by 3D generators}
  \label{fig:synthesis_generator}
\end{minipage}%
~
\begin{minipage}[b]{.64\textwidth}
  \centering
  \rotatebox[origin=l]{90}{\hspace{3mm}\textbf{{\footnotesize toilet}}}
  	\includegraphics[width=.14\linewidth]{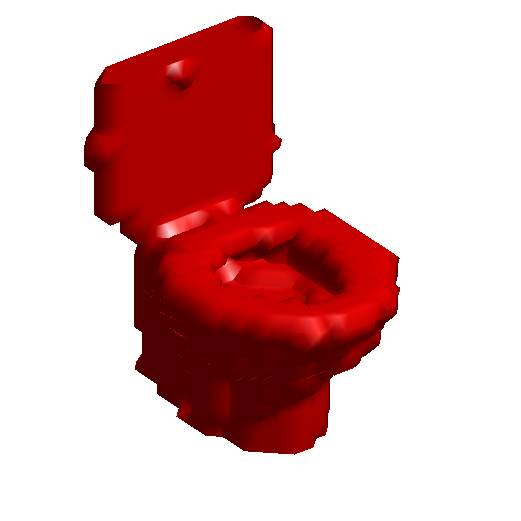}\hspace{-3.5mm} 
	\includegraphics[width=.14\linewidth]{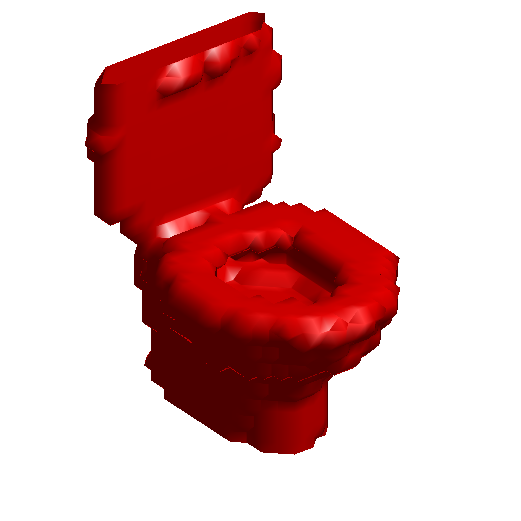}\hspace{-3.5mm} 
	\includegraphics[width=.14\linewidth]{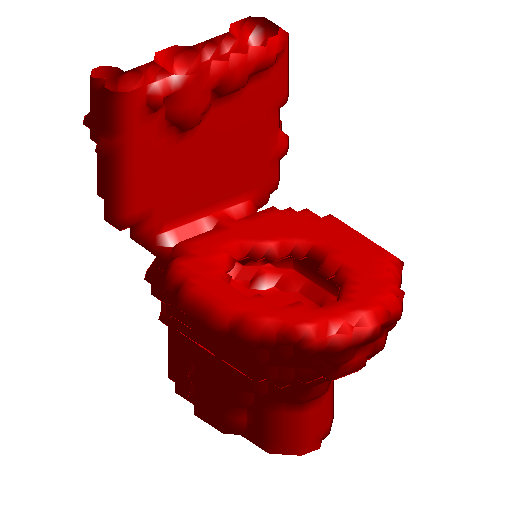}\hspace{-3.5mm} 
    \includegraphics[width=.14\linewidth]{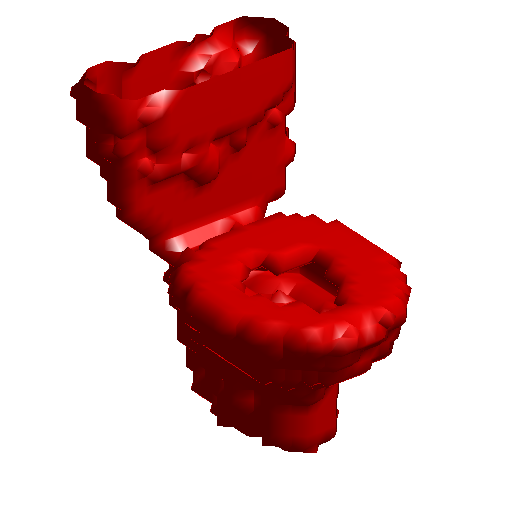}\hspace{-3.5mm} 
    \includegraphics[width=.14\linewidth]{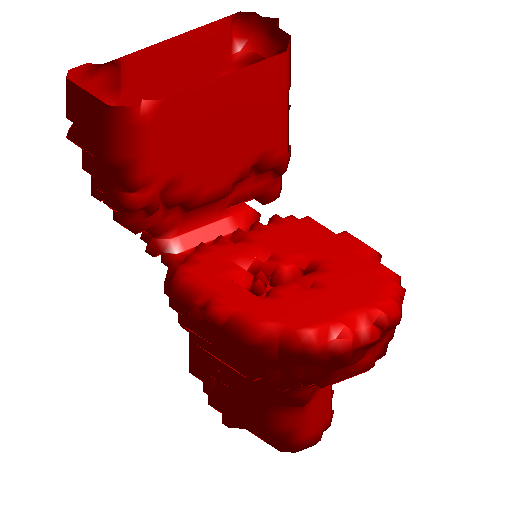}\hspace{-3.5mm} 
    \includegraphics[width=.14\linewidth]{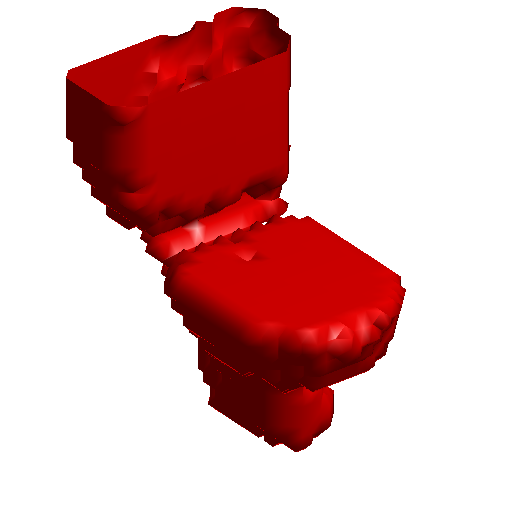}\hspace{-3.5mm} 
    \includegraphics[width=.14\linewidth]{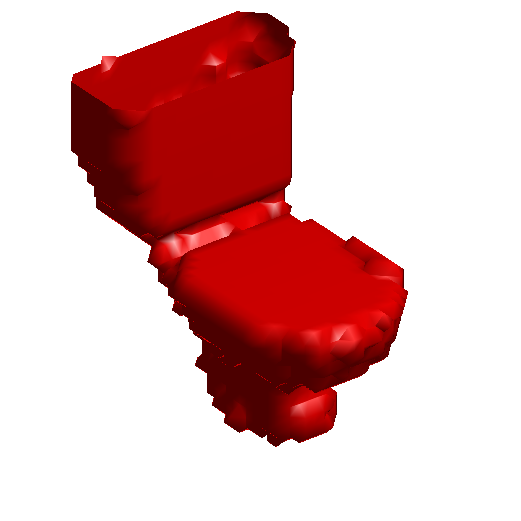}\hspace{-3.5mm} 
    \includegraphics[width=.14\linewidth]{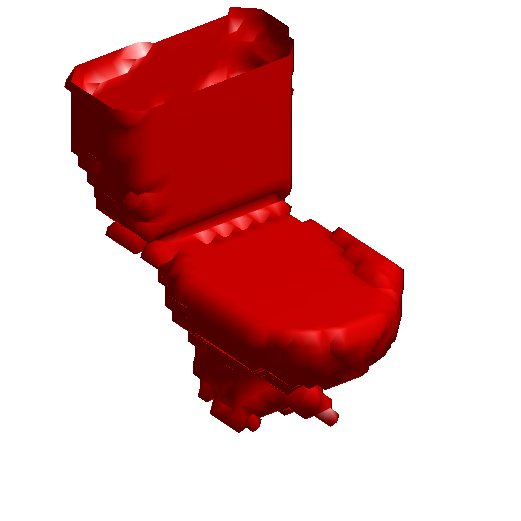}\\
    \rotatebox[origin=l]{90}{\hspace{3mm}\textbf{{\footnotesize sofa}}}
    	\includegraphics[width=.12\linewidth]{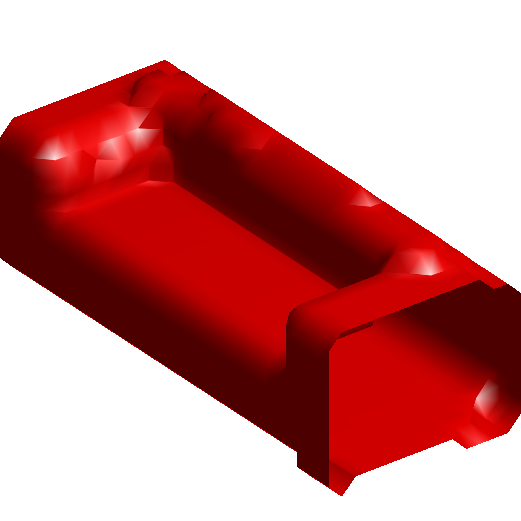} \hspace{-1.7mm} 
	\includegraphics[width=.12\linewidth]{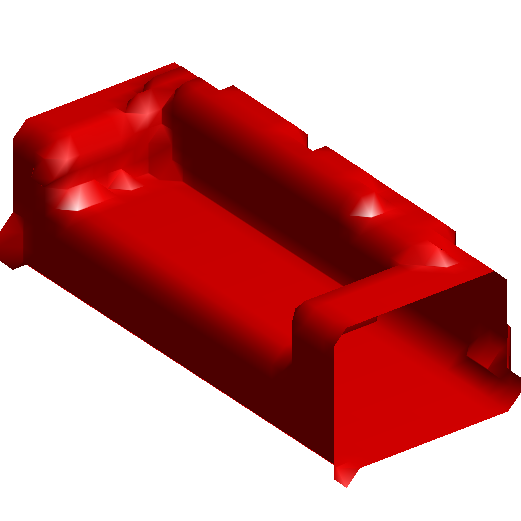} \hspace{-1.7mm}  
	\includegraphics[width=.12\linewidth]{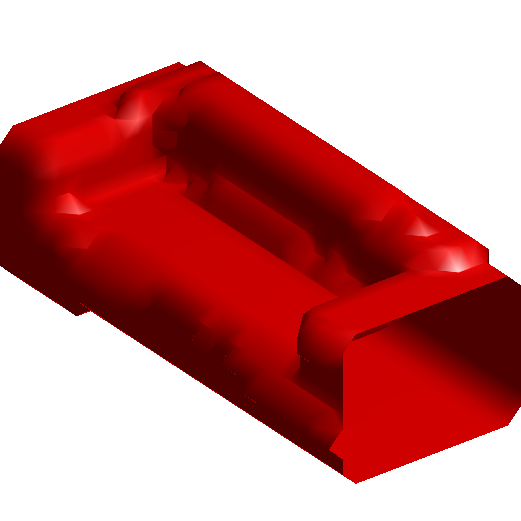} \hspace{-1.7mm} 
	\includegraphics[width=.12\linewidth]{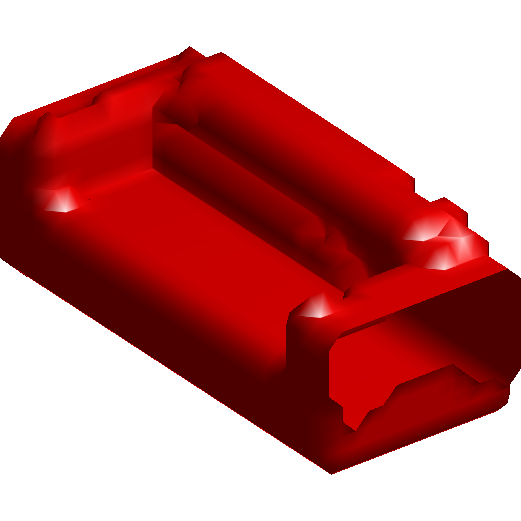} \hspace{-1.7mm}  
	\includegraphics[width=.12\linewidth]{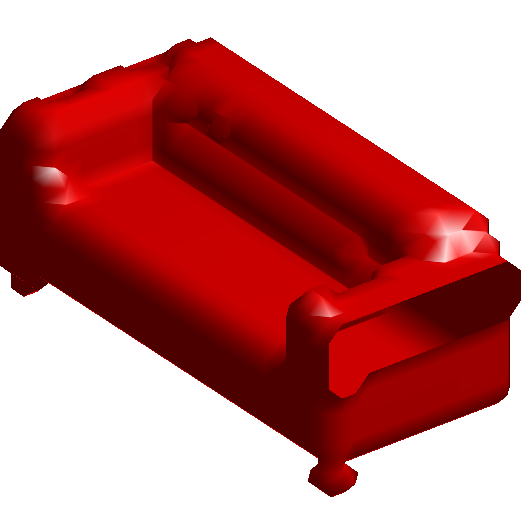} \hspace{-1.7mm} 
	\includegraphics[width=.12\linewidth]{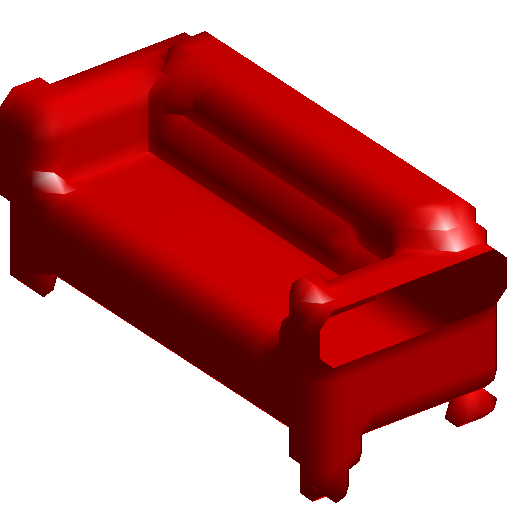} \hspace{-1.7mm} 
	\includegraphics[width=.12\linewidth]{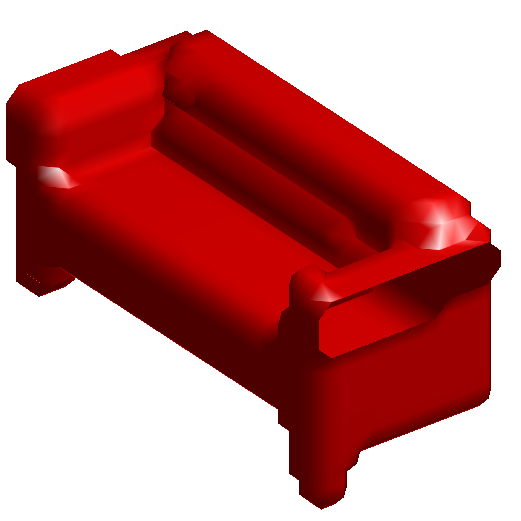} \hspace{-1.7mm} 
	\includegraphics[width=.12\linewidth]{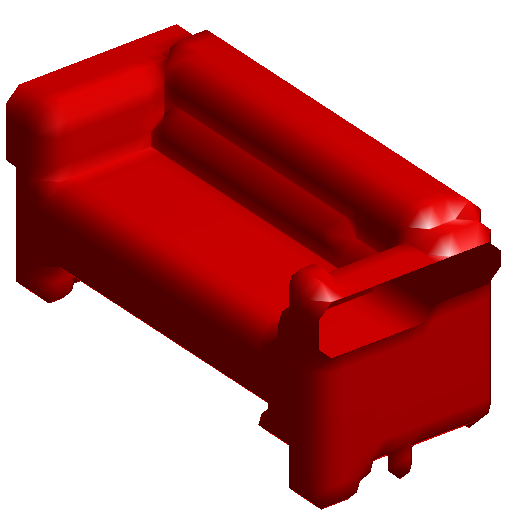}
  \caption{Interpolation between latent vectors of the 3D objects on the two ends}
  \label{fig:interpolation}
\end{minipage}
\end{figure*}

\begin{figure}
	\centering
	\includegraphics[height=.139\linewidth]{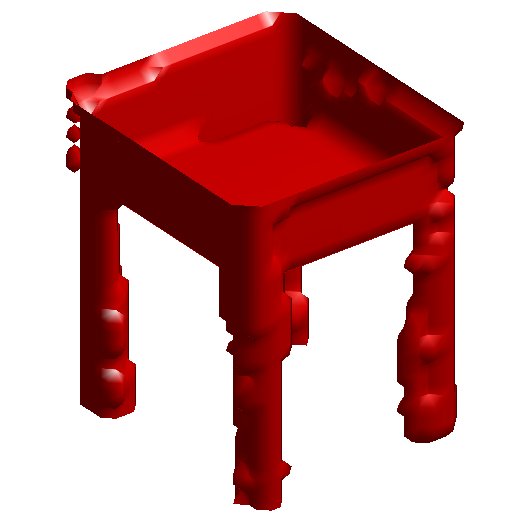} 
	\includegraphics[height=.139\linewidth]{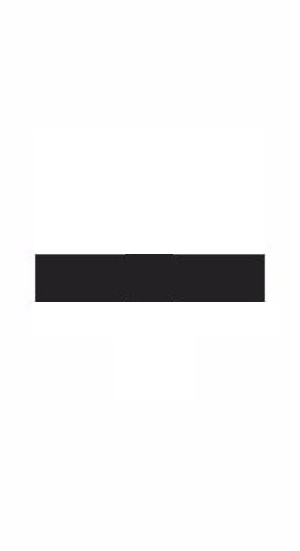} 
	\includegraphics[height=.139\linewidth]{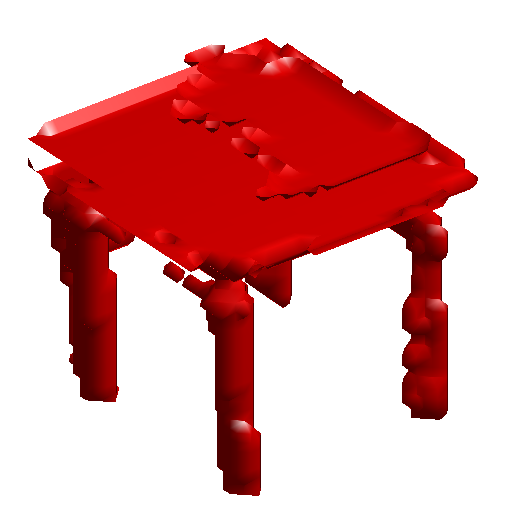} \hspace{-2mm}
	\includegraphics[height=.139\linewidth]{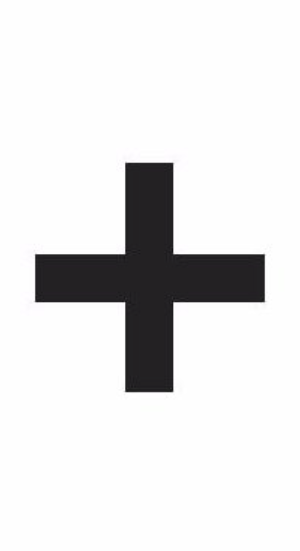} 
	\includegraphics[height=.139\linewidth]{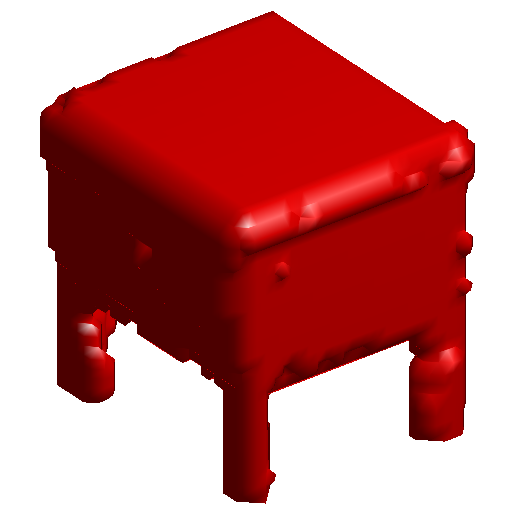} 
	\includegraphics[height=.139\linewidth]{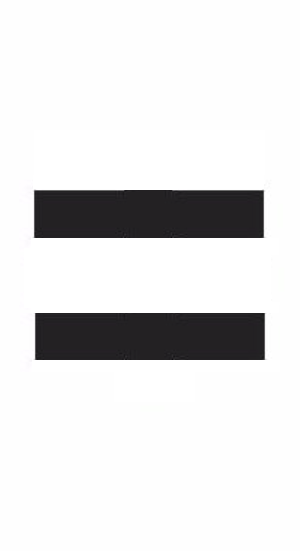} 
	\includegraphics[height=.139\linewidth]{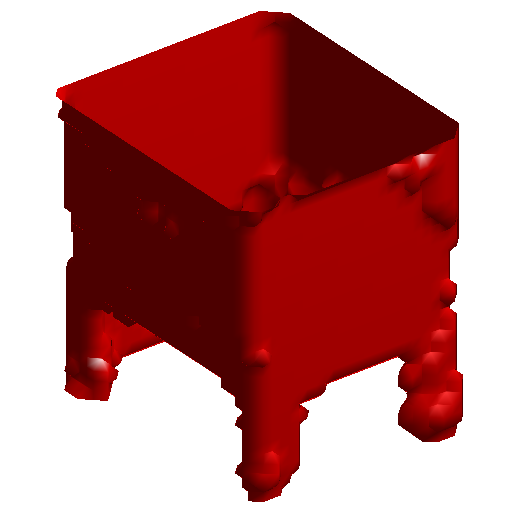} \hspace{-3mm}	\\
	\includegraphics[height=.139\linewidth]{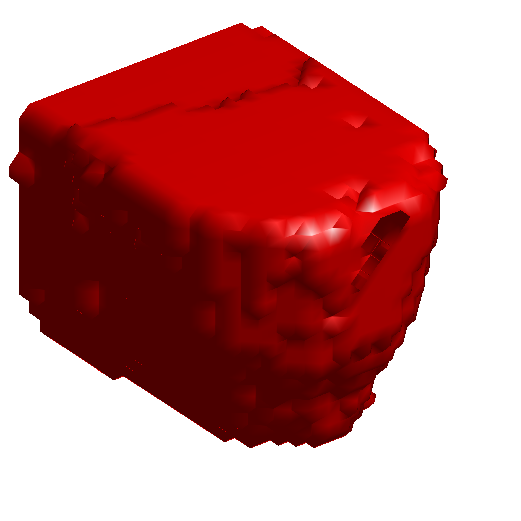} \hspace{-3mm}
	\includegraphics[height=.139\linewidth]{./figures/output_arith/minus.png} 
	\includegraphics[height=.139\linewidth]{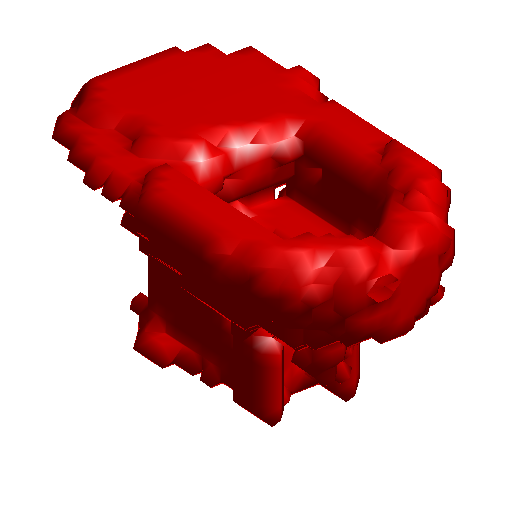} \hspace{-2mm}
	\includegraphics[height=.139\linewidth]{./figures/output_arith/plus.png} 
	\includegraphics[height=.139\linewidth]{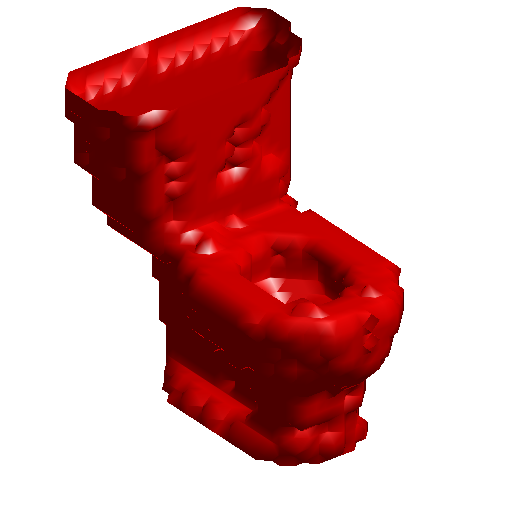} 
	\includegraphics[height=.139\linewidth]{./figures/output_arith/equal.png} 
	\includegraphics[height=.139\linewidth]{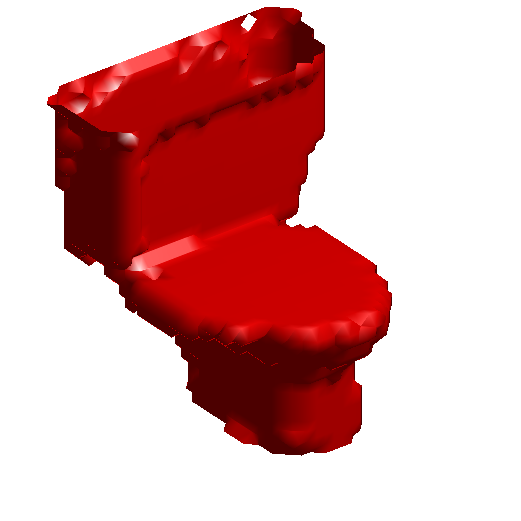}  \hspace{-3mm}
	\caption{3D shape arithmetic in the latent space}
	\label{fig:arith}
\end{figure}	

We test the conditional 3D DescriptorNet on the 3D object super-resolution task. Similar to Experiment \ref{Exp:objectRecovery}, we can perform super-resolution on a low resolution 3D objects by sampling from a conditional 3D DescriptorNet $p(Y_{\rm high}|Y_{\rm low}, \theta)$, where $Y_{\rm high}$ denotes a high resolution version of $Y_{\rm low}$. 
The sampling of the conditional model $p(Y_{\rm high}|Y_{\rm low}, \theta)$ is accomplished by the Langevin dynamics initialized with the given low resolution 3D object that needs to be super-resolutioned. In the learning stage, we learn the conditional model from the fully observed training 3D objects as well as their low resolution versions. To specialize the learned model to this super-resolution task, in the training process, we down-scale each fully observed training 3D object $Y_{\rm high}$ into a low resolution version $Y_{\rm low}$, which leads to information loss. In each iteration, we first up-scale $Y_{\rm low}$ by expanding each voxel of $Y_{\rm low}$ into a $d \times d \times d$ block (where $d$ is the ratio between the sizes of $Y_{\rm high}$ and $Y_{\rm low}$) of constant values to obtain an up-scaled version  $Y^{'}_{\rm high}$ of $Y_{\rm low}$ (The up-scaled $Y^{'}_{\rm high}$ is not identical to the original high resolution $Y_{\rm high}$ since the high resolution details are lost), and then run Langevin dynamics starting from $Y^{'}_{\rm high}$. The parameters $\theta$ are then updated by gradient ascent according to (\ref{eq:lD2}). Figure \ref{exp:superResolution} shows some qualitative results of  3D super-resolution, where we use a 2-layer conditional 3D DescriptorNet. The first layer has 200 $16 \times 16 \times 16$ filters with sub-sampling of 3. The second layer is a fully-connected layer with one single filter. The Langevin step size is 0.01.

To be more specific, let $Y_{\rm low} = C Y_{\rm high}$, where $C$ is the down-scaling matrix, e.g., each voxel of $Y_{\rm low}$ is the average of  the corresponding $d \times d \times d$ block of $Y_{\rm high}$. Let $C^{-}$ be the pseudo-inverse of $C$, e.g., $C^{-} Y_{\rm low}$ gives us a high resolution shape by expanding each voxel of $Y_{\rm low}$ into a $d \times d\times d$ block of constant values. Then the sampling of $p(Y_{\rm high} | Y_{\rm low}; \theta)$ is similar to sampling the unconditioned model $p(Y_{\rm high}; \theta)$, except that for each step of the Langevin dynamics, let $\Delta Y$ be the change of $Y$, we update $Y \leftarrow Y +  (I - C^{-} C) \Delta Y$, i.e., we project $\Delta Y$ to the null space of $C$, so that  the low resolution version of $Y$, i.e., $CY$, remains fixed.    From this perspective, super-resolution is similar to inpainting, except that the visible voxels are replaced by low resolution voxels.

\subsection{Analyzing the learned 3D generator}

We evaluate a 3D generator trained by a 3D DescriptorNet via MCMC teaching. The generator network $g(Z; \alpha)$ has 4 layers of volumetric deconvolution with $4 \times 4 \times 4$ kernels, with up-sampling factors $\{1, 2, 2, 2\}$ at different layers respectively. The numbers of channels at different layers are 256, 128, 64, and 1. There is a fully connected layer under the 100 dimensional latent factors $Z$. The output size is $32 \times 32 \times 32$. Batch normalization and ReLU layers are used between deconvolution layers and tanh non-linearity is added at the bottom-layer. We train a 3D DescriptorNet with the above 3D generator as a sampler in a cooperative training scheme presented in Algorithm \ref{code:3} for the categories of toilet, sofa, and nightstand in ModelNet10 dataset independently. The 3D DescriptorNet has a 4-layer network, where the first layer has 64 $9 \times 9 \times 9$ filters, the second layer has 128 $7 \times 7 \times 7$ filters, the third layer has 256 $4 \times 4 \times 4$ filters, and the fourth layer is a fully connected layer with a single filter. The sub-sampling factors are $\{2, 2, 2, 1\}$. ReLU layers are used between convolutional layers. 
 
We use Adam for optimization of 3D DescriptorNet with $\beta_1=0.4$ and $ \beta_2=0.999$, and for optimization of 3D generator with $\beta_1=0.6$ and $\beta_2=0.999$. The learning rates for 3D DescriptorNet and 3D generator are 0.001 and 0.0003 respectively. The number of parallel chains is 50, and the mini-batch size is 50. The training data are scaled into the range of $[-1, 1]$. The synthesized data are re-scaled back into $[0, 1]$ for visualization. Figure \ref{fig:synthesis_generator} shows some examples of 3D objects generated by the 3D generators trained by the 3D DescriptorNet via MCMC teaching. 
 
We show results of interpolating between two latent vectors of $Z$ in Figure \ref{fig:interpolation}. For each row, the 3D objects at the two ends are generated from $Z$ vectors that are randomly sampled from ${\rm N}(0,I_d)$. Each object in the middle is obtained by first interpolating the $Z$ vectors of the two end objects, and then generating the objects using the 3D generator. We observe smooth transitions in 3D shape structure and that most intermediate objects are also physically plausible.
This experiment demonstrates that the learned 3D generator embeds the 3D object distribution into a smooth low dimensional manifold. Another way to investigate the learned 3D generator is to show shape arithmetic in the latent space. As shown in Figure \ref{fig:arith}, the 3D generator is able to encode semantic knowledge of 3D shapes in its latent space such that arithmetic can be performed on $Z$ vectors for visual concept manipulation of 3D shapes. 

\subsection{3D object classification}

We evaluate the feature maps learned by our 3D DescriptorNet. We perform a classification experiment on ModelNet10 dataset. We first train a single model on all categories of the training set in an unsupervised manner. The network architecture and learning configuration are the same as the one used for synthesis in Section \ref{Exp:objectSynthesis}. 
Then we use the model as a feature extractor. Specifically, for each input 3D object, we use the model to extract its first and second layers of feature maps, apply max pooling of kernel sizes $4 \times 4 \times 4$ and $2 \times 2\times 2$ respectively, and concatenate the outputs as a feature vector of length 8,100. We train a multinomial logistic regression classifier from labeled data based on the extracted feature vectors for classification. We evaluate the classification accuracy of the classifier on the testing data using the one-versus-all rule. For comparison, Table \ref{exp:classification} lists 8 published results on this dataset obtained by other baseline methods. Our method outperforms the other methods in terms of classification accuracy on this dataset.
 
\begin{table}[h]
\caption{3D object classification on ModelNet10 dataset}\label{exp:classification}
\vspace{-2mm}
\centering
\begin{small}
\begin{tabular}{|l|c|}
\hline
          Method   &  Accuracy\\ \hline \hline
     Geometry Image \cite{sinha2016deep} & 88.4$\%$\\ \hline      
      
      PANORAMA-NN    \cite{sfikas2017exploiting} & 91.1$\%$ \\ \hline
    ECC  \cite{simonovsky2017dynamic} & 90.0$\%$\\ \hline    
       3D ShapeNets \cite{wu20153d}  & 83.5$\%$ \\ \hline
        DeepPano \cite{shi2015deeppano} & 85.5$\%$ \\ \hline
 SPH  \cite{kazhdan2003rotation}     & 79.8$\%$    \\ \hline
 VConv-DAE   \cite{sharma2016vconv}   & 80.5$\%$   \\ \hline
 3D-GAN  \cite{3dgan}  & 91.0$\%$  \\ \hline
  3D DescriptorNet (ours) & \textbf{92.4}$\%$\\ \hline 
\end{tabular}
\end{small}
\end{table}
\vspace{-3mm}

\section{Conclusion}
We propose the 3D DescriptorNet for volumetric object synthesis, and the conditional 3D DescriptorNet for 3D object recovery and 3D object super resolution. The proposed model is a  deep convolutional energy-based model, which can be trained by an ``analysis by synthesis'' scheme. The training of the model can be interpreted as a mode seeking and mode shifting process, and the zero temperature limit has an adversarial interpretation. A 3D generator can be taught by the 3D DescriptorNet via MCMC teaching. Experiments demonstrate that our models are able to generate realistic 3D shape patterns and are useful for 3D shape analysis.

\section*{Acknowledgment}

The work is supported by Hikvision gift fund, DARPA SIMPLEX N66001-15-C-4035,  ONR MURI N00014-16-1-2007, DARPA ARO W911NF-16-1-0579, and DARPA  N66001-17-2-4029. We thank Erik Nijkamp for his help on coding. We thank Siyuan Huang for helpful discussions.

{\small
\bibliographystyle{ieee}
\bibliography{mybibfileICML}

\begin{thebibliography}{10}\itemsep=-1pt

\bibitem{arjovsky2017wasserstein}
M.~Arjovsky, S.~Chintala, and L.~Bottou.
\newblock Wasserstein {GAN}.
\newblock {\em arXiv preprint arXiv:1701.07875}, 2017.

\bibitem{blanz1999morphable}
V.~Blanz and T.~Vetter.
\newblock A morphable model for the synthesis of {3D} faces.
\newblock In {\em Proceedings of the 26th annual conference on Computer
  graphics and interactive techniques}, pages 187--194, 1999.

\bibitem{carlson1982algorithm}
W.~E. Carlson.
\newblock An algorithm and data structure for {3D} object synthesis using
  surface patch intersections.
\newblock {\em ACM SIGGRAPH Computer Graphics}, 16(3):255--263, 1982.

\bibitem{chang2015shapenet}
A.~X. Chang, T.~Funkhouser, L.~Guibas, P.~Hanrahan, Q.~Huang, Z.~Li,
  S.~Savarese, M.~Savva, S.~Song, H.~Su, et~al.
\newblock Shapenet: An information-rich {3D} model repository.
\newblock {\em arXiv preprint arXiv:1512.03012}, 2015.

\bibitem{girdhar2016learning}
R.~Girdhar, D.~F. Fouhey, M.~Rodriguez, and A.~Gupta.
\newblock Learning a predictable and generative vector representation for
  objects.
\newblock In {\em European Conference on Computer Vision}, pages 484--499,
  2016.

\bibitem{goodfellow2014generative}
I.~Goodfellow, J.~Pouget-Abadie, M.~Mirza, B.~Xu, D.~Warde-Farley, S.~Ozair,
  A.~Courville, and Y.~Bengio.
\newblock Generative adversarial nets.
\newblock In {\em Advances in Neural Information Processing Systems}, pages
  2672--2680, 2014.

\bibitem{grenander2007pattern}
U.~Grenander and M.~I. Miller.
\newblock {\em Pattern theory: from representation to inference}.
\newblock Oxford University Press, 2007.

\bibitem{jin2017introspective}
L.~Jin, J.~Lazarow, and Z.~Tu.
\newblock Introspective classification with convolutional nets.
\newblock In {\em Advances in Neural Information Processing Systems}, pages
  823--833, 2017.

\bibitem{kalogerakis2012probabilistic}
E.~Kalogerakis, S.~Chaudhuri, D.~Koller, and V.~Koltun.
\newblock A probabilistic model for component-based shape synthesis.
\newblock {\em ACM Transactions on Graphics (TOG)}, 31(4):55, 2012.

\bibitem{kazhdan2003rotation}
M.~Kazhdan, T.~Funkhouser, and S.~Rusinkiewicz.
\newblock Rotation invariant spherical harmonic representation of {3D} shape
  descriptors.
\newblock In {\em Symposium on geometry processing}, volume~6, pages 156--164,
  2003.

\bibitem{kingma2015adam}
D.~P. Kingma and J.~Ba.
\newblock Adam: A method for stochastic optimization.
\newblock In {\em International Conference on Learning Representations}, 2015.

\bibitem{kingma2013auto}
D.~P. Kingma and M.~Welling.
\newblock Auto-encoding variational bayes.
\newblock In {\em International Conference on Learning Representations}, 2014.

\bibitem{lazarow2017introspective}
J.~Lazarow, L.~Jin, and Z.~Tu.
\newblock Introspective neural networks for generative modeling.
\newblock In {\em IEEE Conference on Computer Vision and Pattern Recognition},
  pages 2774--2783, 2017.

\bibitem{Lecun2006}
Y.~LeCun, S.~Chopra, R.~Hadsell, M.~Ranzato, and F.~J. Huang.
\newblock A tutorial on energy-based learning.
\newblock In {\em Predicting Structured Data}. MIT Press, 2006.

\bibitem{lu2015learning}
Y.~Lu, S.-C. Zhu, and Y.~N. Wu.
\newblock Learning {FRAME} models using {CNN} filters.
\newblock {\em arXiv preprint arXiv:1509.08379}, 2015.

\bibitem{maturana2015voxnet}
D.~Maturana and S.~Scherer.
\newblock Voxnet: A {3D} convolutional neural network for real-time object
  recognition.
\newblock In {\em International Conference on Intelligent Robots and Systems
  (IROS)}, pages 922--928, 2015.

\bibitem{qi2016volumetric}
C.~R. Qi, H.~Su, M.~Nie{\ss}ner, A.~Dai, M.~Yan, and L.~J. Guibas.
\newblock Volumetric and multi-view {CNNs} for object classification on {3D}
  data.
\newblock In {\em IEEE Conference on Computer Vision and Pattern Recognition},
  pages 5648--5656, 2016.

\bibitem{radford2015unsupervised}
A.~Radford, L.~Metz, and S.~Chintala.
\newblock Unsupervised representation learning with deep convolutional
  generative adversarial networks.
\newblock {\em arXiv preprint arXiv:1511.06434}, 2015.

\bibitem{sfikas2017exploiting}
K.~Sfikas, T.~Theoharis, and I.~Pratikakis.
\newblock Exploiting the panorama representation for convolutional neural
  network classification and retrieval.
\newblock In {\em Eurographics Workshop on 3D Object Retrieval}, 2017.

\bibitem{sharma2016vconv}
A.~Sharma, O.~Grau, and M.~Fritz.
\newblock {VConv-DAE}: Deep volumetric shape learning without object labels.
\newblock In {\em European Conference on Computer Vision}, pages 236--250,
  2016.

\bibitem{shi2015deeppano}
B.~Shi, S.~Bai, Z.~Zhou, and X.~Bai.
\newblock {DeepPano}: Deep panoramic representation for {3-D} shape
  recognition.
\newblock {\em IEEE Signal Processing Letters}, 22(12):2339--2343, 2015.

\bibitem{simonovsky2017dynamic}
M.~Simonovsky and N.~Komodakis.
\newblock Dynamic edge-conditioned filters in convolutional neural networks on
  graphs.
\newblock In {\em IEEE Conference on Computer Vision and Pattern Recognition},
  2017.

\bibitem{sinha2016deep}
A.~Sinha, J.~Bai, and K.~Ramani.
\newblock Deep learning {3D} shape surfaces using geometry images.
\newblock In {\em European Conference on Computer Vision}, pages 223--240,
  2016.

\bibitem{su2015multi}
H.~Su, S.~Maji, E.~Kalogerakis, and E.~Learned-Miller.
\newblock Multi-view convolutional neural networks for {3D} shape recognition.
\newblock In {\em International Conference on Computer Vision}, pages 945--953,
  2015.

\bibitem{tu2007learning}
Z.~Tu.
\newblock Learning generative models via discriminative approaches.
\newblock In {\em IEEE Conference on Computer Vision and Pattern Recognition},
  pages 1--8, 2007.

\bibitem{warde2016improving}
D.~Warde-Farley and Y.~Bengio.
\newblock Improving generative adversarial networks with denoising feature
  matching.
\newblock In {\em International Conference on Learning Representations}, 2017.

\bibitem{welling2009herding}
M.~Welling.
\newblock Herding dynamical weights to learn.
\newblock In {\em International Conference on Machine Learning}, pages
  1121--1128, 2009.

\bibitem{3dgan}
J.~Wu, C.~Zhang, T.~Xue, W.~T. Freeman, and J.~B. Tenenbaum.
\newblock Learning a probabilistic latent space of object shapes via {3D}
  generative-adversarial modeling.
\newblock In {\em Advances in Neural Information Processing Systems}, pages
  82--90, 2016.

\bibitem{wu20153d}
Z.~Wu, S.~Song, A.~Khosla, F.~Yu, L.~Zhang, X.~Tang, and J.~Xiao.
\newblock {3D} shapenets: A deep representation for volumetric shapes.
\newblock In {\em IEEE Conference on Computer Vision and Pattern Recognition},
  pages 1912--1920, 2015.

\bibitem{xie2015learning}
J.~Xie, W.~Hu, S.-C. Zhu, and Y.~N. Wu.
\newblock Learning sparse {FRAME} models for natural image patterns.
\newblock {\em International Journal of Computer Vision}, 114(2-3):91--112,
  2015.

\bibitem{xie2016cooperative}
J.~Xie, Y.~Lu, R.~Gao, S.-C. Zhu, and Y.~N. Wu.
\newblock Cooperative training of descriptor and generator networks.
\newblock {\em arXiv preprint arXiv:1609.09408}, 2016.

\bibitem{xie2016inducing}
J.~Xie, Y.~Lu, S.-C. Zhu, and Y.~N. Wu.
\newblock Inducing wavelets into random fields via generative boosting.
\newblock {\em Applied and Computational Harmonic Analysis}, 41(1):4--25, 2016.

\bibitem{XieLuICML}
J.~Xie, Y.~Lu, S.-C. Zhu, and Y.~N. Wu.
\newblock A theory of generative convnet.
\newblock In {\em International Conference on Machine Learning}, 2016.

\bibitem{zhu2003statistical}
S.-C. Zhu.
\newblock Statistical modeling and conceptualization of visual patterns.
\newblock {\em IEEE Transactions on Pattern Analysis and Machine Intelligence},
  25(6):691--712, 2003.

\bibitem{zhu1998filters}
S.~C. Zhu, Y.~Wu, and D.~Mumford.
\newblock Filters, random fields and maximum entropy {(FRAME)}: Towards a
  unified theory for texture modeling.
\newblock {\em International Journal of Computer Vision}, 27(2):107--126, 1998.

\end{thebibliography}
}

\end{document}